%% file: paper.tex
\documentclass{article}

\usepackage[preprint]{temp}

\usepackage{hyperref}
\usepackage{lipsum}
\usepackage{natbib}
\usepackage{float, caption}
\usepackage{threeparttable}
\usepackage{algorithm}
\usepackage{algorithmicx}
\usepackage{algpseudocode}
\usepackage{caption}
\usepackage[utf8]{inputenc} 
\usepackage[T1]{fontenc}    
\usepackage{hyperref}       
\usepackage{url}            
\usepackage{booktabs}       
\usepackage{amsfonts}        
\usepackage{nicefrac}        
\usepackage{microtype}      
\usepackage{xcolor,caption}   
\usepackage{subfigure}
\usepackage{amsmath,amsthm,graphics,subcaption}
\input{math_commands.tex}

\usepackage{tikz}
\usepackage[most]{tcolorbox}
\usetikzlibrary{shapes, arrows, positioning}
\tikzset{>=stealth}
\usepackage{graphicx}

\title{Controllable Coupled Image Generation via Diffusion Models
}

\author{
  Chenfei Yuan\thanks{Chenfei Yuan, Nanshan Jia and Hangqi Li contributed equally to this work.} \quad
  Nanshan Jia\footnotemark[1]\hspace{1ex}\thanks{Corresponding author. UC Berkeley IEOR and BAIR. nsjia@berkeley.edu}\quad 
  Hangqi Li\footnotemark[1] \quad
  Peter W. Glynn \quad
  Zeyu Zheng \\
}

\begin{document}

\maketitle

\begin{abstract}
  We provide an attention-level control method for the task of coupled image generation, where “coupled” means that multiple simultaneously generated images are expected to have the same or very similar backgrounds. While backgrounds coupled, the centered objects in the generated images are still expected to enjoy the flexibility raised from different text prompts. The proposed method disentangles the background and entity components in the model’s cross-attention modules, attached with a sequence of time-varying weight control parameters depending on the time step of sampling. We optimize this sequence of weight control parameters with a combined objective that assesses how coupled the backgrounds are as well as text-to-image alignment and overall visual quality. Empirical results demonstrate that our method outperforms existing approaches across these criteria.
\end{abstract}

\section{Introduction}
\label{Introduction}
Diffusion models, since \cite{ho2020denoising,song2020denoising,song2020score,lipman2022flow}, have emerged as a powerful tool for generative modeling, achieving remarkable success in generating high-quality images and videos given textual prompts. By iteratively denoising random noise guided by learned data distributions, these models have surpassed prior generative models, such as GAN \cite{goodfellow2014generative,goodfellow2020generative} and variational autoencoders \cite{kingma2013auto}, in both sample quality and diversity \cite{dhariwal2021diffusion}. Their ability to synthesize diverse and realistic visual content has led to rapid adoption in a wide range of creative and industrial applications, including digital art, animation, design, and content creation \cite{rombach2022high,saharia2022photorealistic}.

On top of these advancements, researchers and practitioners have expressed growing desires and interests in more-controllable generation \cite{epstein2023selfguidance,qin2023unicontrol,gaintseva2025casteer}. More-controllable generation refers to the ability to influence and constrain specific aspects of the generated output, such as maintaining very similar backgrounds \cite{hertz2022prompt}, ensuring precise object placement \cite{tewel2024add,liu2024diffpop,ma2024directed}, or independently modulating style and content \cite{wu2023uncovering,wang2023stylediffusion,qi2024deadiff,jiang2024artist,tarres2024parasol}. This type of fine-grained control, if manageably achieved, can be desirable for certain applications with demand for consistency, reliability, and interpretability, such as media synthesis, image editing, and design automation \cite{zhang2023controlvideo,zhang2023adding,mou2024t2i}. 

In this work, we study an image generation task called coupled image generation, which aims to achieve more controllability in terms of simultaneously generating multiple images with the same or highly similar backgrounds. The input consists of prompts that describe similar backgrounds but differ in the main objects, which we referred to as the entities. This difference may arise from entirely different entities, or from the same entity presented with variations in angle, position, or appearance. While the backgrounds across generated images are expected to be coupled and controlled, the entities are expected to vary according to the prompt-specific description. Below we present an example of the coupled image generation task.

\begin{figure}[htbp]
    \centering
    \begin{minipage}{0.8\textwidth}
        \centering
        \textbf{Baseline method (Flux) for coupled image generation}
        \vspace{0.2cm}
    \end{minipage}
    \begin{minipage}{0.32\textwidth}
        \subfigure[Prompt: A cute Pikachu sits on the floor of a cozy room bathed in warm sunshine. The room has wooden flooring and a peaceful, homely atmosphere.]{\includegraphics[width=\linewidth]{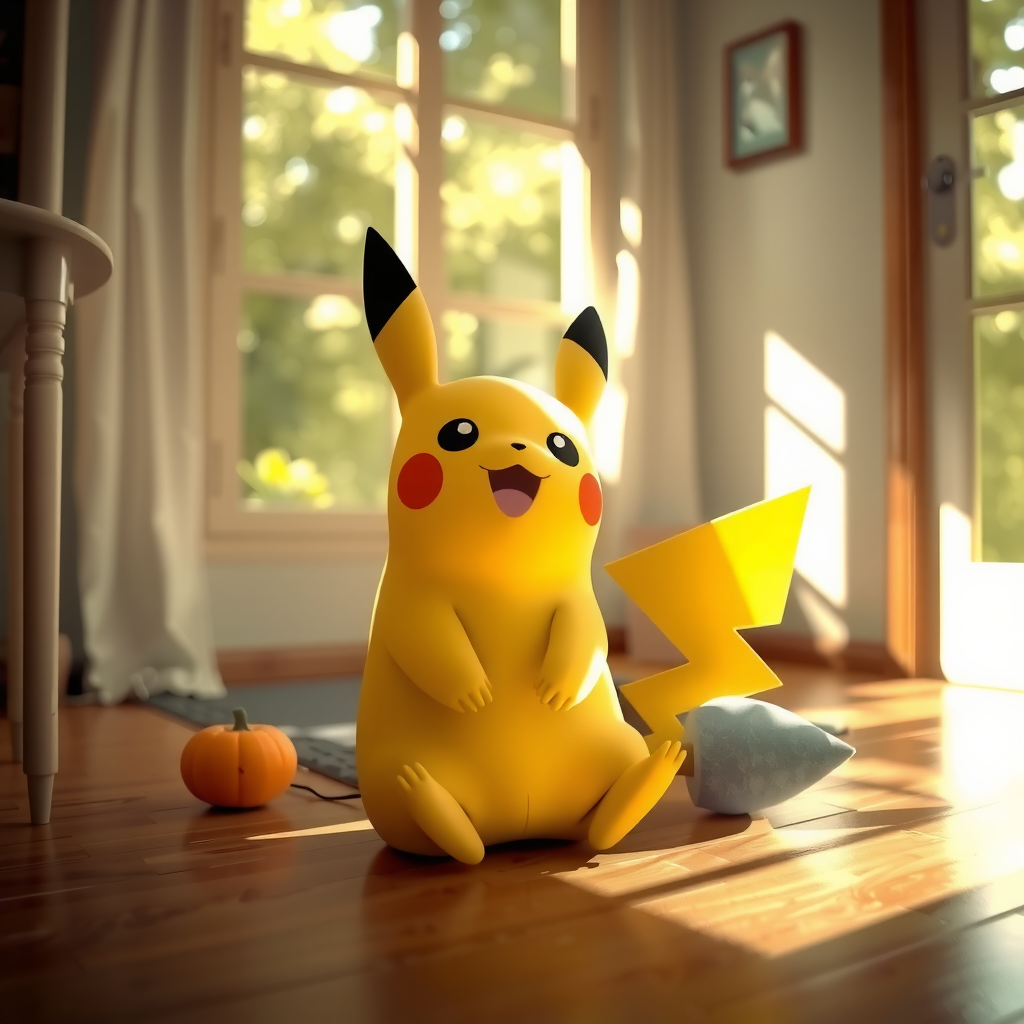}}
    \end{minipage}
    \hfill
    \begin{minipage}{0.32\textwidth}
        \subfigure[Prompt: A cute Pikachu \textbf{stands} on the floor of a cozy room bathed in warm sunshine. The room has wooden flooring and a peaceful, homely atmosphere.]{\includegraphics[width=\linewidth]{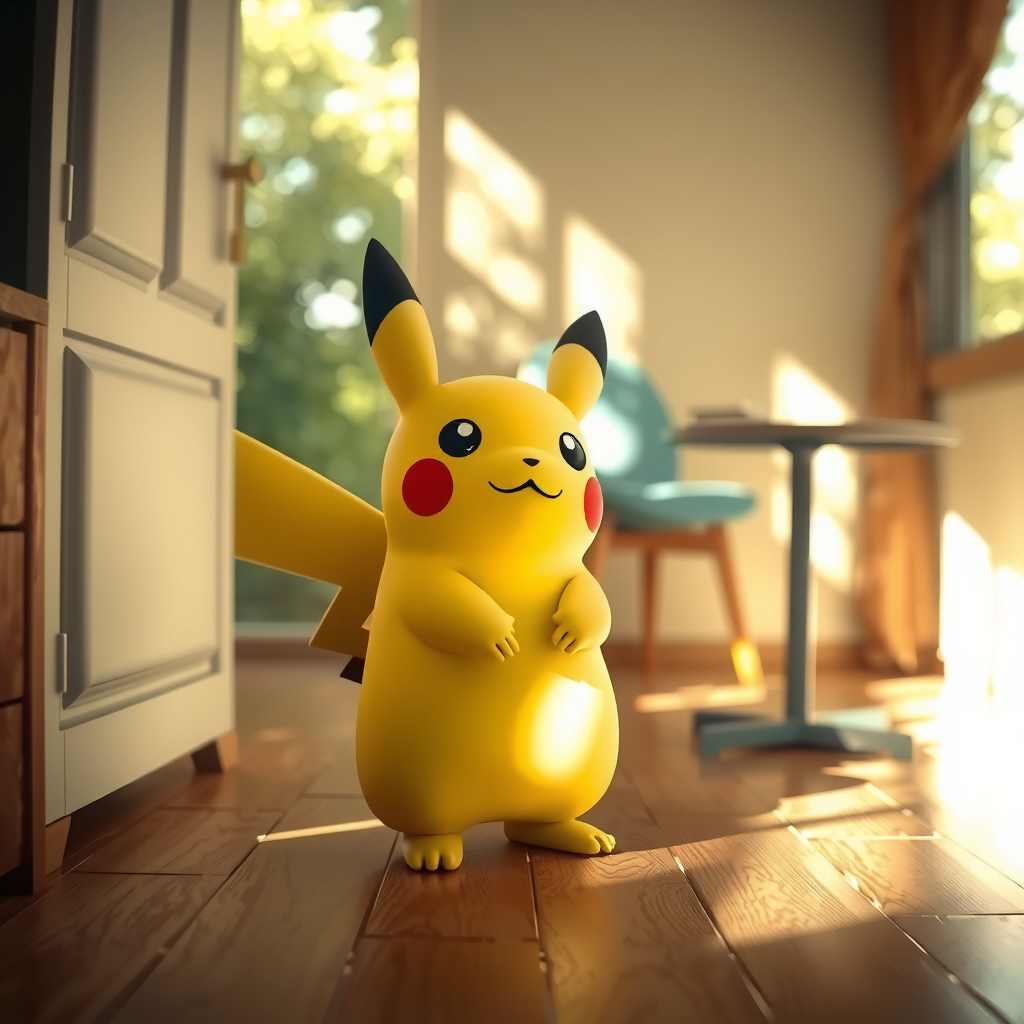}}
    \end{minipage}
    \hfill
    \begin{minipage}{0.32\textwidth}
    \subfigure[Prompt: A \textbf{beautiful girl} stands on the floor of a cozy room bathed in warm sunshine. The room has wooden flooring and a peaceful, homely atmosphere.]{
        \includegraphics[width=\linewidth]{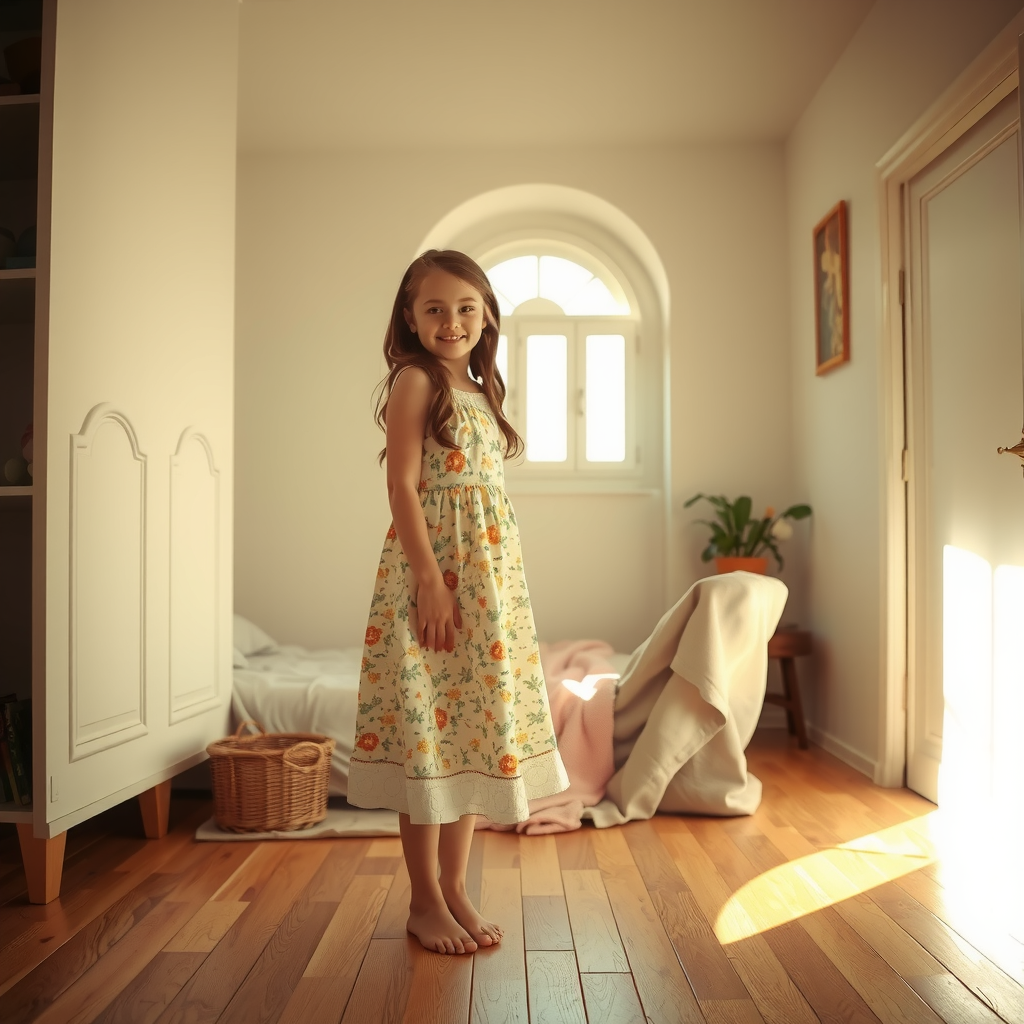}}
    \end{minipage}
    \hfill
    \begin{minipage}{0.8\textwidth}
        \centering
        \vspace{0.2cm}
        \textbf{Our method for coupled image generation}
               \vspace{0.2cm}
    \end{minipage}
    \begin{minipage}{0.32\textwidth}
    \subfigure[Prompt: A cute Pikachu sits on the floor of a cozy room bathed in warm sunshine. The room has wooden flooring and a peaceful, homely atmosphere.]{
        \includegraphics[width=\linewidth]{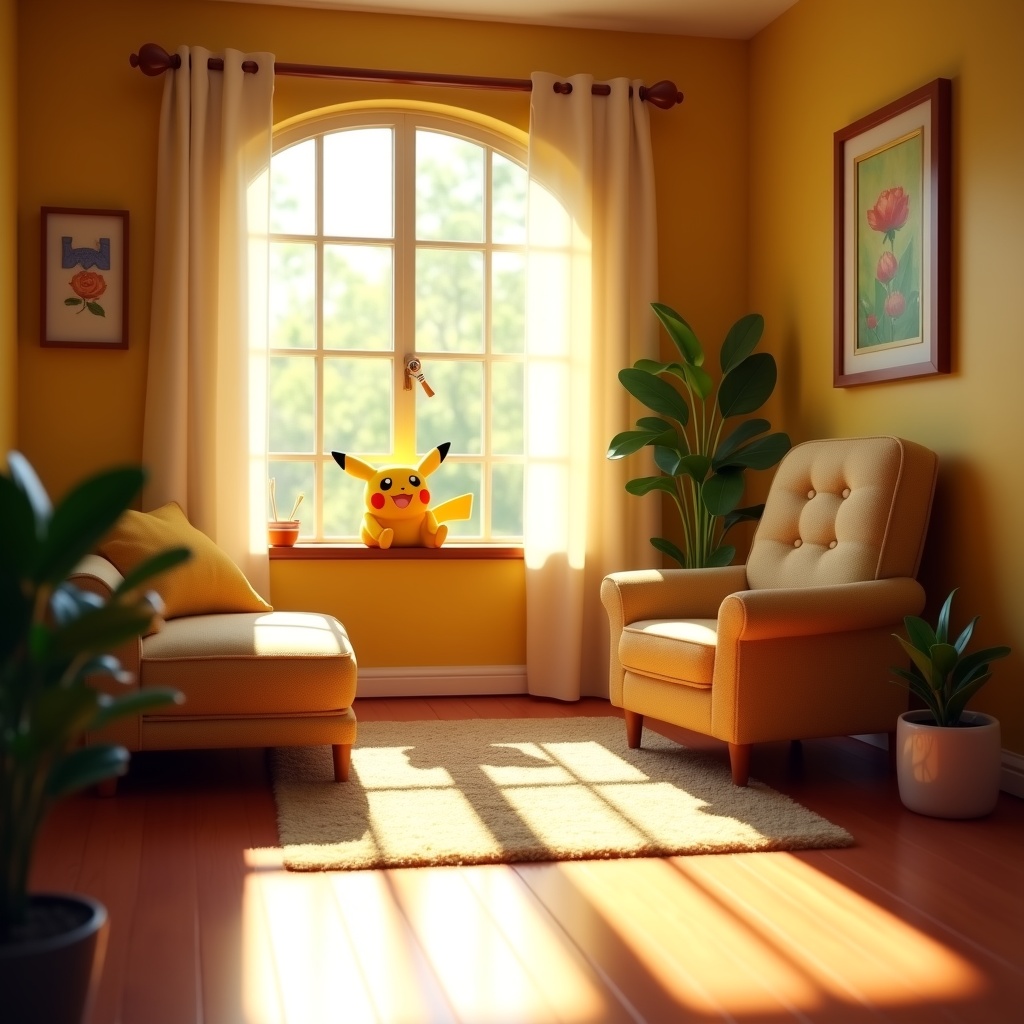}}
    \end{minipage}
    \hfill
        \begin{minipage}{0.32\textwidth}
        \subfigure[Prompt: A cute Pikachu \textbf{stands} on the floor of a cozy room bathed in warm sunshine. The room has wooden flooring and a peaceful, homely atmosphere.]{
        \includegraphics[width=\linewidth]{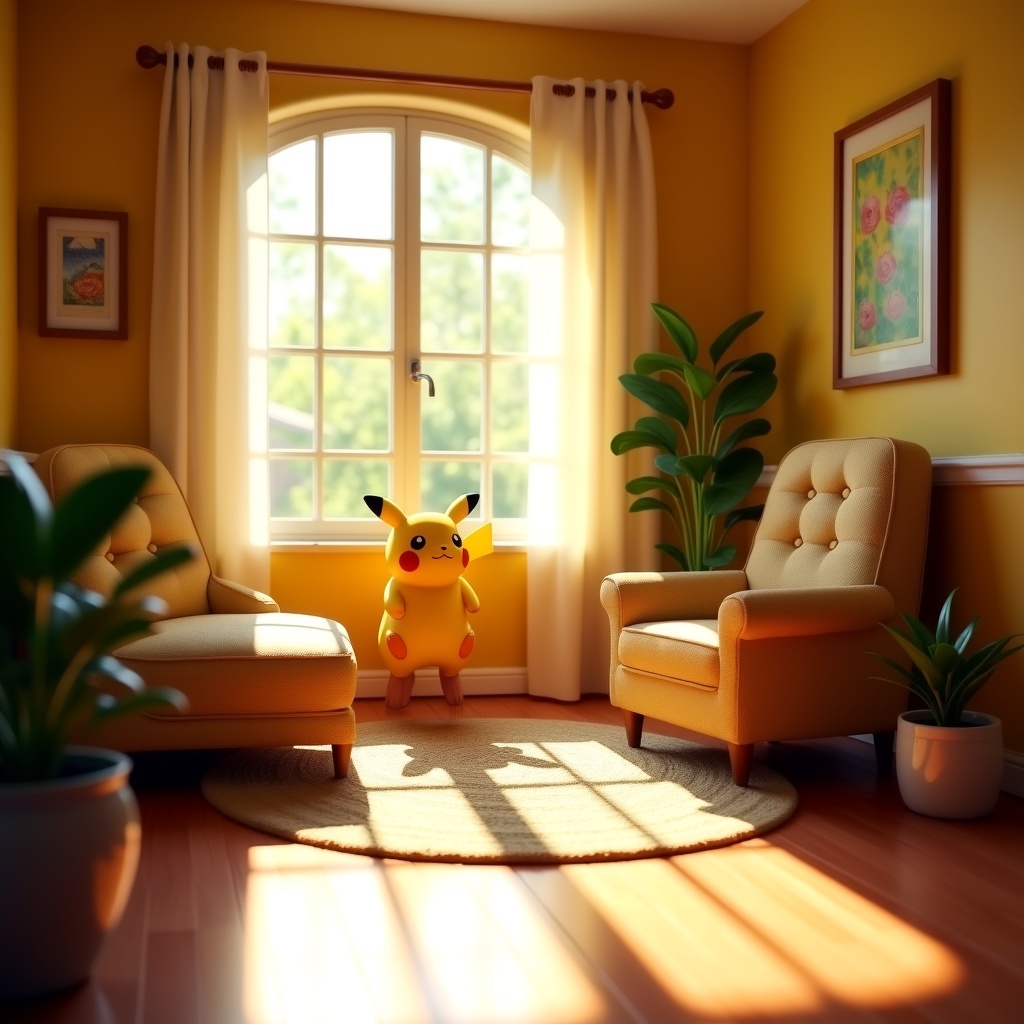}}
        
    \end{minipage}
    \hfill
        \begin{minipage}{0.32\textwidth}
        \subfigure[Prompt: A \textbf{beautiful girl} stands on the floor of a cozy room bathed in warm sunshine. The room has wooden flooring and a peaceful, homely atmosphere.]{
                \includegraphics[width=\linewidth]{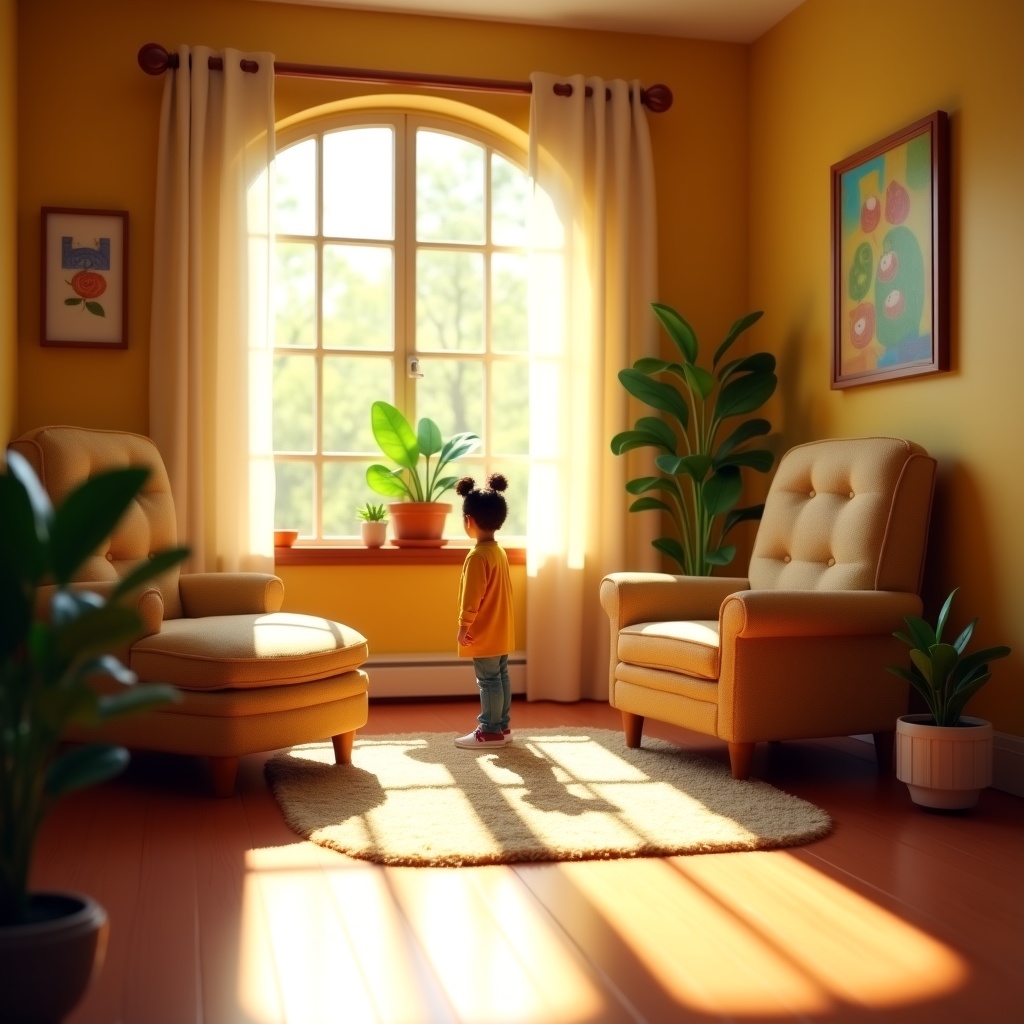}}
    \end{minipage}
    \caption{The first row illustrates that by given prompts that have almost the same description and differ only slightly in the subject, the pictures generated by Flux~\cite{flux1dev2024} have significant background differences, even after using the same random seeds in every step and using the same sampling settings. The second row illustrates the coupled generated images using our method, where the generated images share very similar backgrounds. 
}
    \label{example;couple}
\end{figure}

Coupled image generation has several practical applications. For example, some video generation models take the first and last frames as input to generate intermediate frames while ensuring temporal coherence \cite{kuaishou2024kling,jain2024video,liu2024mardini,wan2025flf2v}. In such cases, maintaining a highly similar background across these frames is essential for temporal consistency among frames. Coupled image generation also benefit 3D modeling, particularly in multi-view reconstruction tasks. As noted in \cite{hu20203d,caliskan2020multi}, accurate 3D reconstruction typically necessitates multiple views of an object captured from different angles. Ensuring a consistent and similar background across these views contribute to isolating the object's shape and enhancing stability during the modeling process. Additionally, coupled image generation can support image editing tasks. Certain editing applications require modifying only the foreground while preserving the background. For example, \cite{li2023layerdiffusion} proposes methods that modify the primary object while keeping the rest of the scene features consistent. These tasks typically rely on datasets consisting of paired images with similar backgrounds and varying foreground entities. Coupled image generation offers an efficient tool to construct such datasets.

Prior to our work, several methods have been proposed that directly or indirectly address the coupled image generation task. These approaches can be broadly classified into three categories: image composition, image editing, and attention control. Image composition methods \cite{ye2023ip,huggingface:flux_controlnet_inpainting} typically begin with a given image depicting the background and then insert an entity based on the input prompt. The insertion step either adopts inpainting-based techniques or employs models specifically designed for direct object addition. These methods require additional input on the background image and often struggle with maintaining global consistency, such as lighting, shadows, and texture, resulting in visual artifacts. Image editing techniques for coupled image generation \cite{rout2024semantic,blackforestlabs2025fluxredux} typically generate images separately. In practice, these methods often begin by generating an image based on one of the prompts and then editing it to align with the descriptions provided by other prompts. While such methods allow for localized control, they often require substantial manual input and tend to generalize not well across diverse prompts or complex scenes. Moreover, these methods are computationally less efficient, as each image must be generated individually. Finally, attention editing methods \cite{hertz2022prompt,yang2023dynamic,wang2024tokencompose,chen2024training} conduct image generation by manipulating the cross-attention maps within diffusion models to emphasize specific content. Techniques such as prompt-to-prompt editing \cite{hertz2022prompt} and token-level attention injection \cite{wang2024tokencompose} enable partial control over shared backgrounds by implicitly disentangling background and foreground descriptions. However, these methods are often sensitive to prompt formulation and may struggle to preserve spatial or semantic consistency when applied across multiple generations. We leave a more thorough discussion on related literature in the supplementary materials.

To address the limitations of existing methods, we propose a cross-attention control framework for coupled image generation. Our approach introduces a combined evaluation metric that jointly measures background similarity and text-to-image alignment for generated images. Inspired by prior work on attention editing, we first employ a pre-trained large language model (LLM) to explicitly disentangle the input prompts. The LLM is prompted to extract a shared background prompt and an entity prompt from each input. This decomposition enables separate control over the background and entity during generation. We then modify the cross-attention module in the diffusion model to accept two text sources. Specifically, we introduce an auxiliary parameter that controls the relative influence of the background and entity prompts within the cross-attention mechanism. To optimize this parameter, we formulate an isotonic optimization problem guided by the proposed evaluation metric. Empirically, we show that our method achieves both high background similarity and strong text-image alignment with minimal additional training. We summarize our contributions as follows: 

\begin{enumerate}
\item We formalize the task of coupled image generation, where multiple prompts share a very similar background while differing in foreground entities. To evaluate performance on this task, we propose an evaluation metric that jointly captures background similarity and text-image alignment.

\item After explicitly disentangling each prompt into a background and an entity component, we adapt the model’s cross-attention layers to handle them jointly and integrate a sequence of time-varying tunable parameters controlling their relative weights.

\item We frame the learning of this sequence of parameters as an isotonic optimization problem, which allows for efficient and stable training. Experiments across standard benchmarks show improvements in both background similarity and alignment with entity prompts. Such improvements are achieved without losing inference and sampling efficiency.
\end{enumerate}
\section{Background on Cross-attention Mechanism}
Cross-attention is an attention mechanism that allows one sequence to attend to another sequence. Instead of computing attention within a single sequence like self-attention, cross-attention aligns the information from one sequence to another through the calculation of attention. This enables the model to dynamically focus on the most relevant parts of the input when generating or interpreting output. The concept was first introduced by \cite{bahdanau2014neural} in the context of neural machine translation, where a decoder RNN is set to align with the encoder hidden states. Since then, cross-attention has become a core architectural component in text-to-image generation. In this setting, it enables alignment between visual features and the input text, allowing the model to condition each generated visual element on specific words or phrases from the prompt. This ensures that the output image accurately and coherently reflects the described content. 

In what follows, we briefly introduce two types of cross-attention mechanisms commonly used in text-to-image generation models.

\begin{enumerate}
    \item \textbf{QKV-level concatenation-based cross-attention.} This cross-attention mechanism first computes the (query, key, value) tuple for both text embeddings and image hidden state. Then, queries, keys and values are concatenated before sending to the attention mechanism. The mathematical formulation is given by
    \begin{equation}
        \mathrm{Attention}(Q_{\text{joint}},K_{\text{joint}},V_{\text{joint}})=\softmax \left(\frac{Q_{\text{joint}}K_{\text{joint}}^\top}{\sqrt{d_{\text{text}}+d_{\text{img}}}}\right)V_{\text{joint}},\label{joint;cattn;v1}
    \end{equation}
    where $Q_{\text{joint}}=[Q_{\text{text}},Q_{\text{img}}]$, $K_{\text{joint}}=[K_{\text{text}},K_{\text{img}}]$, $V_{\text{joint}}=[V_{\text{text}},V_{\text{img}}]$, $d_{\text{text}}$ is the hidden dimension of textual prompts and $d_{\text{img}}$ is the hidden dimension of image hidden state. Following the cross-attention module, the outputs are split into text and image components, each of which is independently processed by subsequent projection layers.

    \item \textbf{Embedding-level concatenation-based cross-attention.} Embeddings from different modalities are first concatenated into a unified sequence. Then, a standard self-attention is applied uniformly across this sequence. Although the concatenated sequence is used for self-attention, the method acts as cross-attention because attention weights can form between tokens from different modalities. This allows each modality to attend to the other. Specifically, define $X_{\text{concat}}=[X_{\text{text}};X_{\text{img}}]$ and calculate corresponding query, key and value, where subscript "concat" are added. The attention comes as
    \begin{equation}
        \mathrm{Attention}(Q_{\text{concat}},K_{\text{concat}},V_{\text{concat}})=\softmax \left(\frac{Q_{\text{concat}}K_{\text{concat}}^\top}{\sqrt{d_{\text{concat}}}}\right)V_{\text{concat}},
    \end{equation}
    where $d_{\text{concat}}$ is the hidden dimension of the concatenated embedding $X_{\text{concat}}$. The attention output from the cross-attention module is further processed by subsequent layers, ultimately producing the output of the block.
\end{enumerate}
For each type of cross-attention, we will introduce our cross-attention control method in Section \ref{Cross-attention Control}.

\section{Problem Formulation}
\label{Problem Formulation}
We now formalize the coupled image generation problem. The term ``coupled" refers to generating multiple images simultaneously that share the same or very similar backgrounds while differing in their depicted entities, based on input prompts. Suppose that we have a pre-trained text-to-image generation model $G: \mathbb{T} \to \mathbb{I}$, where $\mathbb{T}$ is the prompt space and $\mathbb{I}$ is the space of generated images, we parameterize it with an auxiliary vector $\boldsymbol{\theta}\in\boldsymbol{\Theta}$ and denote it by $G_{\boldsymbol{\theta}}$. The detailed parameterization method is included in Section \ref{Cross-attention Control} later. Then, given a batch size $n$ and input prompts $T_1, T_2, \cdots, T_n \in \mathbb{T}$, we feed these inputs to $G_{\boldsymbol{\theta}}$ and denote the generated images by $I_j = G_{\boldsymbol{\theta}}(T_j)$ for $j=1,2,\cdots,n$.

To evaluate whether the generated images meet the requirements of coupled image generation problem, we define the following objectives:
\begin{enumerate}
\item \textbf{Background similarity.} A function $f_{bg}: \mathbb{T}^n\times\mathbb{I}^n \to \mathbb{R}^+$ that measures the degree of background similarity across the generated samples. Higher values indicate stronger background similarity across the function inputs.
\item \textbf{Text-image alignment.} A function $f_{ti}: \mathbb{T} \times \mathbb{I} \to \mathbb{R}^+$, where $f_{ti}(T, I)$ quantifies how well image $I$ aligns with prompt $T$. Higher values indicate stronger alignment between the image and the prompt.
\end{enumerate}

Given weight parameters $\lambda_{bg}, \lambda_{ti}$, we define the combined objective $f_c: \mathbb{T}^n\times \boldsymbol{\Theta} \to \mathbb{R}^+$ as
\begin{equation}
f_c(T_1,\cdots,T_n,\boldsymbol{\theta}) = \lambda_{bg} f_{bg}\left(T_1,\cdots,T_n,G_{\boldsymbol{\theta}}(T_1),\cdots, G_{\boldsymbol{\theta}}(T_n)\right) + \frac{\lambda_{ti}}{n} \sum_{j=1}^n f_{ti}\left(T_j, G_{\boldsymbol{\theta}}(T_j)\right).
\end{equation}

The training of $\boldsymbol{\theta}$ is then formulated as solving the optimization problem:
\begin{equation}
\underset{\boldsymbol{\theta}\in\boldsymbol{\Theta}}{\max}  \frac{1}{M} \sum_{k=1}^M f_c(T_{k,1}, \cdots, T_{k,n},\boldsymbol{\theta}),
\end{equation}
where $\{T_{k,1}, \cdots, T_{k,n}\}$ denotes the $k$-th training set.

\section{Our Method}
As shown in Section~\ref{Introduction} and Figure \ref{example;couple}, the backgrounds of images genrated by the pre-trained model $G$ exhibit noticeable visual differences, even when the prompts describe the same or highly similar background and the random seed is manually fixed. We attribute this phenomenon to the lack of explicit disentanglement between background and entity in most text-to-image generation models, such as Flux \cite{flux2024} and Stable Diffusion \cite{rombach2022high}. The presence of different entities in the prompt can affect spatial layout, lighting and texture through the cross-attention modules during each denoising step. With accumulation occurring across all steps during inference, this affection eventually results in variations and differences in the background of the generated images and the overall image composition.

To overcome this entanglement in prompt, we first propose a disentanglement method in Section \ref{Disentangling Prompts into Background Prompts and Entity Prompts}, which extracts the shared background information from all prompts and presents entity prompt for each individual prompt. Then, we introduce our method to incorporate both background prompt and entity prompt in Section \ref{Cross-attention Control}. We finally formalize the isotonic optimization problem in Section \ref{Formulation of an Isotonic Problem to Optimize a Parameterized Model}.

\subsection{Explicitly Disentangling Prompts into Background Prompts and Entity Prompts}
\label{Disentangling Prompts into Background Prompts and Entity Prompts}
\begin{figure}[htbp]
\centering
\begin{tikzpicture}[
    user/.style={rectangle, rounded corners=3mm, fill=gray!15, draw=gray, minimum width=12cm, align=left, text width=11cm, inner sep=5pt},
    llm/.style={rectangle, rounded corners=3mm, fill=gray!15, draw=gray, minimum width=12cm, align=left, text width=11cm, inner sep=5pt},
    textnode/.style={rectangle, draw=none, fill=none, text width=14cm, align=center},
    node distance=1.2cm and 1.2cm
]

    \node[user] (user1) {
        \textbf{User Input:}\\[2pt]
        Prompt 1: A cute Pikachu sits in a cozy room bathed in warm sunshine. The room has wooden flooring and a peaceful, homely atmosphere. \\
        Prompt 2: A beautiful girl stands in a cozy room bathed in warm sunshine. The room has wooden flooring and a peaceful, homely atmosphere.
    };

    \node[textnode, below=1.2cm of user1] (textlabel) {
        Pre-processing through an LLM with pre-specified decomposition instructions
    };

    \node[llm, below=of textlabel] (llm1) {
        \textbf{LLM Output (then as textual prompts to coupled text-to-image generation):}\\[2pt]
        Background: A cozy room bathed in warm sunshine with wooden flooring and a peaceful, homely atmosphere.\\
        Entity 1: A cute Pikachu sits.\\
        Entity 2: A beautiful girl stands.
    };

    \draw[->, thick] (user1.south) -- (textlabel.north);
    \draw[->, thick] (textlabel.south) -- (llm1.north);

\end{tikzpicture}
    \caption{A zero-shot example of prompting a pre-trained LLM to extract the shared background from a set of prompts and present the distinct entity descriptions for each individual prompt. The pre-specified prompt for LLM can be found in the supplementary materials.}
    \label{example;prompt;2}

\end{figure}
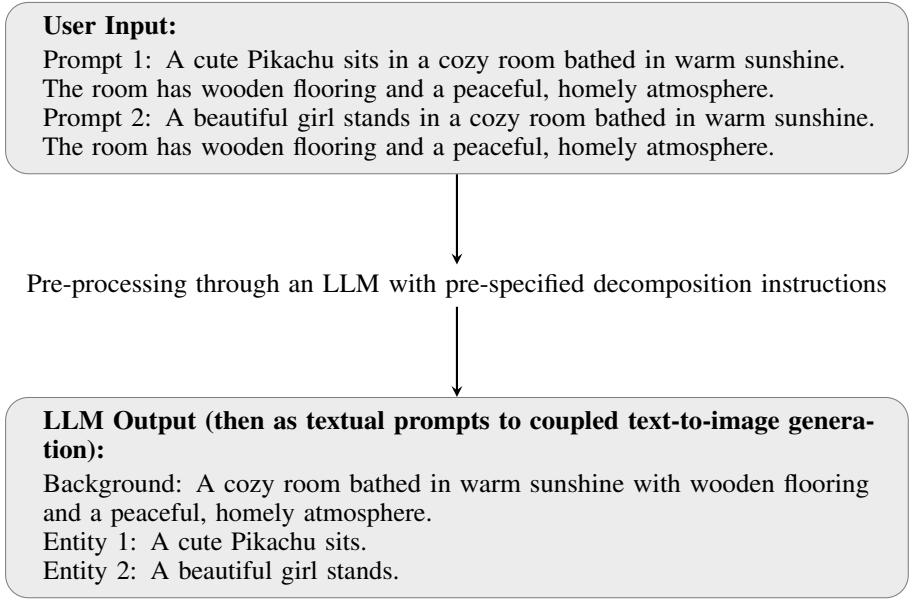

Our method begins by reorganizing the input prompts to disentangle shared and distinct components. Specifically, given input prompts $T_1, \cdots, T_n \in \mathbb{T}$, we use a pre-trained large language model to extract a shared background prompt and distinct entity prompts. The model outputs a background prompt $T_{bg}$ and a set of entity prompts $T_1^{\text{ent}}, \cdots, T_n^{\text{ent}}$. The background prompt captures the shared background information across all inputs and each entity prompt describes the unique entity in its corresponding input. A concrete example of this decomposition process is illustrated in Figure \ref{example;prompt;2}.

\subsection{Cross-attention Control}
\label{Cross-attention Control}
Most cross-attention modules in text-to-image generation models only takes one text embedding and one image hidden state as input. Given our scope of coupled image generation, after the prompt being disentangled into background prompt and entity prompt, we modify the cross-attention module to enable two input text embeddings. In addition, as outlined in Section \ref{Problem Formulation}, we want the generated images to share a very similar background with high alignment between prompt and image. Hence, we have to control the relative weight of background prompt and entity prompt. To this end, we introduce our cross-attention control method for different types of cross-attention as follows. For ease of exposition, we will use subscript bg, ent and img to denote background, entity and image, respectively.

\begin{enumerate}
\item \textbf{QKV-level concatenation-based cross-attention.} To extend the QKV-level concatenation-based cross-attention mechanism, we propose to concatenate the query, key, and value vectors across the background embedding, entity embedding, and image hidden states. Specifically, we define
\begin{equation}
    \mathrm{Attention}(Q_{\text{joint}}, K_{\text{joint}}(\theta), V_{\text{joint}} )
    = \softmax \left(\frac{Q_{\text{joint}} K_{\text{joint}}^\top(\theta)}{\sqrt{d_{\text{text}}+d_{\text{img}}}}\right)V_{\text{joint}}, \label{joint;cattn;v2}
\end{equation}
where $\theta \in [0, 1]$, $Q_{\text{joint}} = [Q_{\text{bg}}, Q_{\text{ent}}, Q_{\text{img}}]$, $K_{\text{joint}}(\theta) = [(1{-}\theta)K_{\text{bg}}, \theta K_{\text{ent}}, K_{\text{img}}]$, and $V_{\text{joint}} = [V_{\text{bg}}, V_{\text{ent}}, V_{\text{img}}]$.

Next, we elaborate on the parameterization in (\ref{joint;cattn;v2}). There are two key design choices. First, we scale the key vectors by $(1{-}\theta)$ for the background and by $\theta$ for the entity. The auxiliary parameter $\theta$ serves as a soft weighting factor between the background and the entity. When $\theta$ is close to 1, the entity contributes more prominently to the attention scores. Conversely, when $\theta$ is close to 0, the background dominates.

Second, although the concatenated query, key, and value vectors have total dimension $2d_{\text{text}}+d_{\text{img}}$, we normalize the dot product in the softmax using $\sqrt{d_{\text{text}}+d_{\text{img}}}$ instead of $\sqrt{2d_{\text{text}}+d_{\text{img}}}$. This choice of normalization by $\sqrt{d_{\text{text}}+d_{\text{img}}}$ comes from the fact that only the background and entity contribute to the modulated key vectors in the attention score. Since the image component remains unscaled, using $\sqrt{d_{\text{text}}+d_{\text{img}}}$ instead of $\sqrt{2d_{\text{text}}+d_{\text{img}}}$ ensures that the overall scale of the attention scores remains consistent with the original case (\ref{joint;cattn;v1}). In addition, this normalization ensures that, when $\theta = 1$, the attention weights are reduced exactly to the original cross-attention weights (\ref{joint;cattn;v1}) using only the entity embedding, and, when $\theta = 0$, they are reduced to the cross-attention weights that only use the background embedding. Together, these two design choices allow our parameterization to provide fine-grained control over attention composition while preserving semantic consistency in the output.

\item \textbf{Embedding-level concatenation-based cross-attention.} In the embedding-level concatenation-based cross-attention, directly concatenating the background and entity embeddings with the image hidden state would alter the architecture, as the multi-layer perceptron following the cross-attention module is designed to accept inputs of dimension $2d$. To avoid the need for auxiliary training or architectural adjustments in each block, we propose a method that accommodates both background and entity embeddings while preserving the original architecture. We first perform embedding-level concatenation:
\begin{equation}
    X_{\text{bg-img}} = [X_{\text{bg}}; X_{\text{img}}], \quad X_{\text{ent-img}} = [X_{\text{ent}}; X_{\text{img}}],
\end{equation}
where $X_{\text{img}}$ is the image hidden state, and $X_{\text{bg}}, X_{\text{ent}}$ are the background and entity embeddings, respectively. We then compute the cross-attention separately for each concatenated input:
\begin{align}
    \mathrm{Attention}(Q_{\text{bg-img}}, K_{\text{bg-img}}, V_{\text{bg-img}}) &= \softmax \left( \frac{Q_{\text{bg-img}} K_{\text{bg-img}}^\top}{\sqrt{d_{\text{concat}}}} \right) V_{\text{bg-img}}, \\
    \mathrm{Attention}(Q_{\text{ent-img}}, K_{\text{ent-img}}, V_{\text{ent-img}}) &= \softmax \left( \frac{Q_{\text{ent-img}} K_{\text{ent-img}}^\top}{\sqrt{d_{\text{concat}}}} \right) V_{\text{ent-img}}.
\end{align}

After the cross-attention operation, the background-image and entity-image attention outputs are propagated separately through subsequent layers. Let the outputs be denoted as:
\begin{equation}
    X_{\text{bg-img}}' = [X_{\text{bg}}'; X_{\text{img-bg}}], \quad X_{\text{ent-img}}' = [X_{\text{ent}}'; X_{\text{img-ent}}].
\end{equation}

We then merge the image hidden states by weighted interpolation:
\begin{equation}
    X_{\text{img}}' = \theta X_{\text{img-ent}} + (1 - \theta) X_{\text{img-bg}},
\end{equation}
and redefine the block outputs as \begin{equation}
    X_{\text{bg-img}}' = [X_{\text{bg}}'; X_{\text{img}}'], \quad X_{\text{ent-img}}' = [X_{\text{ent}}'; X_{\text{img}}'].
\end{equation}
This formulation ensures that the image hidden state remains unified and consistent across both branches after each block. Each block output serves as the input to the subsequent block.

We also note that, although the introduction of $\theta$ occurs after the cross-attention module, the method still enjoys the following property: when $\theta = 1$, the entity-image branch reduces exactly to the original block output that uses only the entity embedding; and when $\theta = 0$, the background-image branch reduces to the output of the original cross-attention block that uses only the background embedding. This design ensures compatibility with existing architectures while enabling interpolation between background embedding and entity embedding in shaping and guiding the image hidden states.
\end{enumerate}

\subsection{Formulation of an Isotonic Problem to Optimize a Parameterized Model}
\label{Formulation of an Isotonic Problem to Optimize a Parameterized Model}
Given a pre-trained text-to-image generation model $G$, we first introduce our parameterization method. Assume a given time schedule that supports an $N$-step inference. We propose a parameterized model $G_{\boldsymbol{\theta}}$, where $\boldsymbol{\theta}\in\boldsymbol{\Theta}$, such that each cross-attention module at the $i$-th inference step is parameterized by $\boldsymbol{\theta}_i$, following the approach detailed in Section~\ref{Cross-attention Control}. Inspired by previous findings~\cite{choi2022perception,park2023understanding,yue2024exploring,wang2024towards} that diffusion models initially focus on generating coarse structure and subsequently refining fine-grained details, we impose an ascending order constraint on the parameter sequence $\{\boldsymbol{\theta}_i\}$ to align with this progressive refinement process and stabilize the training. That said, we formulate the problem of learning $\boldsymbol{\theta}$ as an isotonic optimization problem:
\begin{align}
    \underset{\boldsymbol{\theta}}{\max}  &\frac{1}{M} \sum_{k=1}^M f_c(T_{k,1}, \cdots, T_{k,n},\boldsymbol{\theta}),\\
    \text{subject to } &\boldsymbol{\theta}_1\leq\boldsymbol{\theta}_2\leq\cdots\leq \boldsymbol{\theta}_N,\nonumber\\
    &\boldsymbol{\theta}\in\boldsymbol{\Theta}.\nonumber
\end{align}

In the constraints, we impose an ascending order on the parameter sequence $\{\boldsymbol{\theta}_i\}$. This ordering reflects the growing influence of entity-specific information over time, alongside the diminishing effect of background information. It reveals how the model gradually shifts its attention from shared background features in the early inference steps to entity-specific details in the later steps. This progressive shift ensures that the generated images share a consistent background while achieving high text-image alignment through the later incorporation of entity prompts.

\section{Experiments}
\subsection{Experimental Settings}
\subsubsection{Base Text-to-Image Generation Model}
Throughout the paper, we use Flux.1dev \cite{flux1dev2024} as the base model $G$. In the Flux architecture, there are 19 double-stream blocks and 38 single-stream blocks. Each double-stream block employs QKV-level concatenation-based cross-attention, followed by independent feedforward layers and layer normalization for the text embedding and image hidden states separately. After the double-stream blocks, the text embedding is concatenated with the image hidden states and passed to the single-stream blocks. These blocks use embedding-level concatenation-based cross-attention, followed by a unified feedforward network and layer normalization. We consider generating images with resolution $1024\times1024$ with $50$ steps. Guidance scale is set to be $3.5$. In this case, the time-varying parameter $\boldsymbol{\theta}$ is a 50-dimensional vector under the isotomic constraint $$0\leq\boldsymbol{\theta}_1\leq\boldsymbol{\theta}_2\leq\cdots\leq\boldsymbol{\theta}_{50}\leq1$$. At the $i$-th step, all cross-attention modules are parameterized with $\boldsymbol{\theta}_i$ following the cross-attention control method introduced in Section \ref{Cross-attention Control}. All experiments are conducted on a single NVIDIA A100 GPU.

\subsubsection{Implementation Details on Proposed Metrics.}
We have proposed a combined metric to evaluate both background similarity and text-image alignment in Section~\ref{Problem Formulation}. In this section, we introduce our implementation of the functions $f_{bg}$ and $f_{ti}$. We begin with $f_{bg}$. For each prompt $T_j$, we follow the proposed method to generate the corresponding image $I_j$. We then use the Segment Anything model~\cite{kirillov2023segment} to segment the foreground entity. Specifically, let $A_j$ denote the region in image $I_j$ that is segmented as the entity. We define the Joint Entity Region (JER) as 
\begin{equation}
\text{JER} = A_1 \cup A_2 \cup \cdots \cup A_n.
\end{equation}
For each image, we apply the same mask over the JER. We then compute the average $\mathbb{L}^2$ distance between the masked images. To account for the proportion of unmasked regions, we define a validity ratio:
\begin{equation}
    R = 1 - \frac{\# \text{JER}}{\text{width} \times \text{height}},
\end{equation}
where $\# \text{JER}$ denotes the number of pixels contained in the JER, and \text{width}, \text{height} refer to the resolution of the generated images.

We scale the average $\mathbb{L}^2$ distance by this ratio and define the background similarity function as
\begin{equation}
    f_{bg} =- \frac{2}{n(n-1)R} \sum_{1 \leq j < k \leq n} \left\| \text{mask} \circ I_j - \text{mask} \circ I_k \right\|^2,
\end{equation}
where $\text{mask} \circ$ denotes the masking operation on the JER, and $\|\cdot\|$ denotes the $\mathbb{L}^2$ norm.

For $f_{ti}$, we adopt the CLIPScore \cite{radford2021learning,hessel2021clipscore}. CLIPScore is a reference-free evaluation metric for text-to-image generation that uses the CLIP model's cross-modal embedding space. It operates by encoding both the image and the associated textual prompt into a shared semantic space using CLIP’s pre-trained encoders. The metric then computes the cosine similarity between these two embeddings to quantify the degree to which the image content aligns with the textual description. A higher CLIPScore typically indicates better text-image alignment. In the following experiments, we always choose $\lambda_{bg}=300$ and $\lambda_{ti}=\frac{1}{30}$ for proper normalization.

\subsection{Comparison to Other Methods}
\begin{figure}[!htbp]
    \centering
    \begin{minipage}[b]{0.24\textwidth}
        \centering
        \textbf{Ours}
        \vspace{0.2cm}
    \end{minipage}
    \begin{minipage}[b]{0.24\textwidth}
        \centering
        \textbf{Random seed}
               \vspace{0.2cm}
    \end{minipage}
    \begin{minipage}[b]{0.24\textwidth}
        \centering
        \textbf{P2P}
               \vspace{0.2cm}
    \end{minipage}
    \begin{minipage}[b]{0.24\textwidth}
        \centering
        \textbf{RF-inversion}
               \vspace{0.2cm}
    \end{minipage}
    \begin{minipage}[b]{0.24\textwidth}
        \centering
        \subfigure{
            \includegraphics[width=\textwidth]{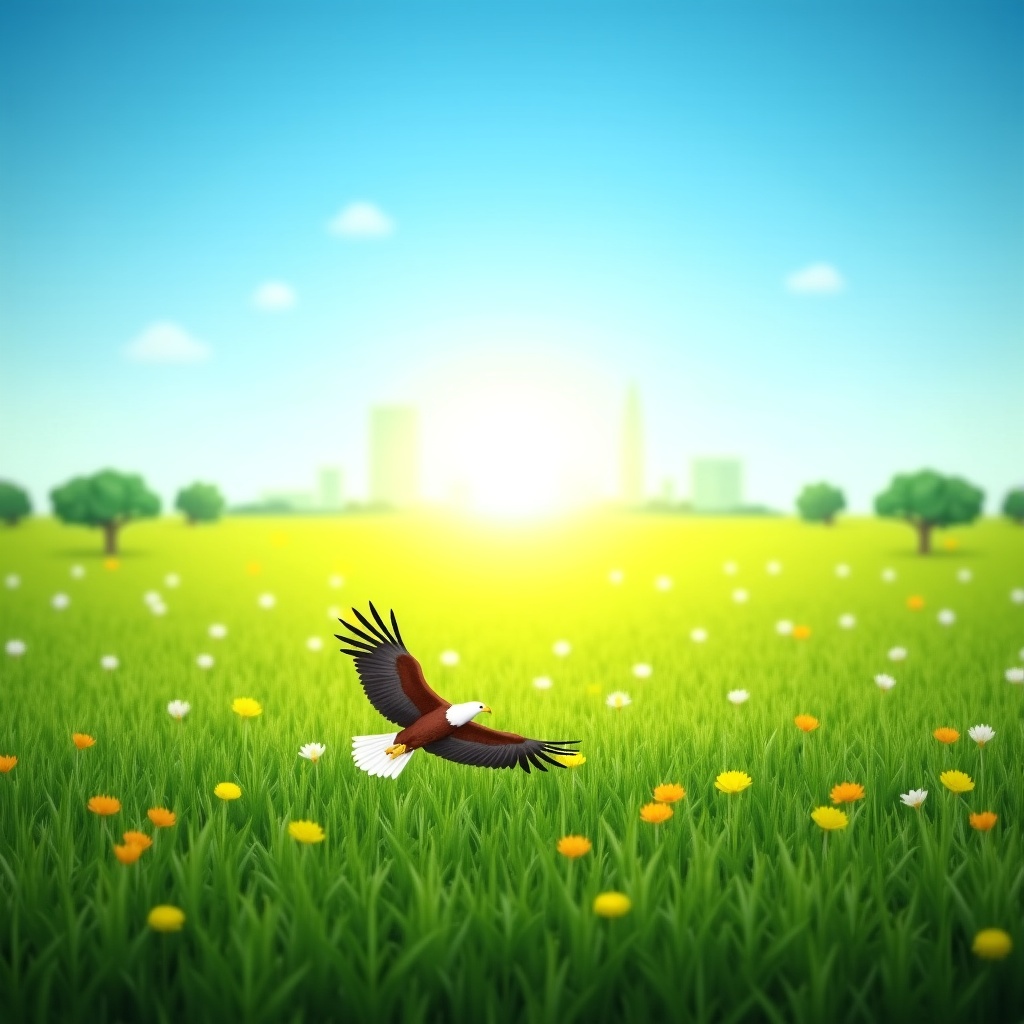}
        }
    \end{minipage}
    \begin{minipage}[b]{0.24\textwidth}
        \centering
        \subfigure{
            \includegraphics[width=\textwidth]{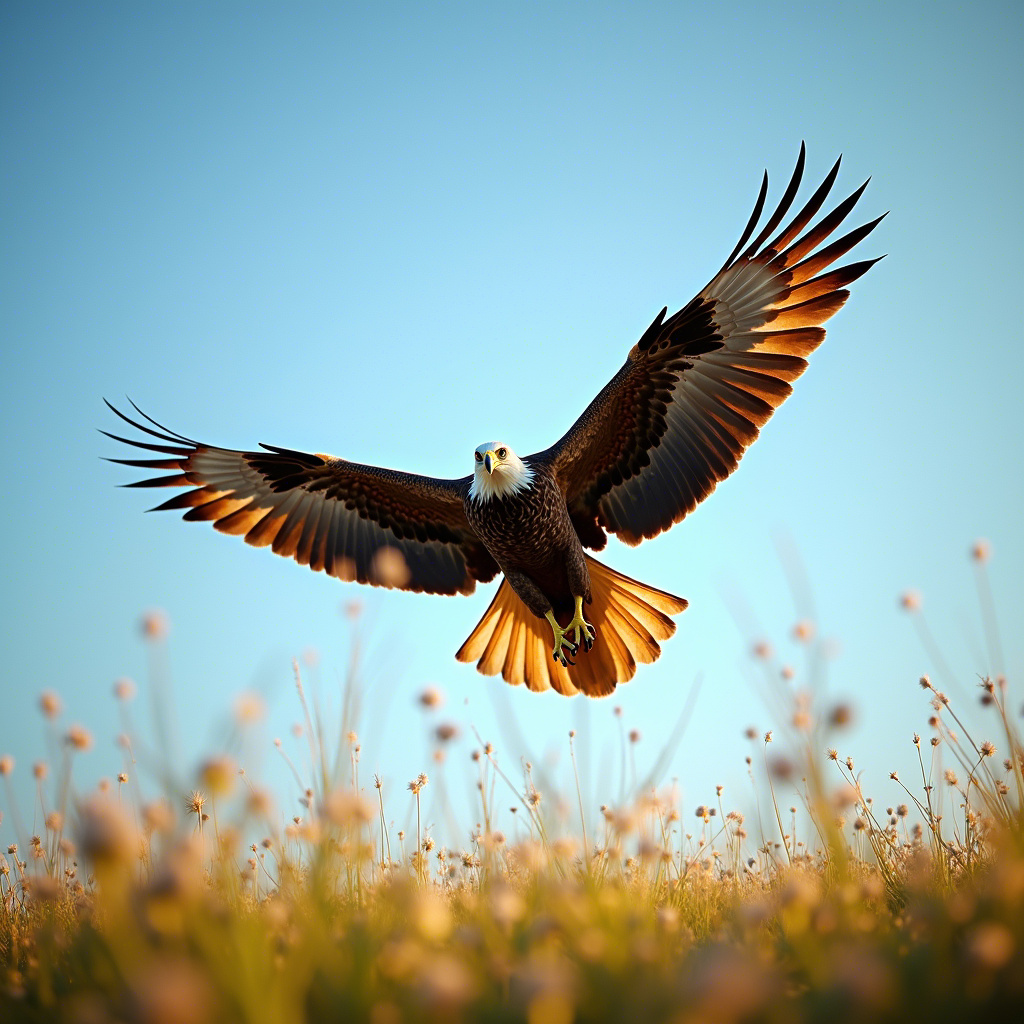}
        }
    \end{minipage}
    \begin{minipage}[b]{0.24\textwidth}
        \centering
        \subfigure{
            \includegraphics[width=\textwidth]{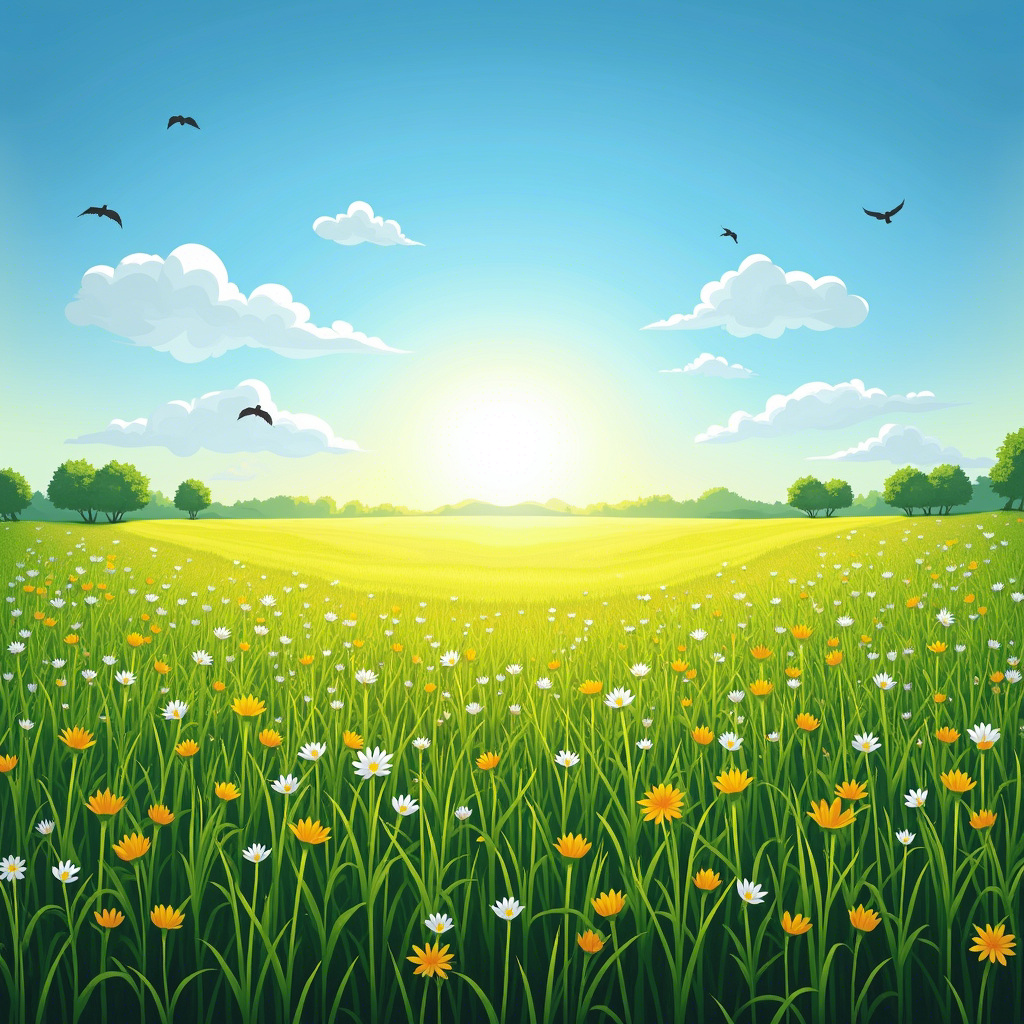}
        }
    \end{minipage}
    \begin{minipage}[b]{0.24\textwidth}
        \centering
        \subfigure{
            \includegraphics[width=\textwidth]{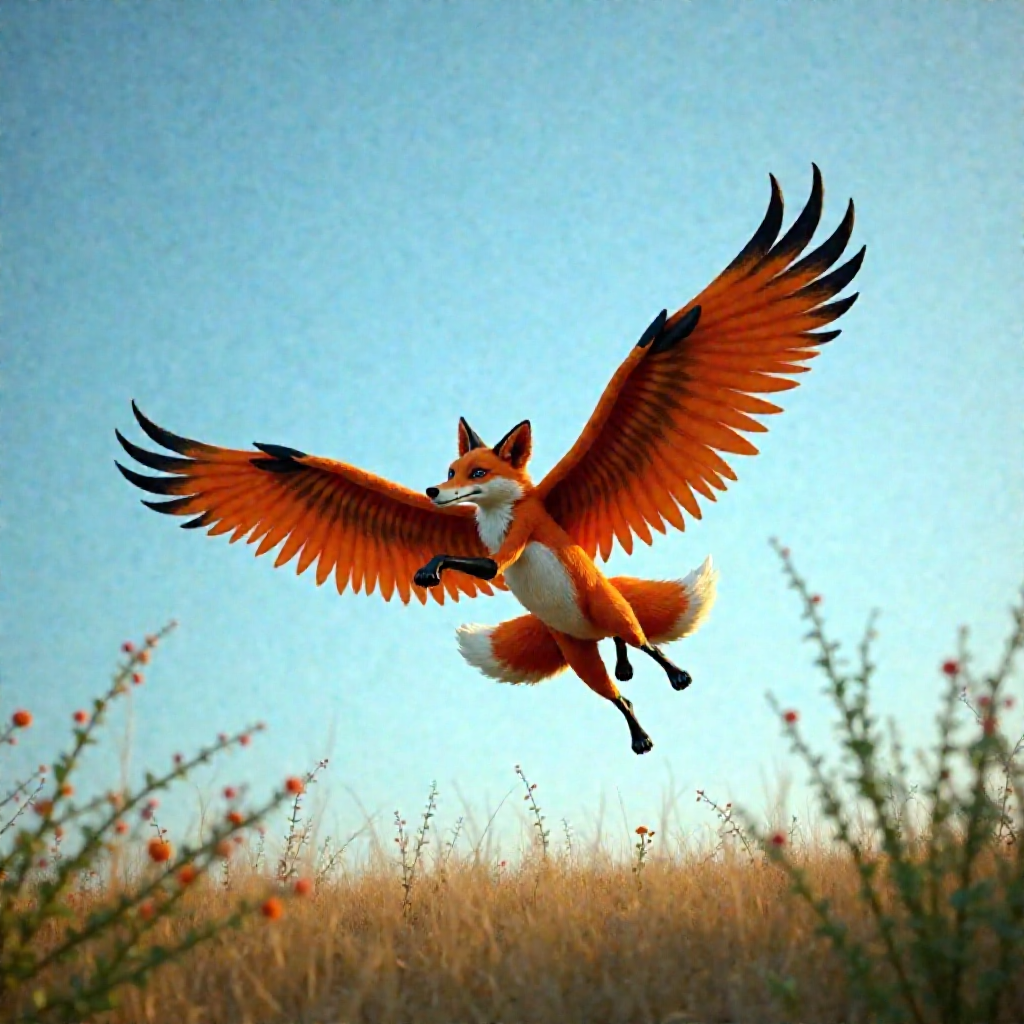}
        }
    \end{minipage}
    \begin{minipage}[b]{0.24\textwidth}
        \centering
        \subfigure{
            \includegraphics[width=\textwidth]{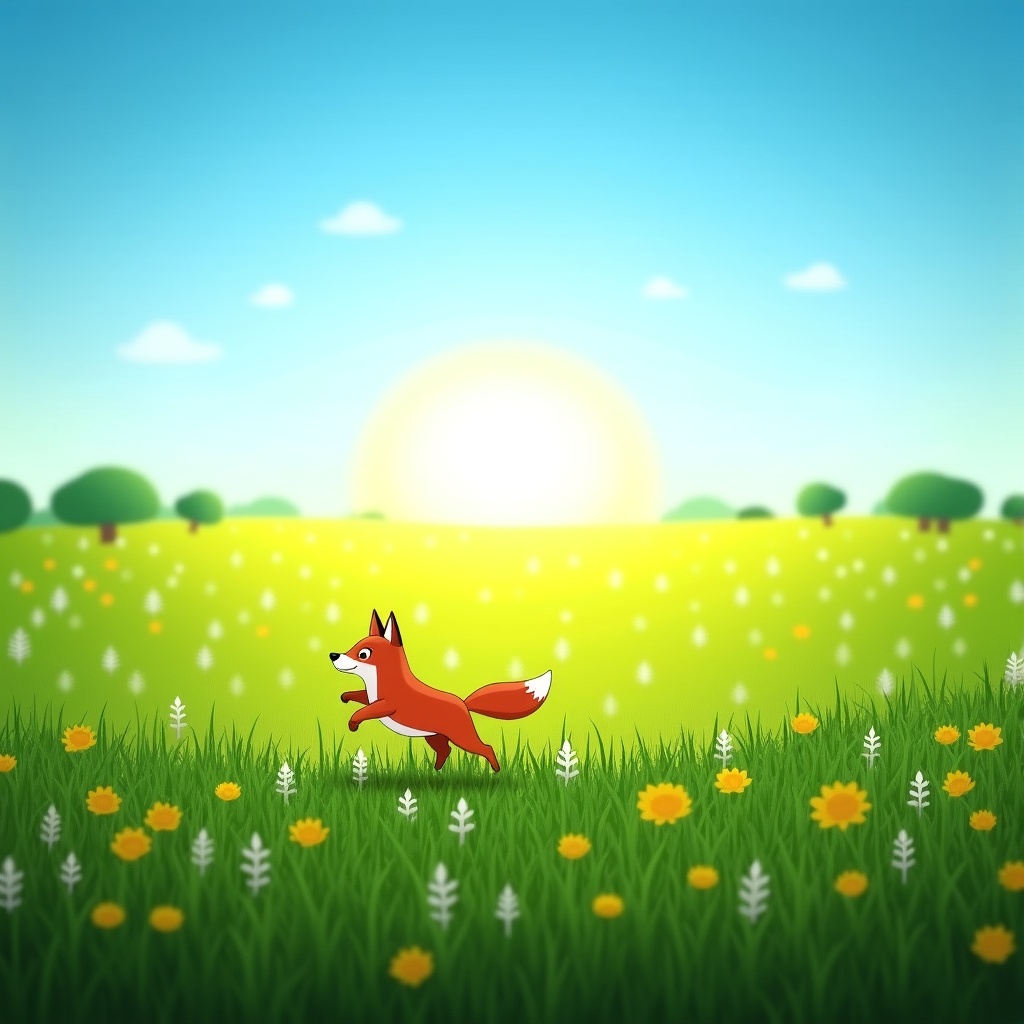}
        }
    \end{minipage}
    \begin{minipage}[b]{0.24\textwidth}
        \centering
        \subfigure{
            \includegraphics[width=\textwidth]{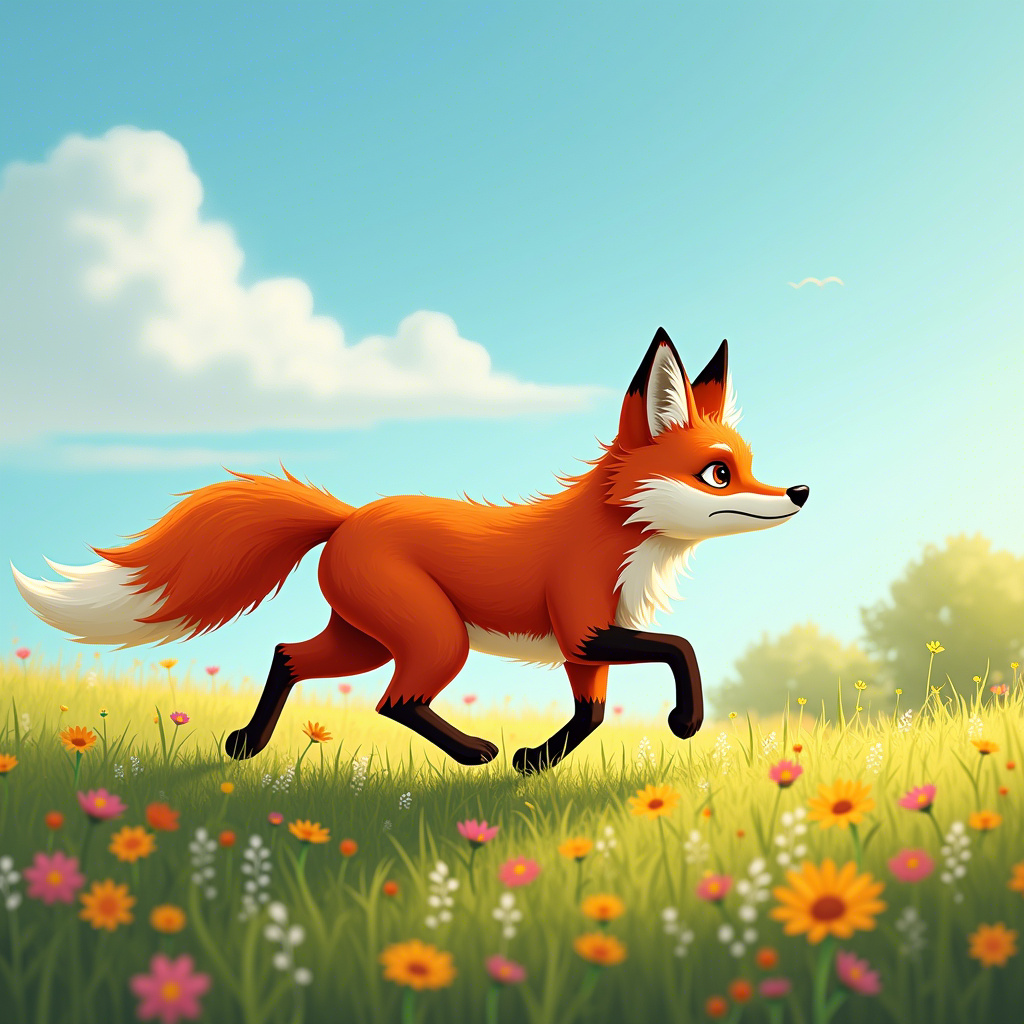}
        }
    \end{minipage}
    \begin{minipage}[b]{0.24\textwidth}
        \centering
        \subfigure{
            \includegraphics[width=\textwidth]{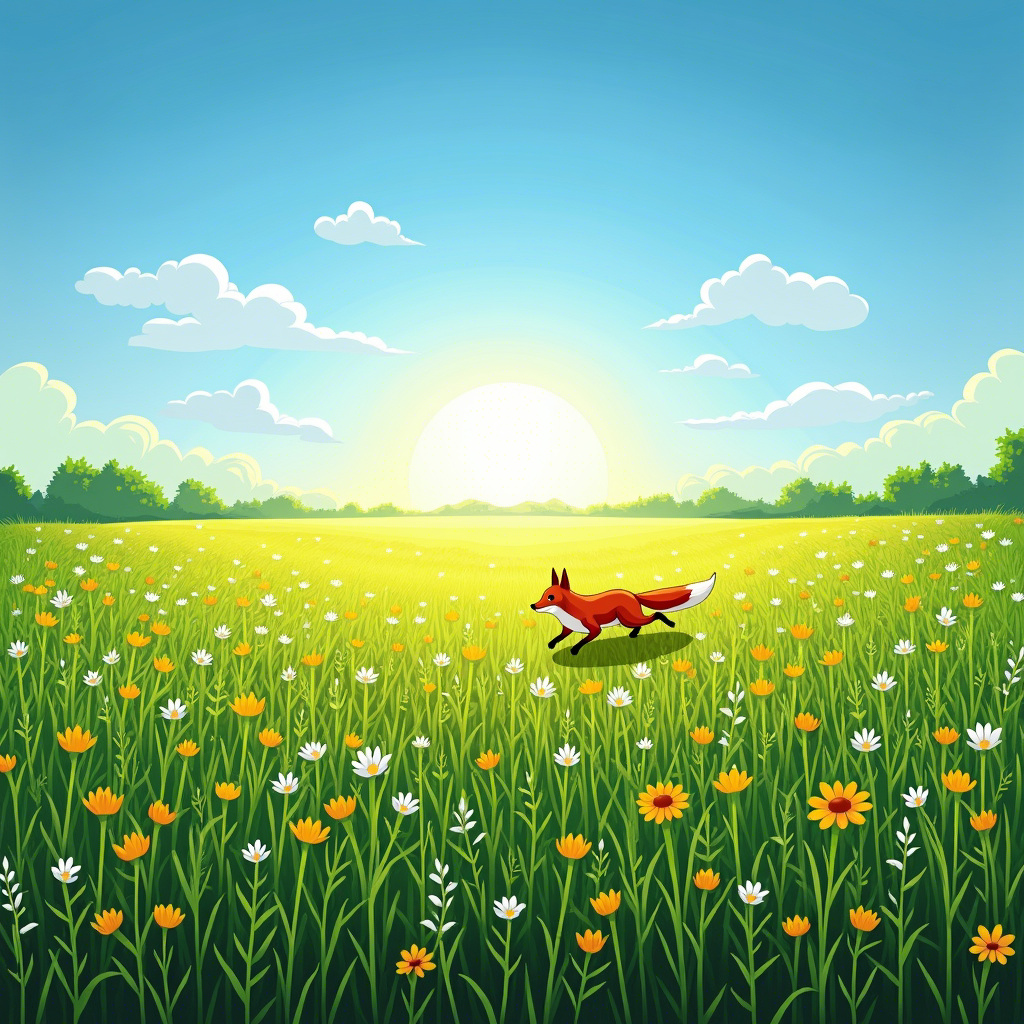}
        }
    \end{minipage}
    \begin{minipage}[b]{0.24\textwidth}
        \centering
        \subfigure{
            \includegraphics[width=\textwidth]{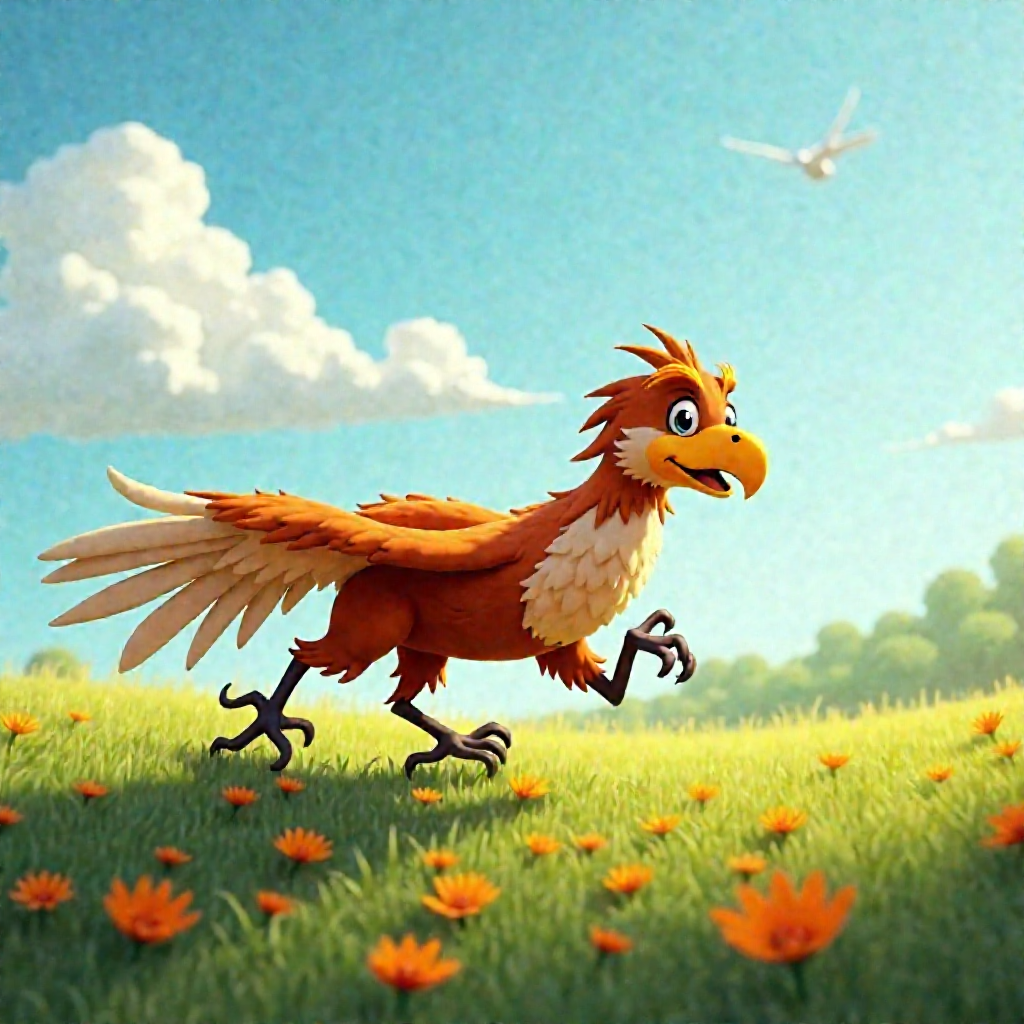}
        }
    \end{minipage}
    \caption{Comparison of our method to other baselines. Generated pictures in the first row correspond to prompt: \textit{"An eagle soars gracefully in a vibrant meadow under a clear blue sky, bathed in sunlight."}. The second row corresponds to prompt: \textit{"A red fox runs in a vibrant meadow under a clear blue sky, bathed in sunlight."} Our method performs better than baselines in both background similarity and text-image alignment, where our method and all baseline methods share the same sampling procedures and random numbers.}
    \label{celebahq-fig}
\end{figure}

In this section, we compare our method with existing approaches for the coupled image generation task. We consider the following baselines:
\begin{enumerate}
    \item \textbf{Random seed.} This baseline generates images using the Flux model with a fixed random seed to encourage consistency across outputs.
    \item \textbf{P2P.} The Prompt-to-Prompt method~\cite{hertz2022prompt} edits the cross-attention maps within the Flux model to enforce alignment between similar prompts.
    \item \textbf{RF-inversion.} The RF-inversion method~\cite{rout2024semantic} first generates an image from a given prompt and then applies an inversion-based editing process to align it with other prompts.
\end{enumerate}
We present the generated images in Figure~\ref{celebahq-fig}. The prompts used are: \textit{"An eagle soars gracefully in a vibrant meadow under a clear blue sky, bathed in sunlight."} and \textit{"A red fox runs in a vibrant meadow under a clear blue sky, bathed in sunlight."}. We observe that both our method and P2P produce images with highly similar backgrounds. In contrast, editing-based methods, such as RF-inversion, fail to ensure strong background similarity. Moreover, our method outperforms P2P in terms of text-image alignment, as it more accurately and clearly depicts the specified entities. Overall, our method yields the highest combined metric. Additionally, we report the evaluation metrics for all methods in Table~\ref{metrics}. These data supports our visual observations.

\begin{table}[htbp]
\centering
\caption{Metrics for coupled image generation}
\begin{tabular}{ccccccc}\toprule
\label{metrics}
Metrics \textbackslash Model. &  Ours & Random seed & P2P & RF-inversion \\\midrule
Background similarity ($\times 10^{-4}$) ($\uparrow$) &  -2.080 & -67.42 & -4.604 & -48.47 \\
Text-image alignment ($\uparrow$) & 22.61 & 18.75 & 18.14 & 19.59 \\
Combined metric ($\uparrow$) & 1.558 & -0.919 & 1.246 & -0.214
\\\bottomrule
\end{tabular}
\end{table}

\section{Conclusion and Future Work}
\label{Conclusion and Future Work}
In this work, we formalize the coupled image generation task, which aims to generate multiple images simultaneously while preserving a highly similar background. We propose a combined metric that jointly measures background similarity and text-image alignment. This metric can be used to evaluate the performance on coupled image generation task. To achieve coupled generation, we first explicitly disentangle each input prompt into a shared background prompt and individual entity prompts. We then introduce a cross-attention control mechanism that enables the model to process and balance these disentangled components. A sequence of time-varying parameters is introduced to control the relative influence of background and entity prompts across the inference steps. We formulate the learning of these parameters as an isotonic optimization problem to enhance training stability and efficiency. Through extensive comparisons with prior methods, we demonstrate that our approach improves both background similarity and text-image fidelity. Notably, our method does not sacrifice inference or sampling efficiency.

We would love to point out a few limitations of our work. First, our work only handles text-to-image generation and does not consider the task of image generations with multi-modal inputs. Second, the approach introduces time-varying parameters on cross-attention mechanisms, which may introduce scalability issues when generating complex scenes or long sequences (e.g., video frames), where balancing multiple components becomes more challenging.

\bibliographystyle{plainnat}  
\bibliography{reference}
\appendix
\newpage
\section{Literature Review}
Prior to our work, several methods have been proposed that, while not explicitly designed for coupled image generation, can nevertheless be adapted to this task due to their compatible input-output frameworks. These existing approaches can be broadly classified into three categories: image inpainting and composition methods, image editing methods, and attention editing methods. Image inpainting and composition methods typically require a background image as input. They then compose entities guided by textual prompts onto the background image. Image editing methods initially generate an image based on one prompt and subsequently modify this image to align with additional prompts while preserving the background context. Attention editing methods specifically target the cross-attention modules within model architectures and control background consistency through the manipulation of cross-attention maps. In the following sections, we provide a detailed literature review of each of these method categories.
\subsection{Image Inpainting and Composition}
Image inpainting techniques allow regenerating missing or removed regions of an image with plausible content, which can be leveraged to replace a foreground object while preserving the original background. Early deep learning approaches like Context Encoders used convolutional auto-encoders with adversarial loss to fill large missing regions \cite{kohler2014mask,xu2014deep,yeh2017semantic,liu2018image}. Follow-up works addressed limitations such as blurriness and artifacts. For example, partial convolution layers were introduced to condition only on valid pixels and ignore masked-out regions. This improves inpainting of irregular holes. Another milestone was using contextual attention to explicitly copy information from distant background patches into the hole \cite{pathak2016context,yu2018generative}, which greatly improved structural coherence. Subsequent models incorporated multiple stages and dual discriminators \cite{iizuka2017globally,yu2018generative} to ensure both local detail and global consistency. These advances enable removing an object from an image and hallucinating the covered background, providing a clean slate on which a new foreground can be inserted.

In parallel, image composition methods focus on inserting a new object into a background image in a realistic manner. This involves determining a plausible location, scale, and blending for the foreground object \cite{lee2018context,avrahami2022blended,ye2023ip,huggingface:flux_controlnet_inpainting}. Traditional “copy-paste” \cite{mortensen1995intelligent} often yields implausible results, so learning-based methods emerged to automate object placement and harmonization. For instance, ST-GAN (Spatial Transformer GAN) applied a learned geometric transform to warp a foreground object such that it fits naturally into the target scene \cite{li2023mt}. Later approaches used conditional GANs and transformers to model semantic context: given a background (or its segmentation map), they predict where an object of a certain class could be placed and even generate its shape if needed. These models learn common-sense placement by analyzing the background context. Beyond placement, the composite must be visually seamless. Research on image harmonization and blending addresses color and illumination matching, as well as boundary smoothing and shadow generation for inserted objects. Incorporating these aspects, multitask frameworks have been proposed to produce realistic results \cite{dong2024internlm}.

\subsection{Image Editing}
A broad class of image editing methods allows modifying an image’s content while keeping other aspects unchanged. Early conditional generative models demonstrated this capability on constrained domains: for instance, facial attribute editing GANs could add or remove features like glasses or hair color “by only changing what you want” and leaving all other facial details intact \cite{xu2018attngan}.  Another line of work uses direct image-to-image diffusion or transformer models conditioned on editing instructions \cite{kawar2023imagic,zhang2023adding,rout2024semantic,rout2024semantic,mou2024t2i,blackforestlabs2025fluxredux,liu2025step1x}. These models take an image and a high-level edit description, then produce a new image with similar background and different entity. For example, \cite{brooks2023instructpix2pix} is a diffusion-based editor trained on synthetic “before and after” image pairs with text instructions. 

\subsection{Attention Editing}
Attention editing method for coupled image generation leverage the internal attention mechanisms of deep generative models to control what content is preserved or changed between images \cite{hertz2022prompt,yang2023dynamic,cao2023masactrl,mokady2023null,kim2023dense,chen2024training,phung2024grounded,wang2024tokencompose}. In particular, diffusion models with cross-attention (e.g. Stable Diffusion \cite{rombach2022high}, Imagen) allow a fine-grained alignment between text tokens and image regions. Researchers have found that by intervening in these attention maps, one can achieve text-driven edits that lock certain parts of the image in place. For example, \cite{hertz2022prompt} demonstrates that simply replacing a word in the text prompt usually yields a completely different image with changed background and composition. However, by injecting the cross-attention maps from the original generation into the new diffusion process, they can replace centered object while keeping the background similar. In other words, the model is guided to reuse the layout and scene details from the first image and only alter the features tied to the changed word. This ability to perform localized prompt-based editing via attention editing is a powerful tool for generating coupled images that share background/content except for specific elements. Another notable advantage is that these methods can operate without explicit masks: the “mask” is effectively created by the attention focusing on the tokens of interest \cite{mokady2023null}.

\section{Additional Experiments}
In this section, we present additional experiments and ablation studies to compare our method to existing methods. The results further validate the effectiveness of our method in both keeping a similar background and maintaining text-image alignment for the coupled image generation task.
\subsection{Additional Examples}

\begin{figure}[!htbp]
    \centering
    \begin{minipage}[b]{0.24\textwidth}
        \centering
        \textbf{Ours}
        \vspace{0.2cm}
    \end{minipage}
    \begin{minipage}[b]{0.24\textwidth}
        \centering
        \textbf{Random seed}
               \vspace{0.2cm}
    \end{minipage}
    \begin{minipage}[b]{0.24\textwidth}
        \centering
        \textbf{P2P}
               \vspace{0.2cm}
    \end{minipage}
    \begin{minipage}[b]{0.24\textwidth}
        \centering
        \textbf{RF-inversion}
               \vspace{0.2cm}
    \end{minipage}
    \begin{minipage}[b]{0.24\textwidth}
        \centering
        \subfigure{
            \includegraphics[width=\textwidth]{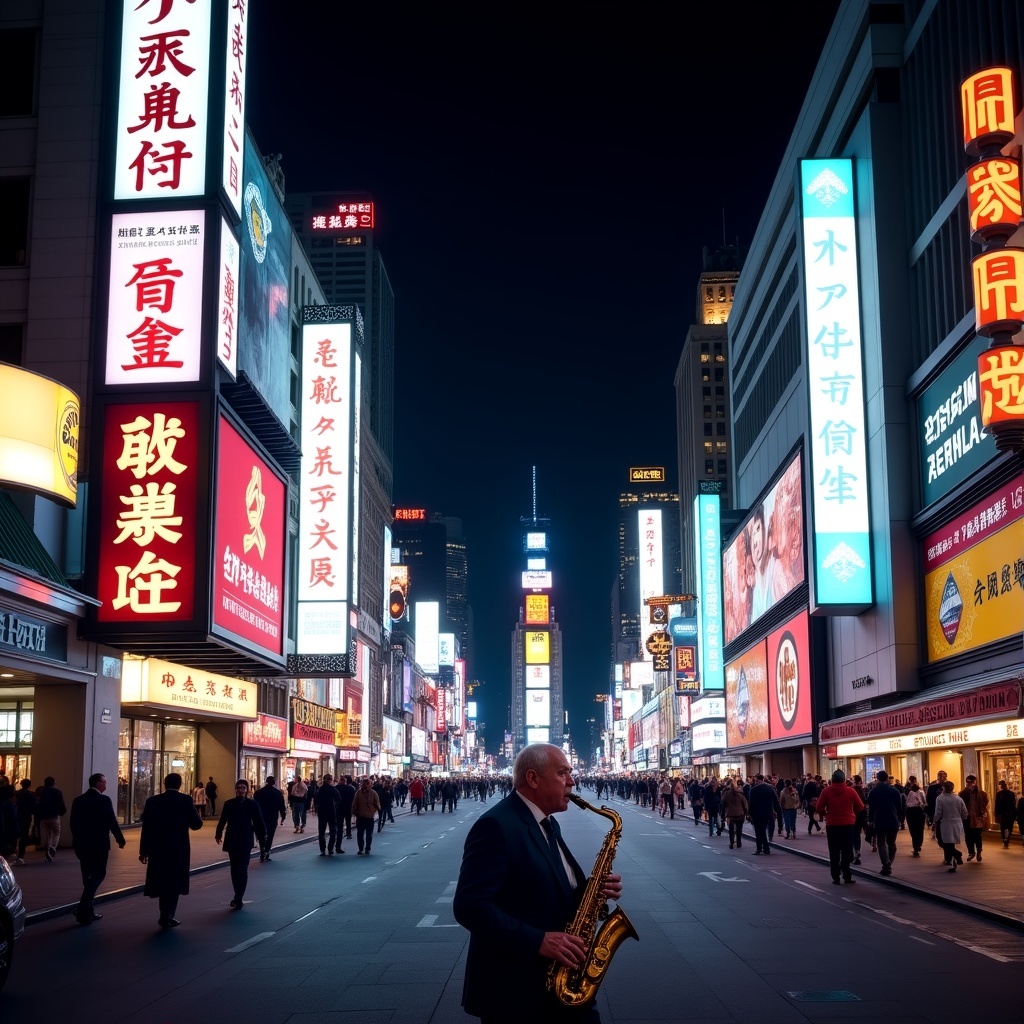}
        }
    \end{minipage}
    \begin{minipage}[b]{0.24\textwidth}
        \centering
        \subfigure{
            \includegraphics[width=\textwidth]{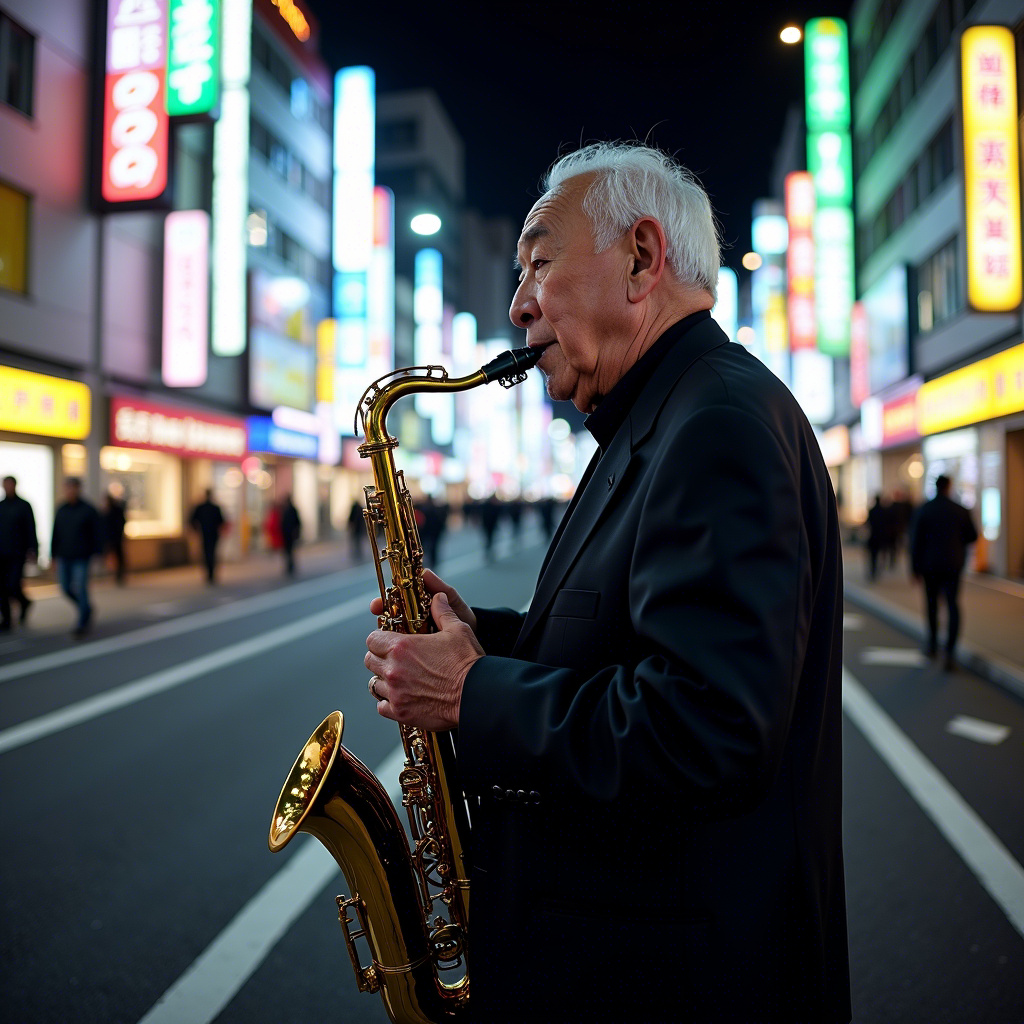}
        }
    \end{minipage}
    \begin{minipage}[b]{0.24\textwidth}
        \centering
        \subfigure{
            \includegraphics[width=\textwidth]{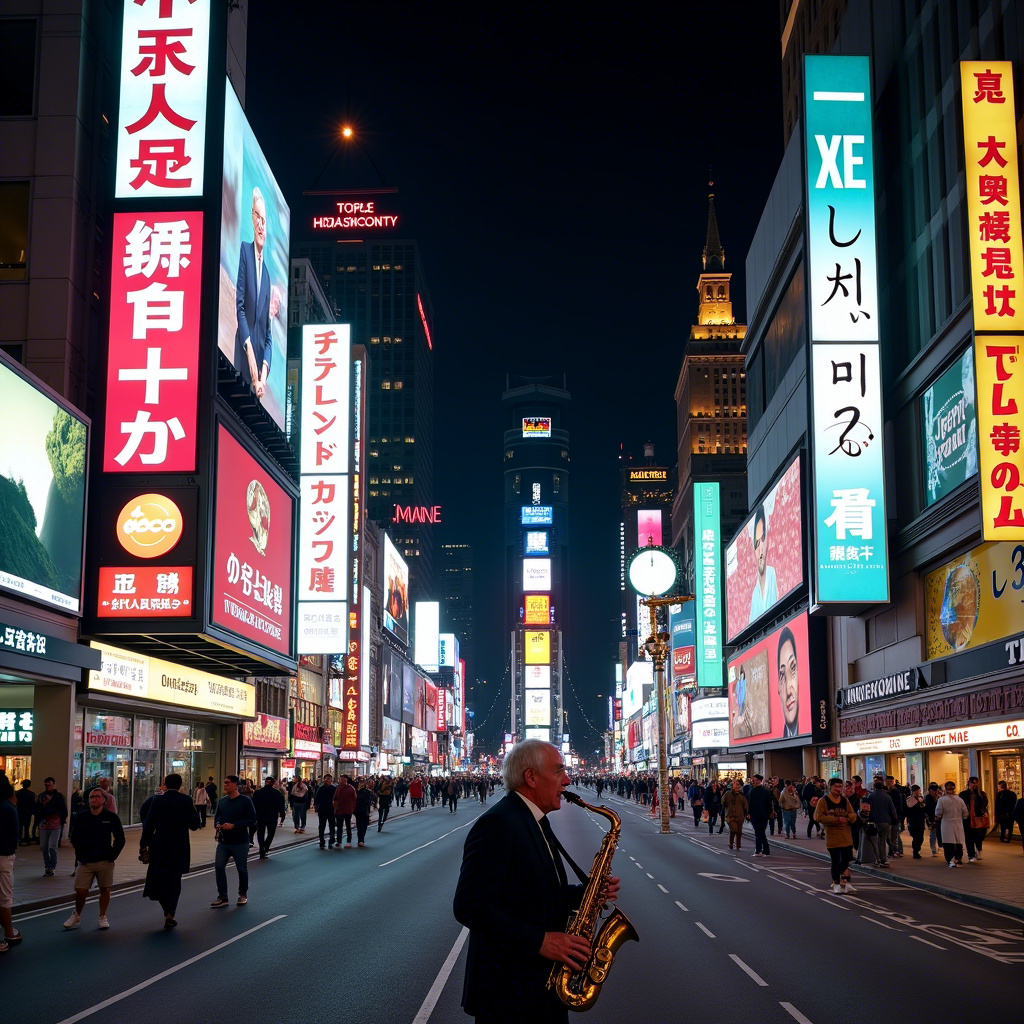}
        }
    \end{minipage}
    \begin{minipage}[b]{0.24\textwidth}
        \centering
        \subfigure{
            \includegraphics[width=\textwidth]{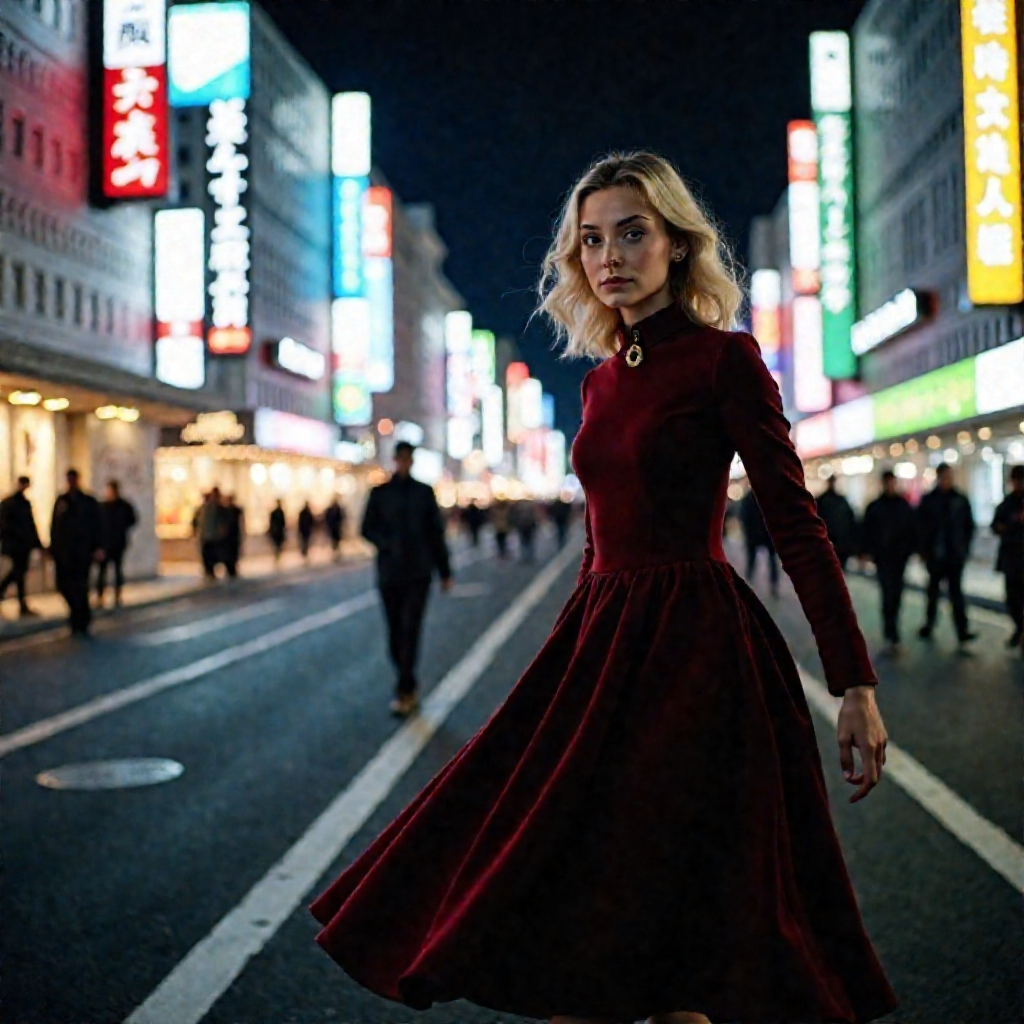}
        }
    \end{minipage}
    \begin{minipage}[b]{0.24\textwidth}
        \centering
        \subfigure{
            \includegraphics[width=\textwidth]{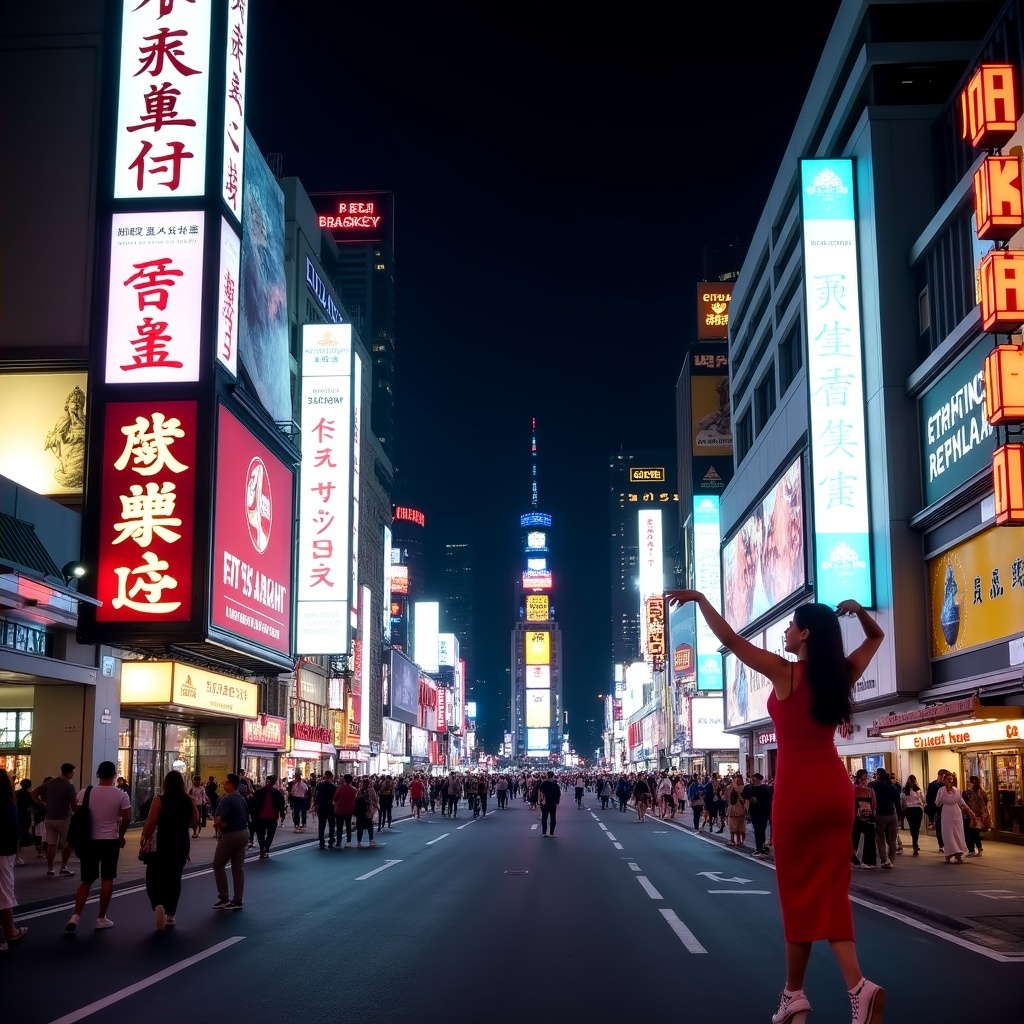}
        }
    \end{minipage}
    \begin{minipage}[b]{0.24\textwidth}
        \centering
        \subfigure{
            \includegraphics[width=\textwidth]{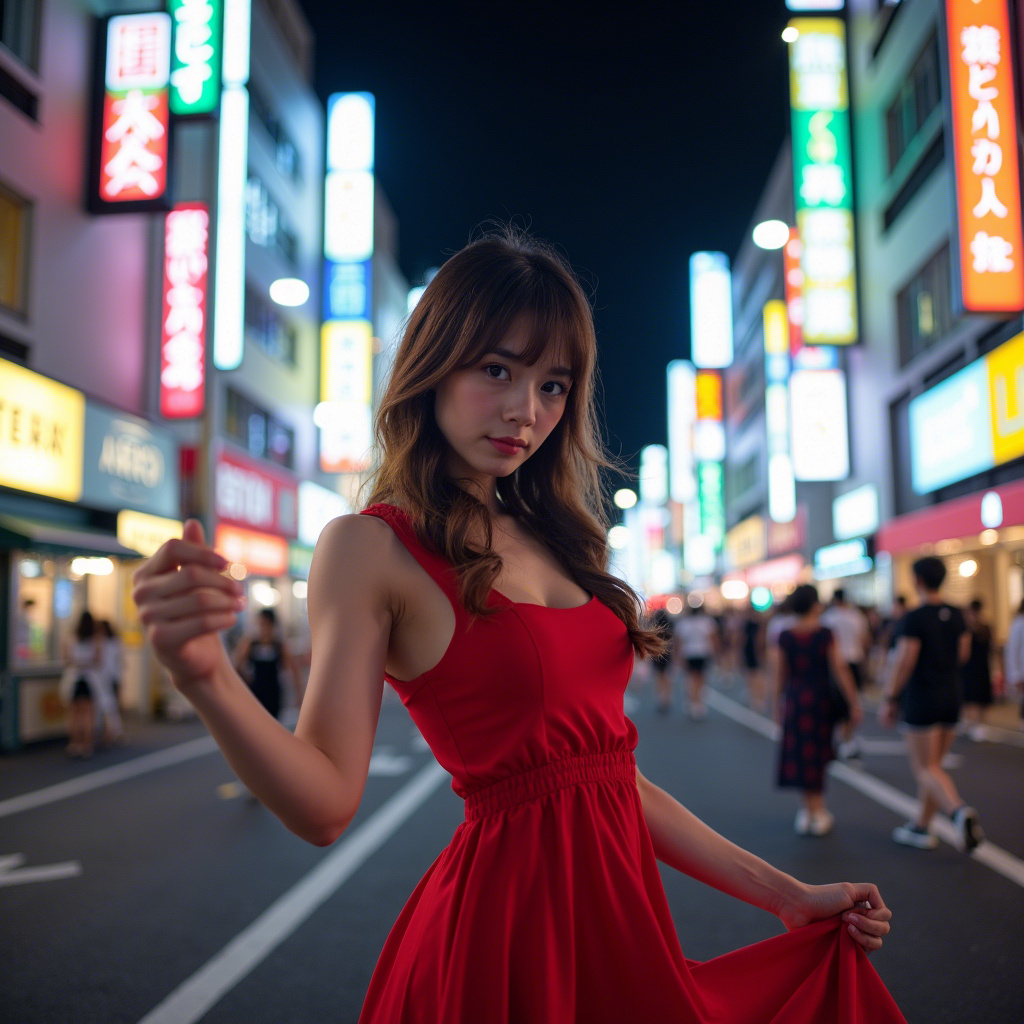}
        }
    \end{minipage}
    \begin{minipage}[b]{0.24\textwidth}
        \centering
        \subfigure{
            \includegraphics[width=\textwidth]{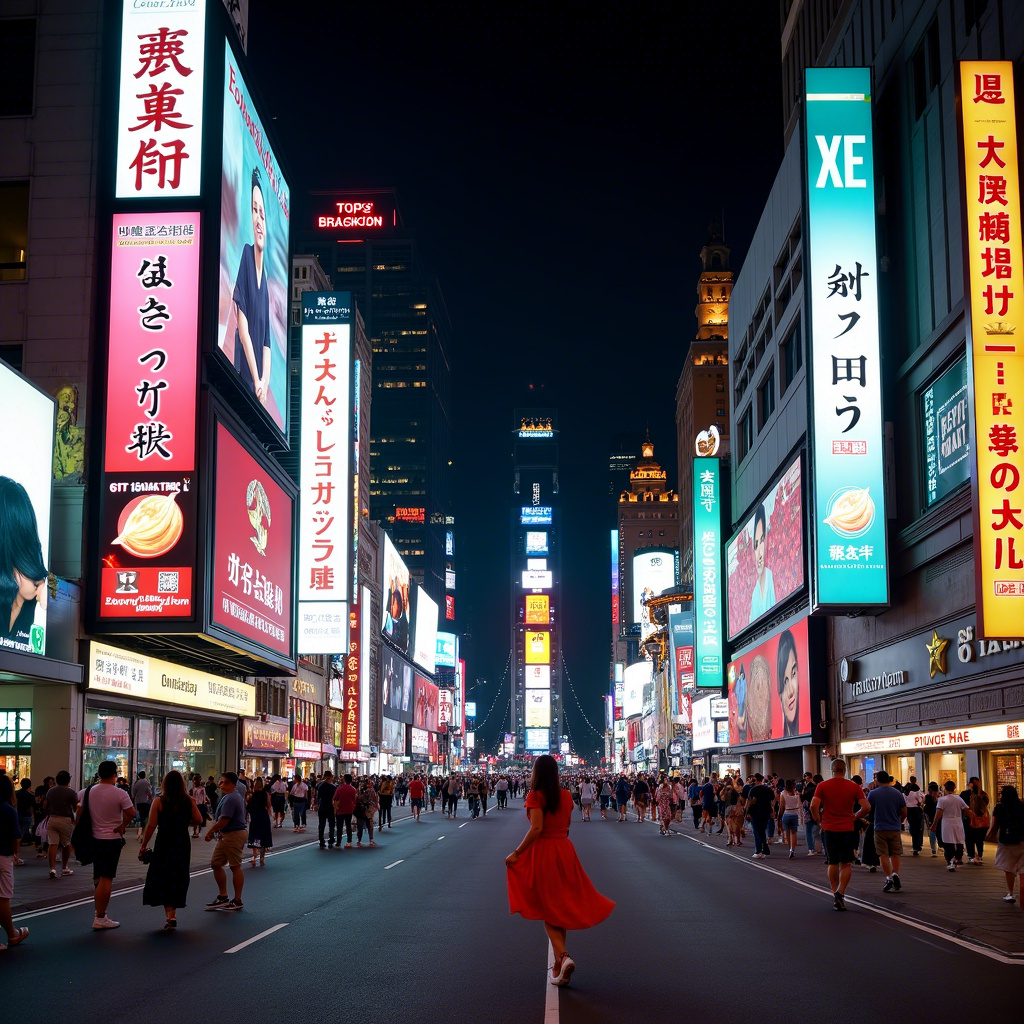}
        }
    \end{minipage}
    \begin{minipage}[b]{0.24\textwidth}
        \centering
        \subfigure{
            \includegraphics[width=\textwidth]{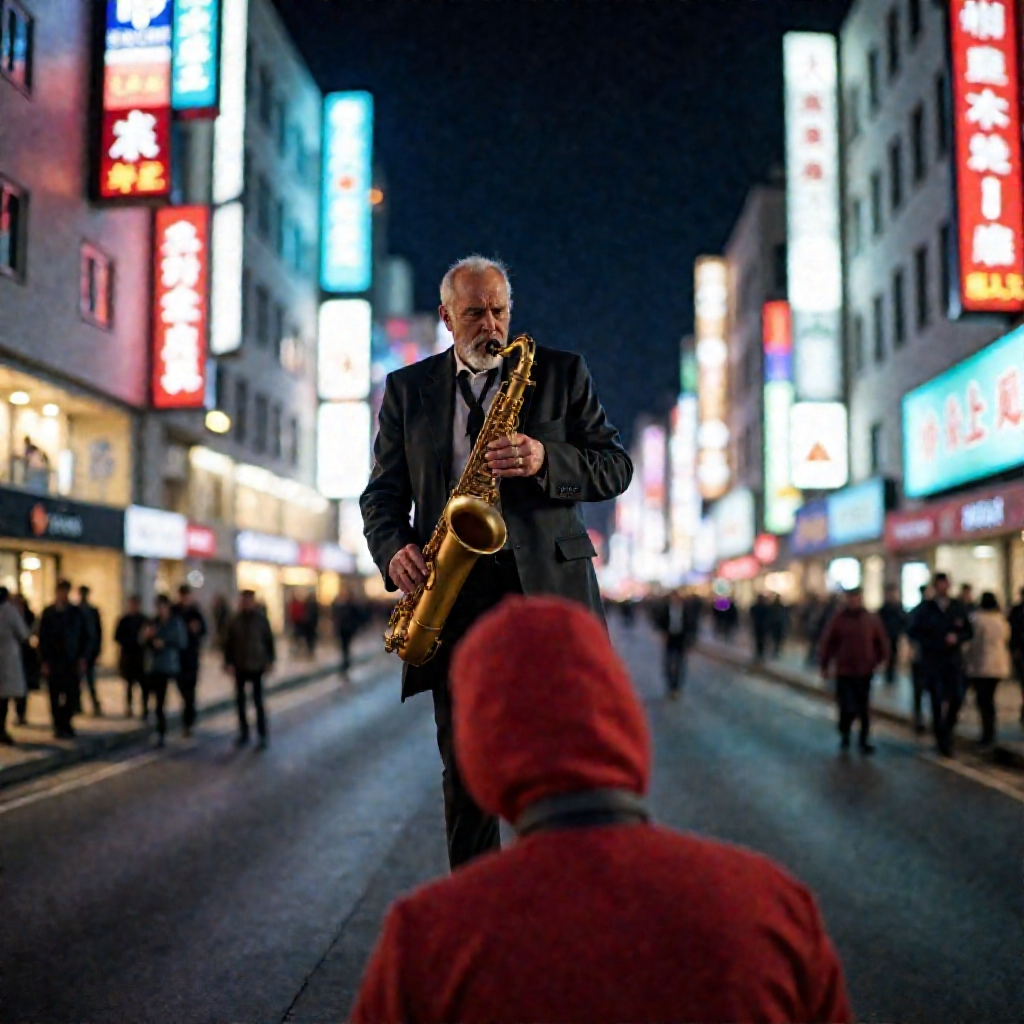}
        }
    \end{minipage}
    \caption{Comparison of our method to other baselines. Generated pictures in the first row correspond to prompt: \textit{"An elderly man in a suit playing saxophone on Tokyo night street, adorned with neon lights and billboards."}. The second row corresponds to prompt: \textit{"A young woman in a red dress dancing on Tokyo night street, adorned with neon lights and billboards."} Our method performs better than baselines in both background similarity and text-image alignment, where our method and all baseline methods share the same sampling procedures and random numbers.}
    \label{1}
\end{figure}

\begin{table}[!htbp]
\centering
\caption{Metrics for coupled image generation in Figure \ref{1}}
\begin{tabular}{ccccccc}\toprule
\label{metrics;2}
Metrics \textbackslash Model. &  Ours & Random seed & P2P & RF-inversion \\\midrule
Background similarity ($\times 10^{-4}$) ($\uparrow$) &
-8.663 & -38.62 & -11.43 & -50.5 \\
Text-image alignment ($\uparrow$) & 
17.31 & 16.72 & 15.36 & 11.21 \\
Combined metric ($\uparrow$) & 
1.063 & -0.3458 & 0.8636 & -0.7018
\\\bottomrule
\end{tabular}
\end{table}

\begin{figure}[!htbp]
    \centering
    \begin{minipage}[b]{0.24\textwidth}
        \centering
        \textbf{Ours}
        \vspace{0.2cm}
    \end{minipage}
    \begin{minipage}[b]{0.24\textwidth}
        \centering
        \textbf{Random seed}
               \vspace{0.2cm}
    \end{minipage}
    \begin{minipage}[b]{0.24\textwidth}
        \centering
        \textbf{P2P}
               \vspace{0.2cm}
    \end{minipage}
    \begin{minipage}[b]{0.24\textwidth}
        \centering
        \textbf{RF-inversion}
               \vspace{0.2cm}
    \end{minipage}
    \begin{minipage}[b]{0.24\textwidth}
        \centering
        \subfigure{
            \includegraphics[width=\textwidth]{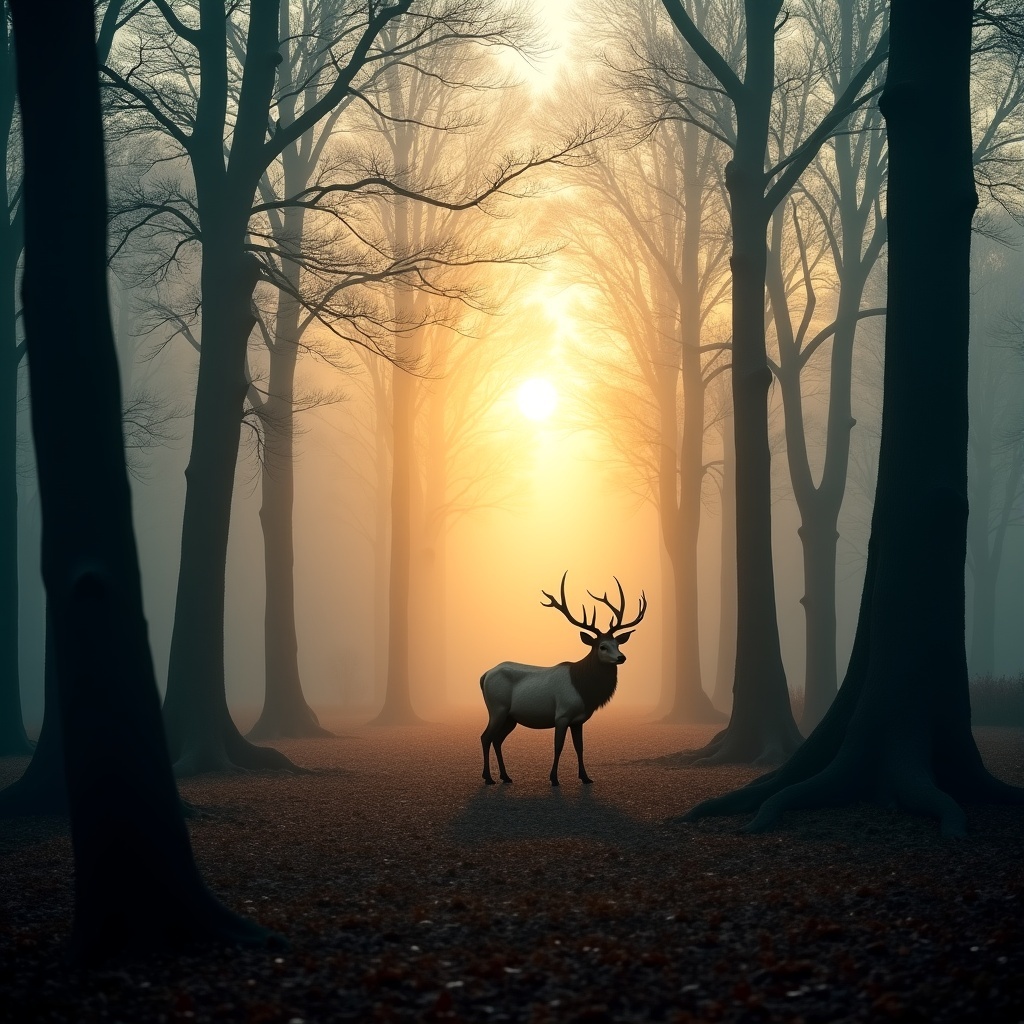}
        }
    \end{minipage}
    \begin{minipage}[b]{0.24\textwidth}
        \centering
        \subfigure{
            \includegraphics[width=\textwidth]{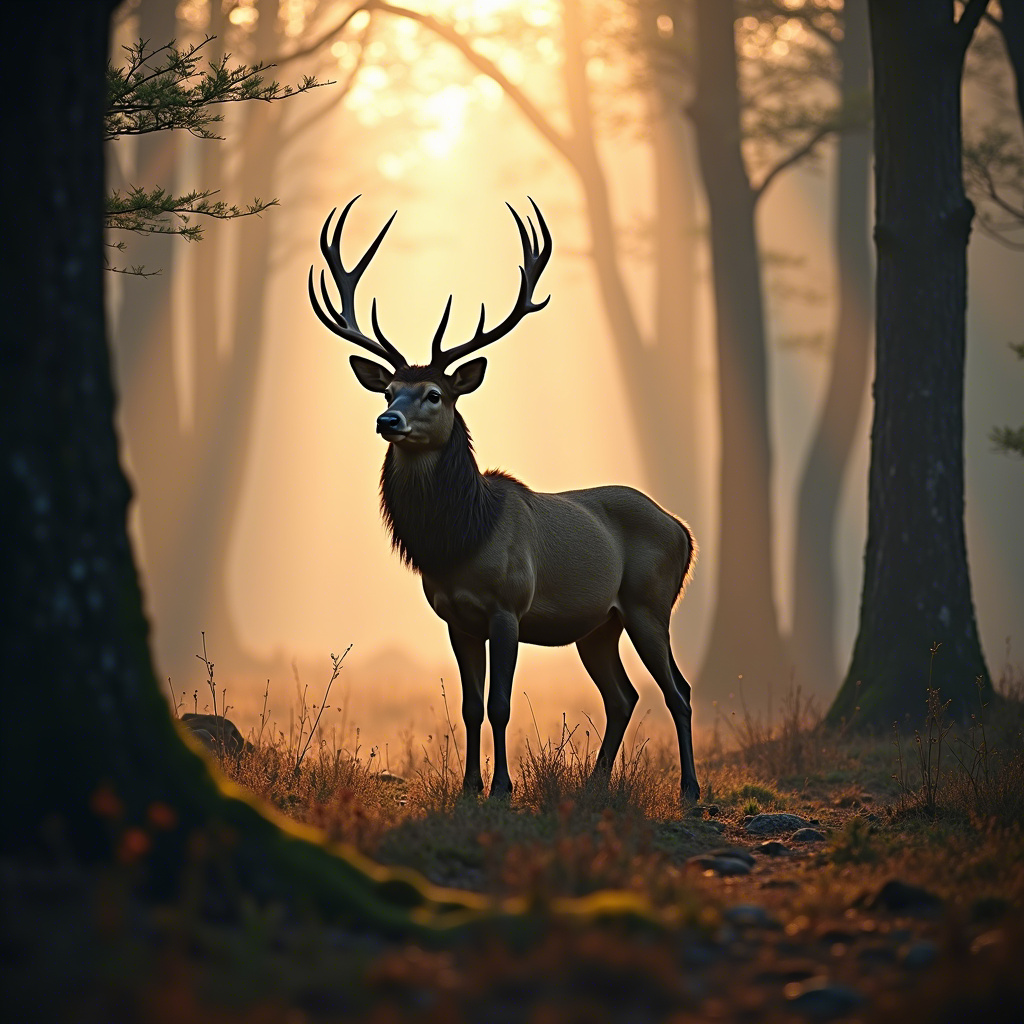}
        }
    \end{minipage}
    \begin{minipage}[b]{0.24\textwidth}
        \centering
        \subfigure{
            \includegraphics[width=\textwidth]{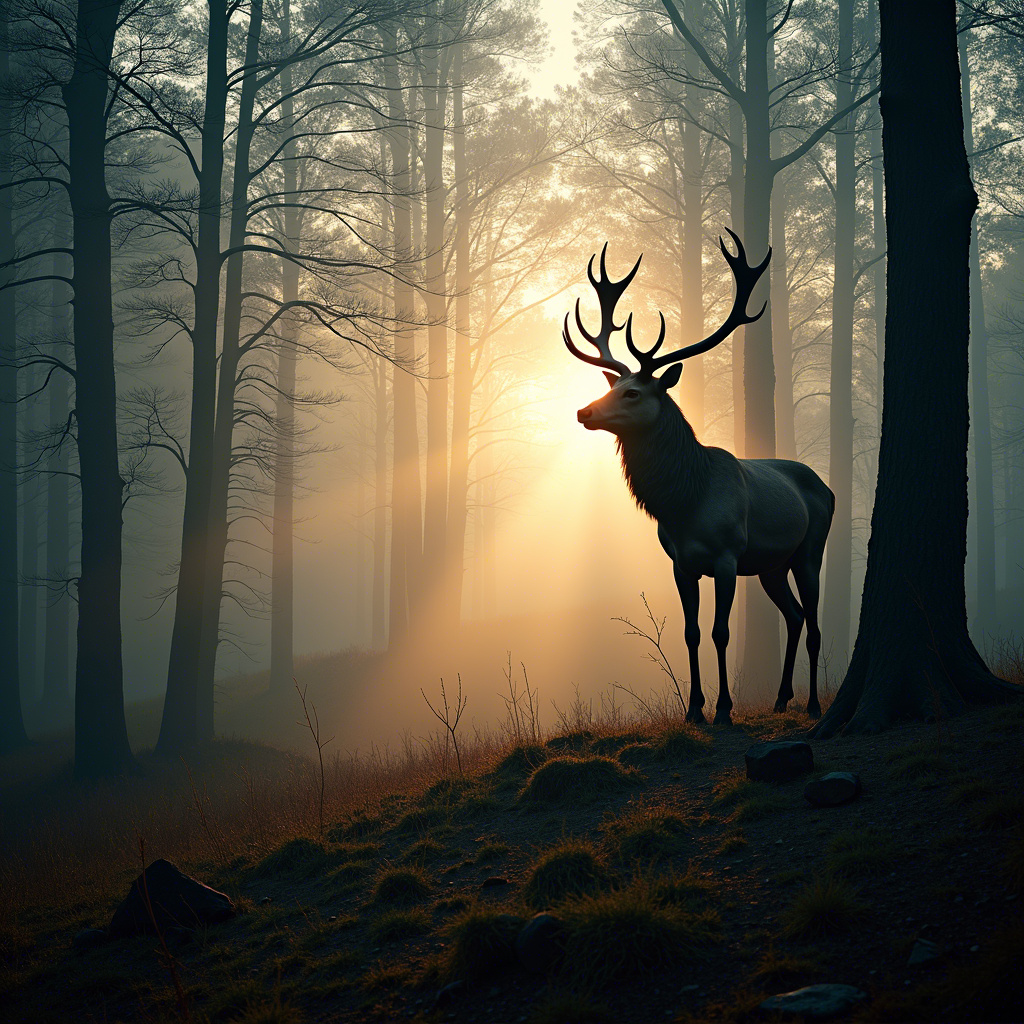}
        }
    \end{minipage}
    \begin{minipage}[b]{0.24\textwidth}
        \centering
        \subfigure{
            \includegraphics[width=\textwidth]{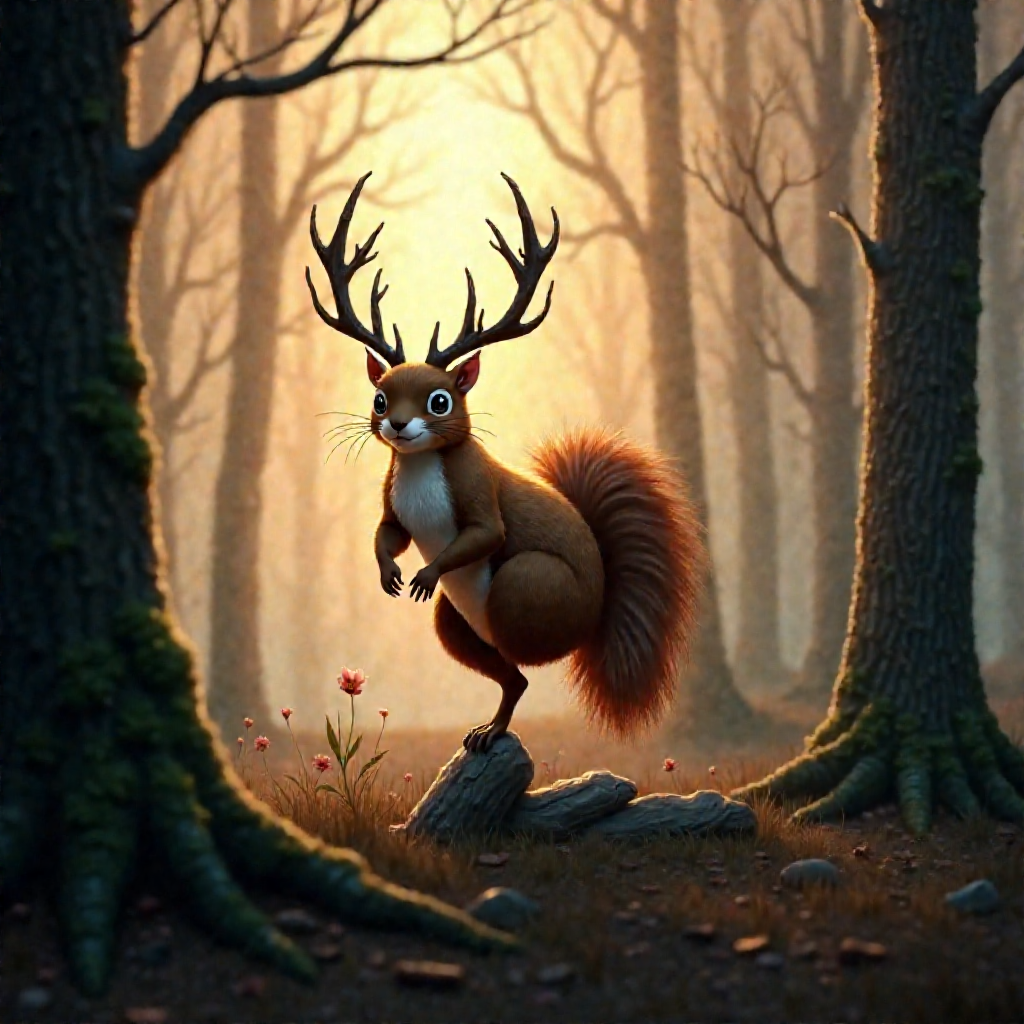}
        }
    \end{minipage}
    \begin{minipage}[b]{0.24\textwidth}
        \centering
        \subfigure{
            \includegraphics[width=\textwidth]{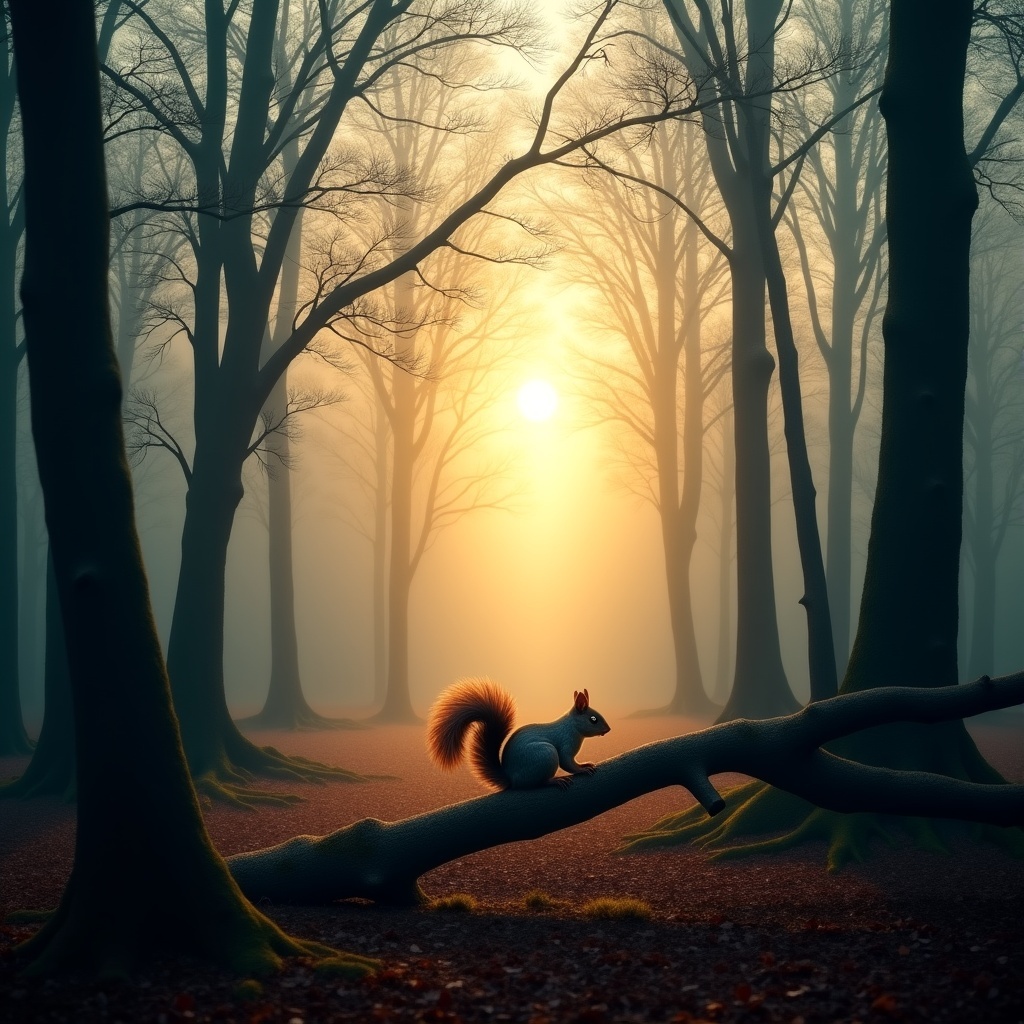}
        }
    \end{minipage}
    \begin{minipage}[b]{0.24\textwidth}
        \centering
        \subfigure{
            \includegraphics[width=\textwidth]{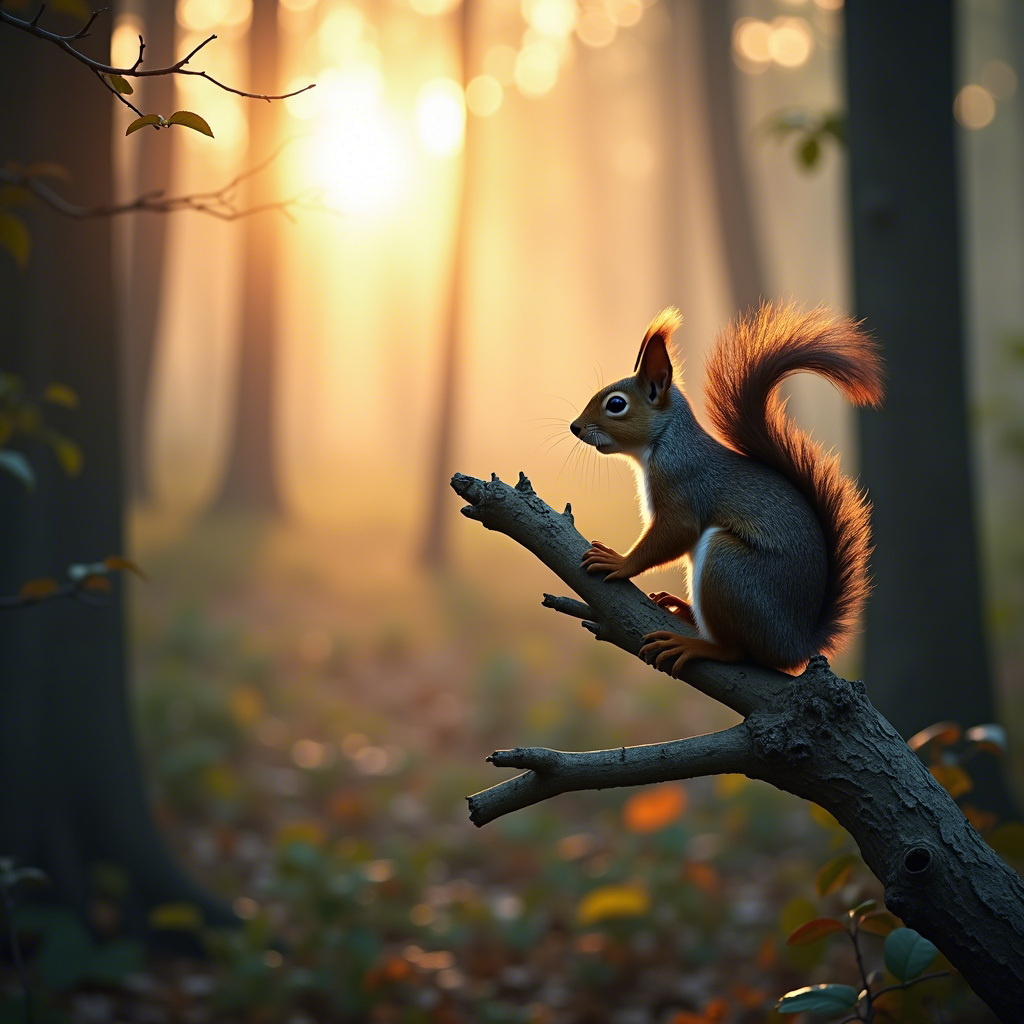}
        }
    \end{minipage}
    \begin{minipage}[b]{0.24\textwidth}
        \centering
        \subfigure{
            \includegraphics[width=\textwidth]{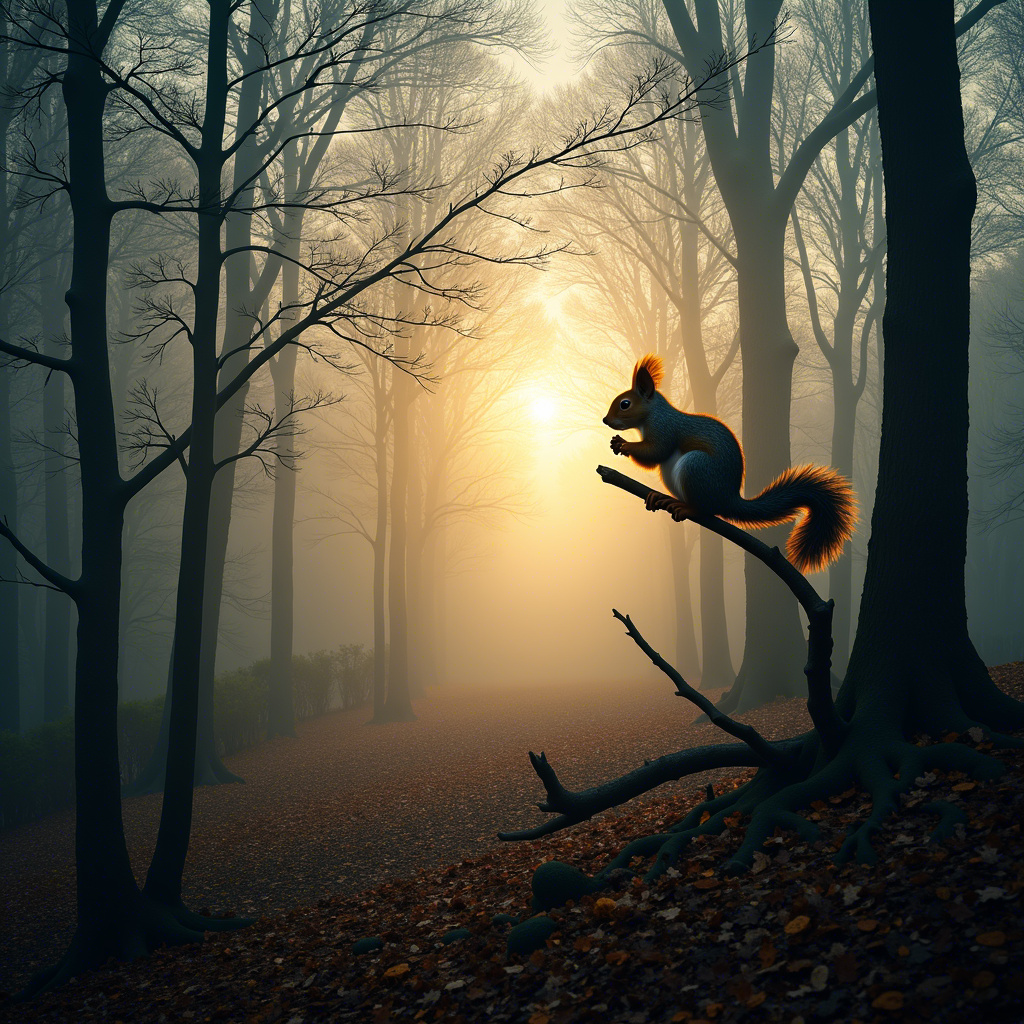}
        }
    \end{minipage}
    \begin{minipage}[b]{0.24\textwidth}
        \centering
        \subfigure{
            \includegraphics[width=\textwidth]{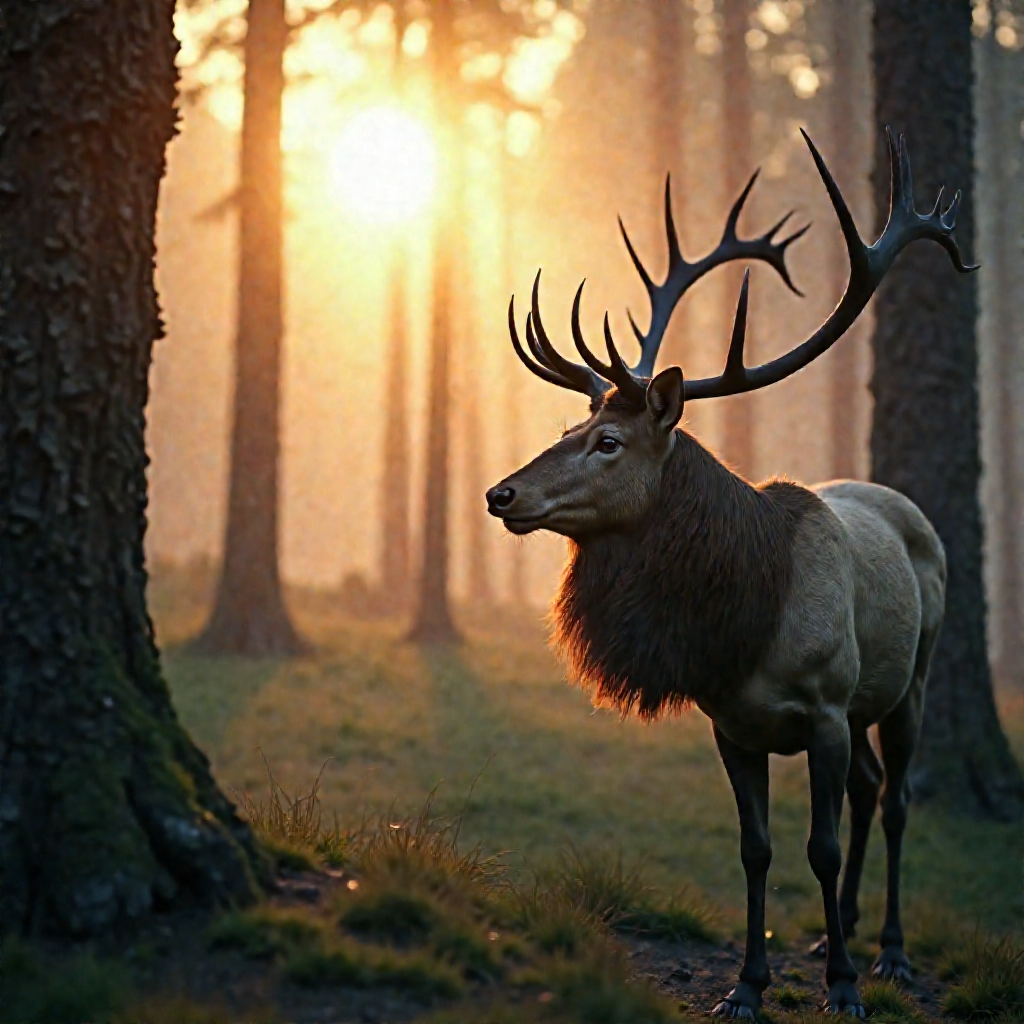}
        }
    \end{minipage}
    \caption{Comparison of our method to other baselines. Generated pictures in the first row correspond to prompt: \textit{"A majestic stag standing proud in a misty forest at dawn, sunlight through trees."}. The second row corresponds to prompt: \textit{"A squirrel perched on a branch in a misty forest at dawn, sunlight through trees."} Our method performs better than baselines in both background similarity and text-image alignment, where our method and all baseline methods share the same sampling procedures and random numbers.}
    \label{2}
\end{figure}

\begin{table}[!htbp]
\centering
\caption{Metrics for coupled image generation in Figure \ref{2}}
\begin{tabular}{ccccccc}\toprule
\label{metrics;3}
Metrics \textbackslash Model. &  Ours & Random seed & P2P & RF-inversion \\\midrule
Background similarity ($\times 10^{-4}$) ($\uparrow$) & 
-4.953 & -12.13 & -5.585 & -30.79 \\
Text-image alignment ($\uparrow$) & 
21.72 & 20.47 & 19.81 & 16.59 \\
Combined metric ($\uparrow$) & 
1.413 & 1.099 & 1.296 & 0.253
\\\bottomrule
\end{tabular}
\end{table}

\newpage
\subsection{Visualization of the Time-varying Parameters} 
In this section, we directly visualize the time-varying parameters after solving the isotonic optimization problem, as proposed in Section 4.3. We observe that these time-varying parameters follow a common pattern, starting near 0 and gradually approaching 1 over time. This aligns with the previous findings in \cite{choi2022perception,park2023understanding,yue2024exploring,wang2024towards} that diffusion models initially focus on generating coarse structure and subsequently refining fine-grained details. Indeed, the time-varying parameters are close to 0 at the initial stage of inference to ensure the generated images share a similar background. As timestep increases, the time-varying parameters also increase and approximate 1 to refine the entity features and maintain text-image alignment.

\begin{figure}[!htbp]
    \centering
    \begin{minipage}[b]{0.45\textwidth}
        \centering
        \textbf{Prompts}
        \vspace{0.4cm}
    \end{minipage}
    \begin{minipage}[b]{0.5\textwidth}
        \centering
        \textbf{Visualization of the parameters}
        \vspace{0.4cm}
    \end{minipage}
    \begin{minipage}[b]{0.45\textwidth}
        An eagle soars gracefully in a vibrant meadow under a clear blue sky, bathed in sunlight. \\
        A red fox runs in a vibrant meadow under a clear blue sky, bathed in sunlight.
        \vspace{0.4cm}
    \end{minipage}
    \begin{minipage}[b]{0.5\textwidth}
        \centering
        \subfigure{
            \includegraphics[width=\textwidth]{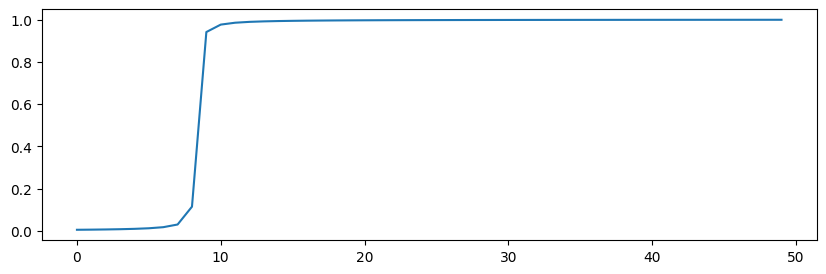}
        }
    \end{minipage}
    \begin{minipage}[b]{0.45\textwidth}
        An elderly man in a suit playing saxophone on Tokyo night street, adorned with neon lights and billboards. \\
        A young woman in a red dress dancing on Tokyo night street, adorned with neon lights and billboards.
        \vspace{0.4cm}
    \end{minipage}
    \begin{minipage}[b]{0.5\textwidth}
        \centering
        \subfigure{
            \includegraphics[width=\textwidth]{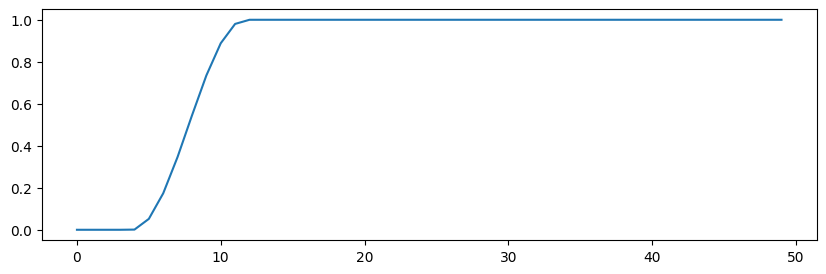}
        }
    \end{minipage}
    \begin{minipage}[b]{0.45\textwidth}
        A majestic stag standing proud in a misty forest at dawn, sunlight through trees. \\
        A squirrel perched on a branch in a misty forest at dawn, sunlight through trees.
        \vspace{0.4cm}
    \end{minipage}
    \begin{minipage}[b]{0.5\textwidth}
        \centering
        \subfigure{
            \includegraphics[width=\textwidth]{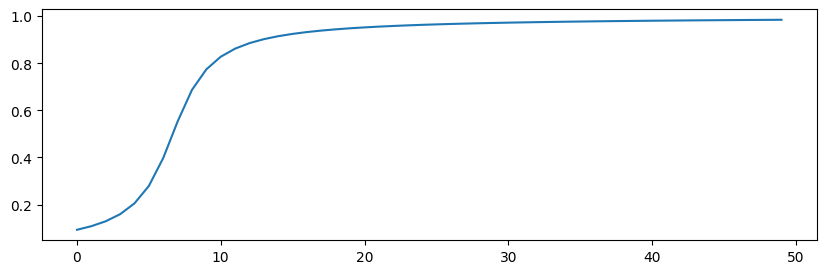}
        }
    \end{minipage}
    \caption{Visualization of the trained time-varying parameters used in the three examples.}
\end{figure}

\subsection{Ablation Studies}
\subsubsection{Ablation Studies on Parameterized Functions}
To further simplify the training load for optimizing the time-varying parameters, we propose to use three classes of parameterized functions to substitute the time-varying parameters. These functions reduce the computational efforts by restricting the number of independent variables.

\begin{itemize}
    \item \textbf{"01" Function:} This function defines $\boldsymbol{\theta}$ as a binary value. To the left of the 'center' point, $\boldsymbol{\theta}$ is 0, and to the right of the 'center' point, $\boldsymbol{\theta}$ is 1.
    Mathematically, if $t$ is the time variable and $c$ is the 'center' point:
    $$ \boldsymbol{\theta}(t) = \begin{cases} 0 & \text{if } t < c \\ 1 & \text{if } t \ge c \end{cases} $$
    
    \item \textbf{"arctan" Function:} This function utilizes a transformed arctangent curve. The standard $\arctan(x)$ function is scaled and shifted such that its output is $0.5$ at the 'center' point $c$. A common transformation to achieve $\boldsymbol{\theta}(c) = 0.5$ and scale the output to approximately $[0,1]$ could be:
    $$ \boldsymbol{\theta}(t) = \frac{1}{\pi} \arctan(k(t-c)) + 0.5 $$
    where $k$ is a scaling factor controlling the steepness of the transition.

    \item \textbf{"sin" Function:} This function uses a transformed sine wave. The sine function is shifted so that its value is $0.5$ at the 'center' point $c$. Furthermore, the function is clamped: for input values that would map to an angle of $-\pi/2$ or less in the sine function's argument (after transformation), $\boldsymbol{\theta}$ is set to 0. For input values that would map to an angle of $\pi/2$ or more, $\boldsymbol{\theta}$ is set to 1. This effectively uses a portion of the sine wave to transition from 0 to 1.
    The function can be defined as:
    $$ \boldsymbol{\theta}(t) = \begin{cases} 0 & \text{if } t \le t_{-\pi/2} \\ \frac{1}{2}\sin\left(k(t-c)\right) + 0.5 & \text{if } t_{-\pi/2} < t < t_{\pi/2} \\ 1 & \text{if } t \ge t_{\pi/2} \end{cases} $$
    Here, $c$ is the 'center' where $k(t-c)=0$, resulting in $\boldsymbol{\theta}(c) = \frac{1}{2}\sin(0) + 0.5 = 0.5$. The points $t_{-\pi/2}$ and $t_{\pi/2}$ are defined such that $k(t_{-\pi/2}-c) = -\pi/2$ (where $\boldsymbol{\theta}$ transitions to 0) and $k(t_{\pi/2}-c) = \pi/2$ (where $\boldsymbol{\theta}$ transitions to 1). The parameter $k$ controls the width of the transition.

\end{itemize}

\begin{figure}[!htbp]
    \centering
    \begin{minipage}[b]{0.3\textwidth}
        \centering
        \textbf{01}
        \vspace{0.2cm}
    \end{minipage}
    \begin{minipage}[b]{0.3\textwidth}
        \centering
        \textbf{arctan}
               \vspace{0.2cm}
    \end{minipage}
    \begin{minipage}[b]{0.3\textwidth}
        \centering
        \textbf{sin}
               \vspace{0.2cm}
    \end{minipage}
    \begin{minipage}[b]{0.3\textwidth}
        \centering
        \subfigure{
            \includegraphics[width=\textwidth]{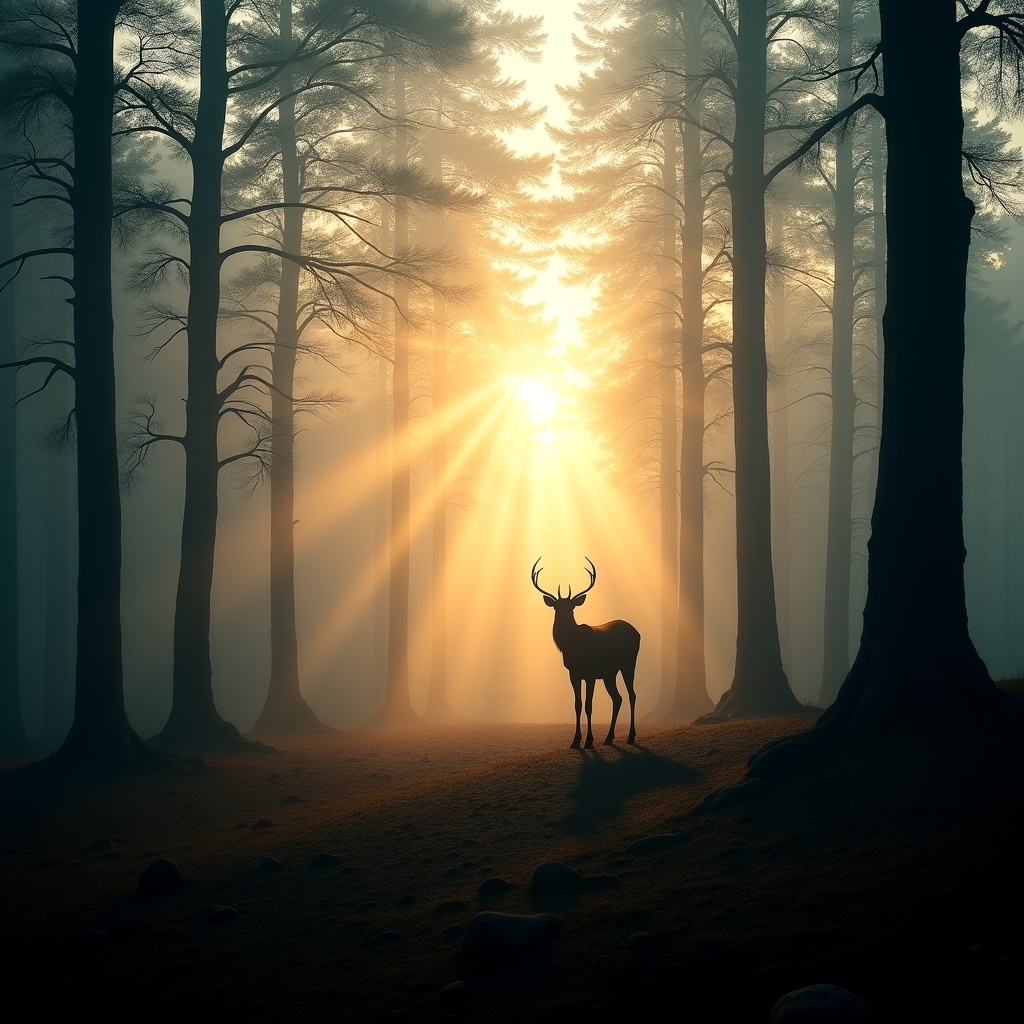}
        }
    \end{minipage}
    \begin{minipage}[b]{0.3\textwidth}
        \centering
        \subfigure{
            \includegraphics[width=\textwidth]{figure/prompt3/ours1.jpg}
        }
    \end{minipage}
    \begin{minipage}[b]{0.3\textwidth}
        \centering
        \subfigure{
            \includegraphics[width=\textwidth]{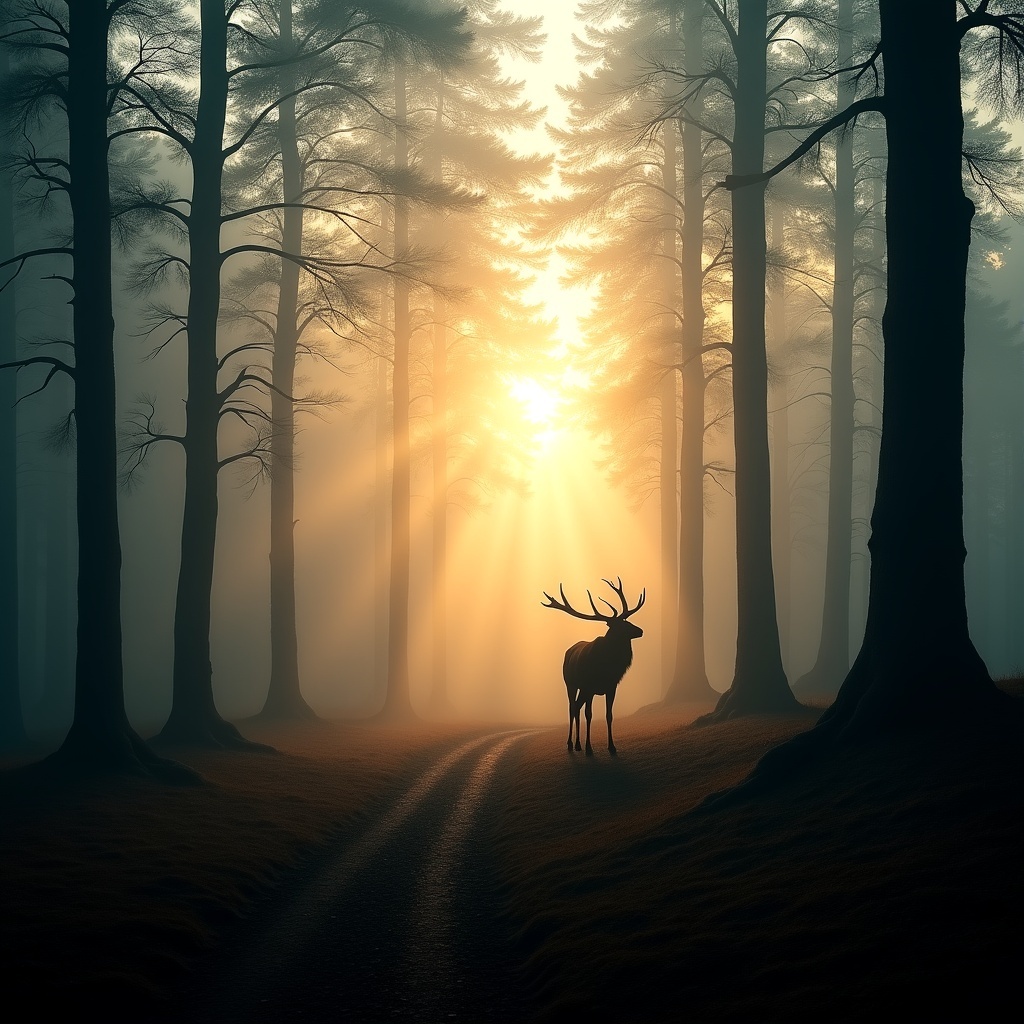}
        }
    \end{minipage}
    \begin{minipage}[b]{0.3\textwidth}
        \centering
        \subfigure{
            \includegraphics[width=\textwidth]{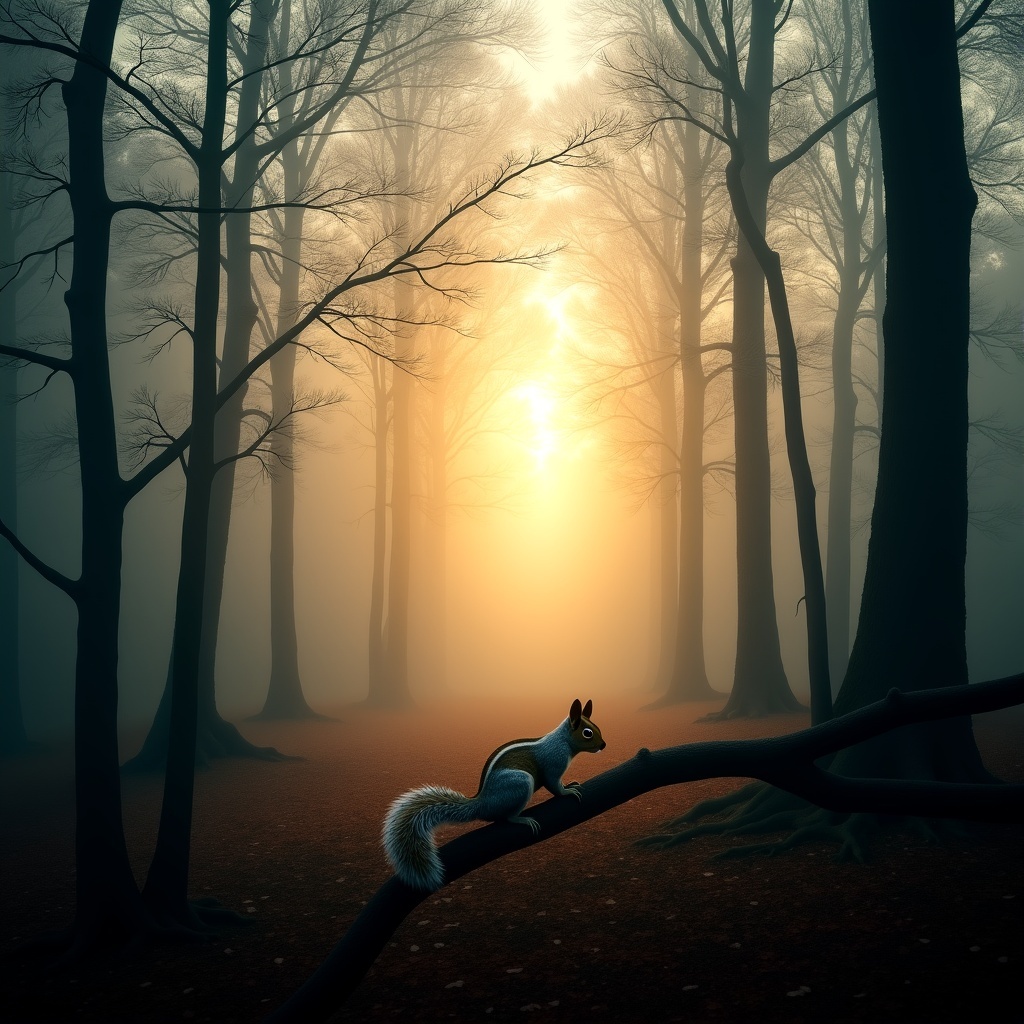}
        }
        \vspace{-0.5cm}
        \caption*{\centering center=10}
    \end{minipage}
    \begin{minipage}[b]{0.3\textwidth}
        \centering
        \subfigure{
            \includegraphics[width=\textwidth]{figure/prompt3/ours2.jpg}
        }
        \vspace{-0.5cm}
        \caption*{\centering center=6.7, scale=0.5}
    \end{minipage}
    \begin{minipage}[b]{0.3\textwidth}
        \centering
        \subfigure{
            \includegraphics[width=\textwidth]{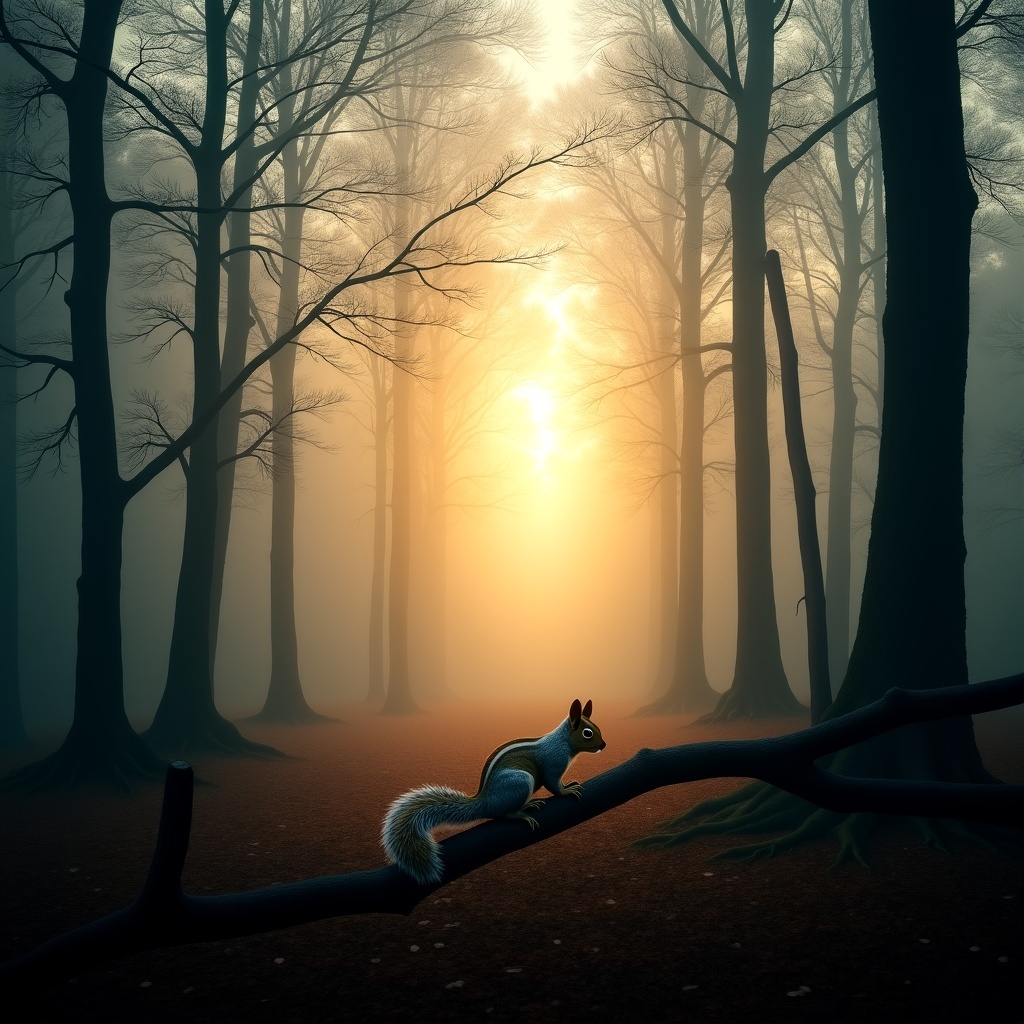}
        }
        \vspace{-0.5cm}
        \caption*{\centering center=10, scale=0.8}
    \end{minipage}
    \caption{Comparison of different parameterized functions. The prompts are \textit{"A majestic stag standing proud in a misty forest at dawn, sunlight through trees."} and \textit{"A squirrel perched on a branch in a misty forest at dawn, sunlight through trees."}}
    \label{3}
\end{figure}

\begin{table}[htbp]
\centering
\caption{Metrics for coupled image generation in Figure \ref{3}}
\begin{tabular}{ccccccc}\toprule
\label{metrics;4}
Metrics \textbackslash Model. &  01 & arctan & sin \\\midrule
Background similarity ($\times 10^{-4}$) ($\uparrow$) & 
-4.55 & -4.953 & -4.263 \\
Text-image alignment ($\uparrow$) & 
19.05 & 21.72 & 22.22 \\
Combined metric ($\uparrow$) & 
1.293 & 1.413 & 1.462
\\\bottomrule
\end{tabular}
\end{table}

The results indicate that the arctan and sin functions slightly outperform the binary 0-1 function. We attribute this improvement to the inherent smoothness of the arctan and sin functions. Unlike the 0-1 function, which introduces abrupt transitions, the gradual transitions provided by arctan and sin functions offer greater flexibility and smoother progression from 0 to 1.

\newpage
\subsubsection{Ablation Studies on Center and Scale of the Parameterized Functions}
\begin{figure}[!htbp]
    \centering
    \begin{minipage}[b]{0.18\textwidth}
        \centering
        \subfigure{
            \includegraphics[width=\textwidth]{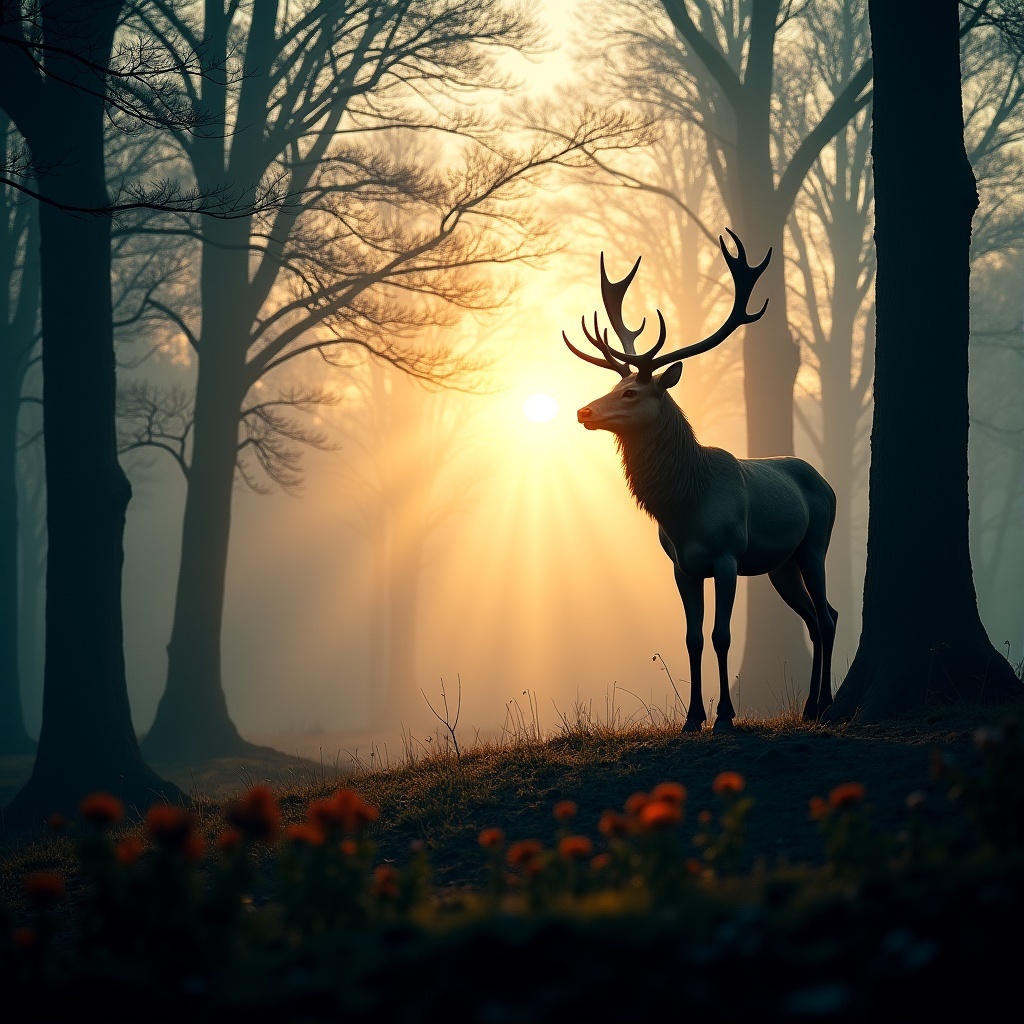}
        }
    \end{minipage}
    \begin{minipage}[b]{0.18\textwidth}
        \centering
        \subfigure{
            \includegraphics[width=\textwidth]{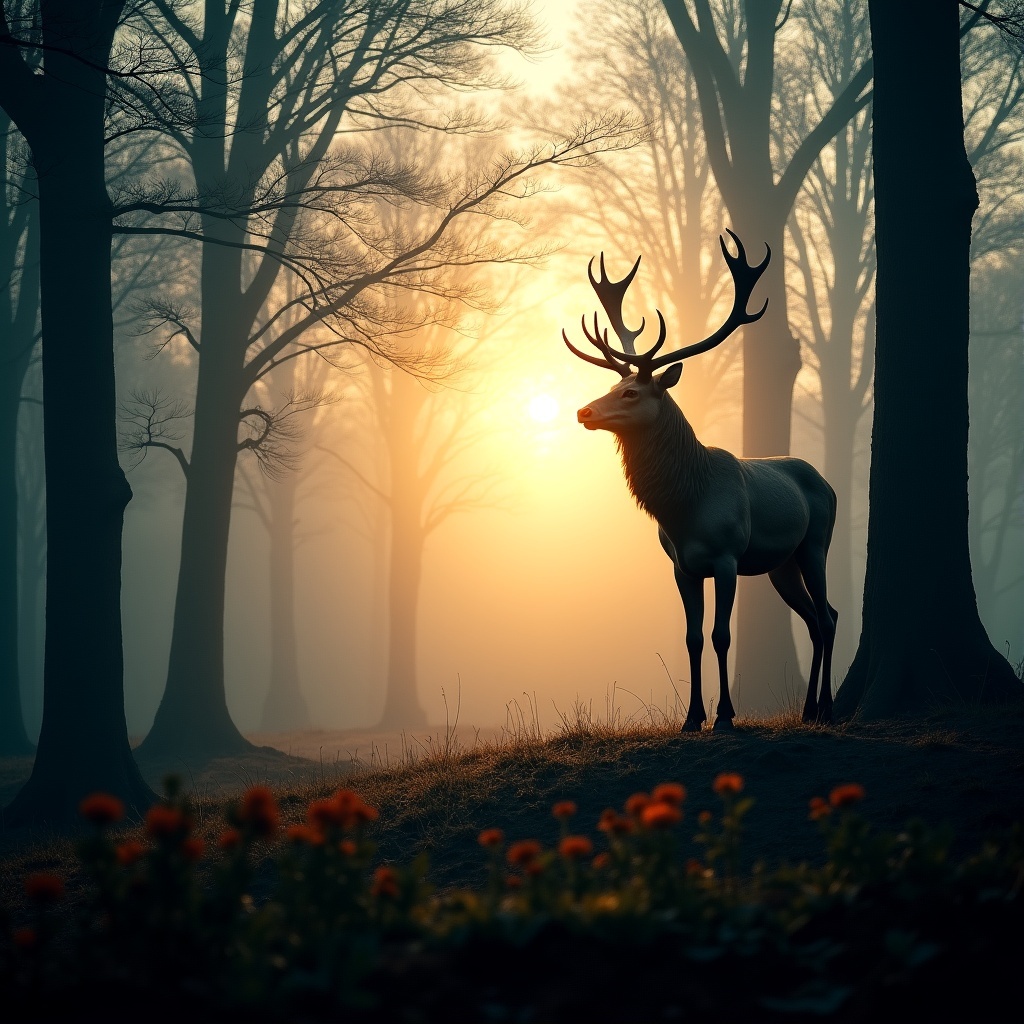}
        }
    \end{minipage}
    \begin{minipage}[b]{0.18\textwidth}
        \centering
        \subfigure{
            \includegraphics[width=\textwidth]{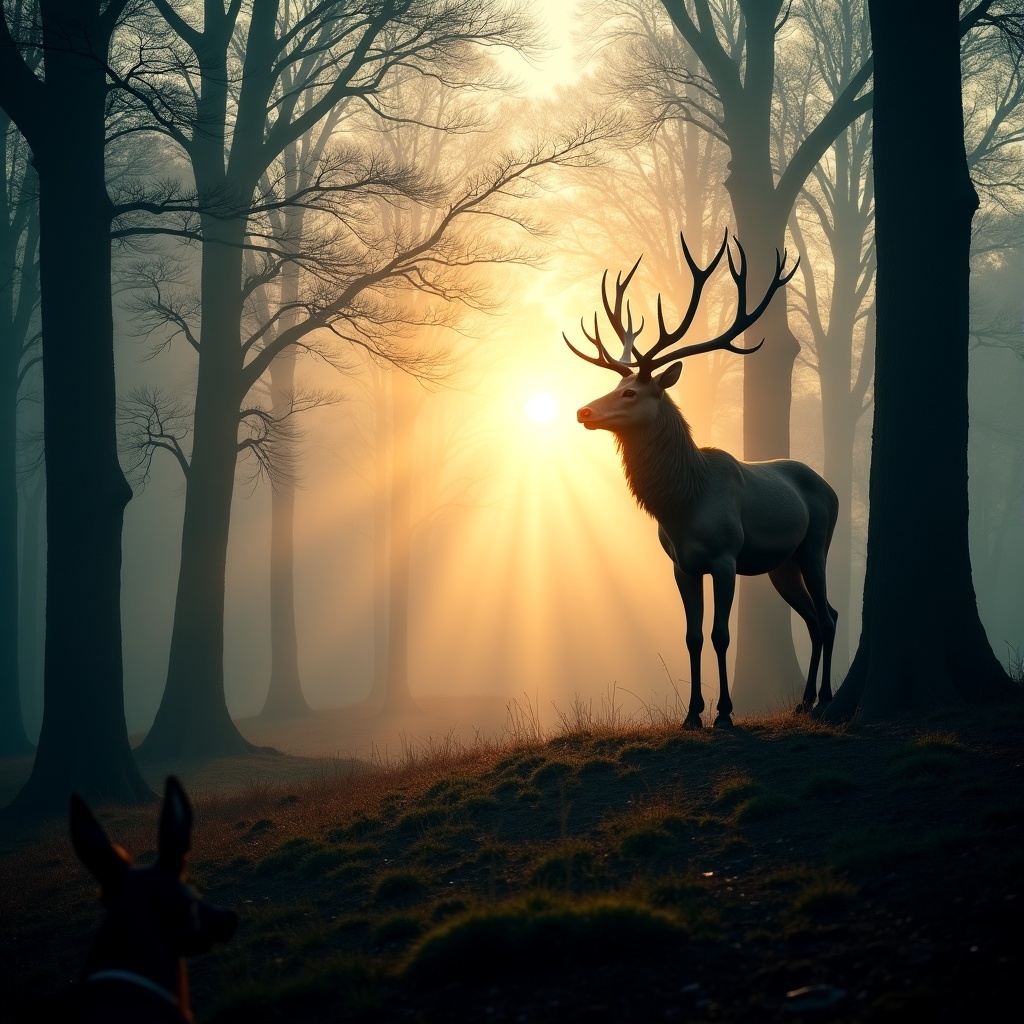}
        }
    \end{minipage}
    \begin{minipage}[b]{0.18\textwidth}
        \centering
        \subfigure{
            \includegraphics[width=\textwidth]{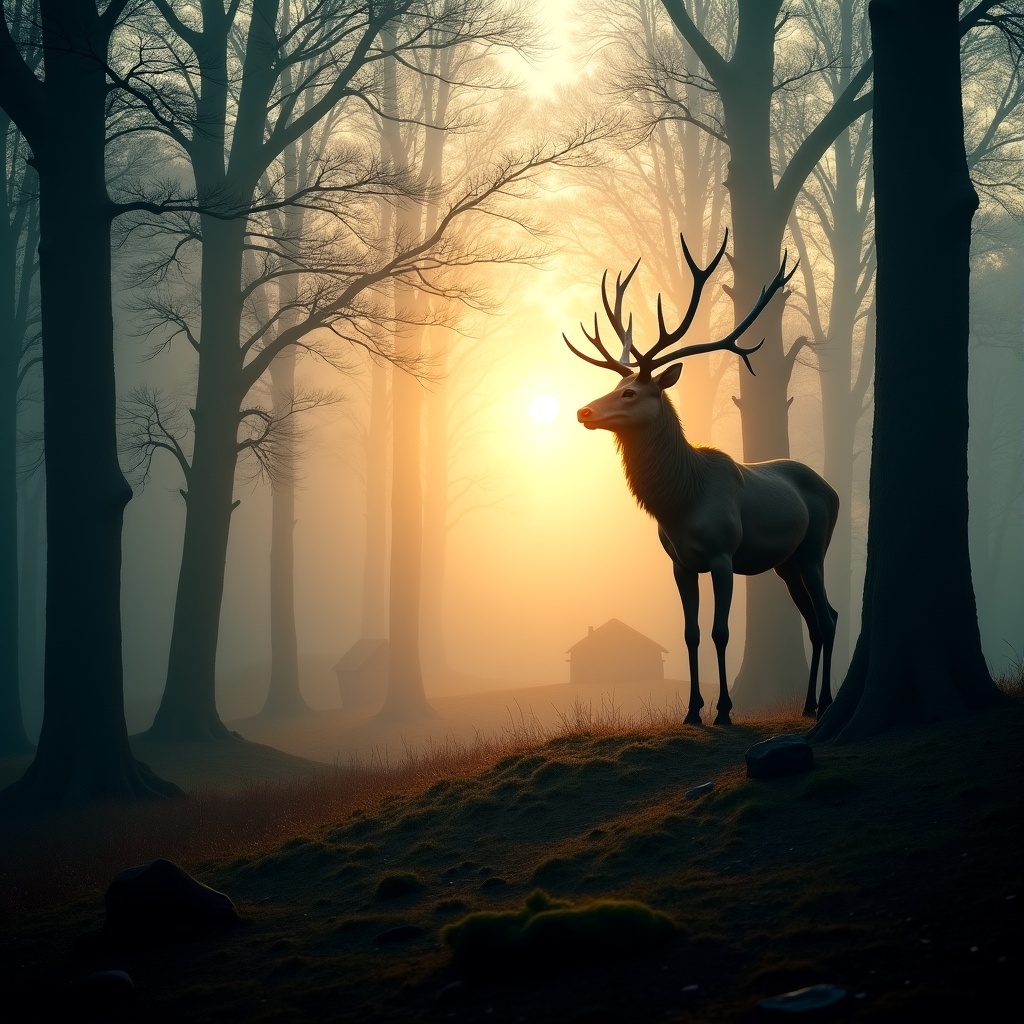}
        }
    \end{minipage}
    \begin{minipage}[b]{0.18\textwidth}
        \centering
        \subfigure{
            \includegraphics[width=\textwidth]{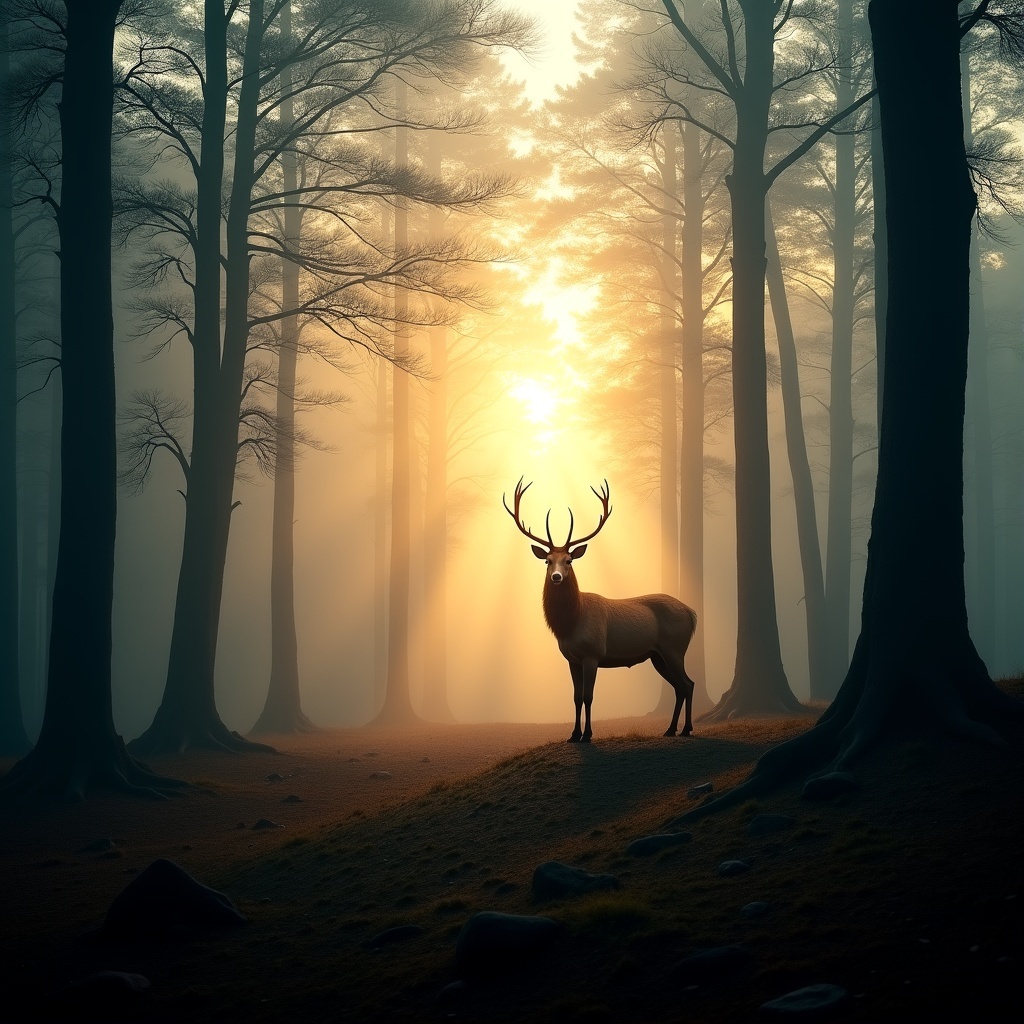}
        }
    \end{minipage}

    \begin{minipage}[b]{0.18\textwidth}
        \centering
        \subfigure{
            \includegraphics[width=\textwidth]{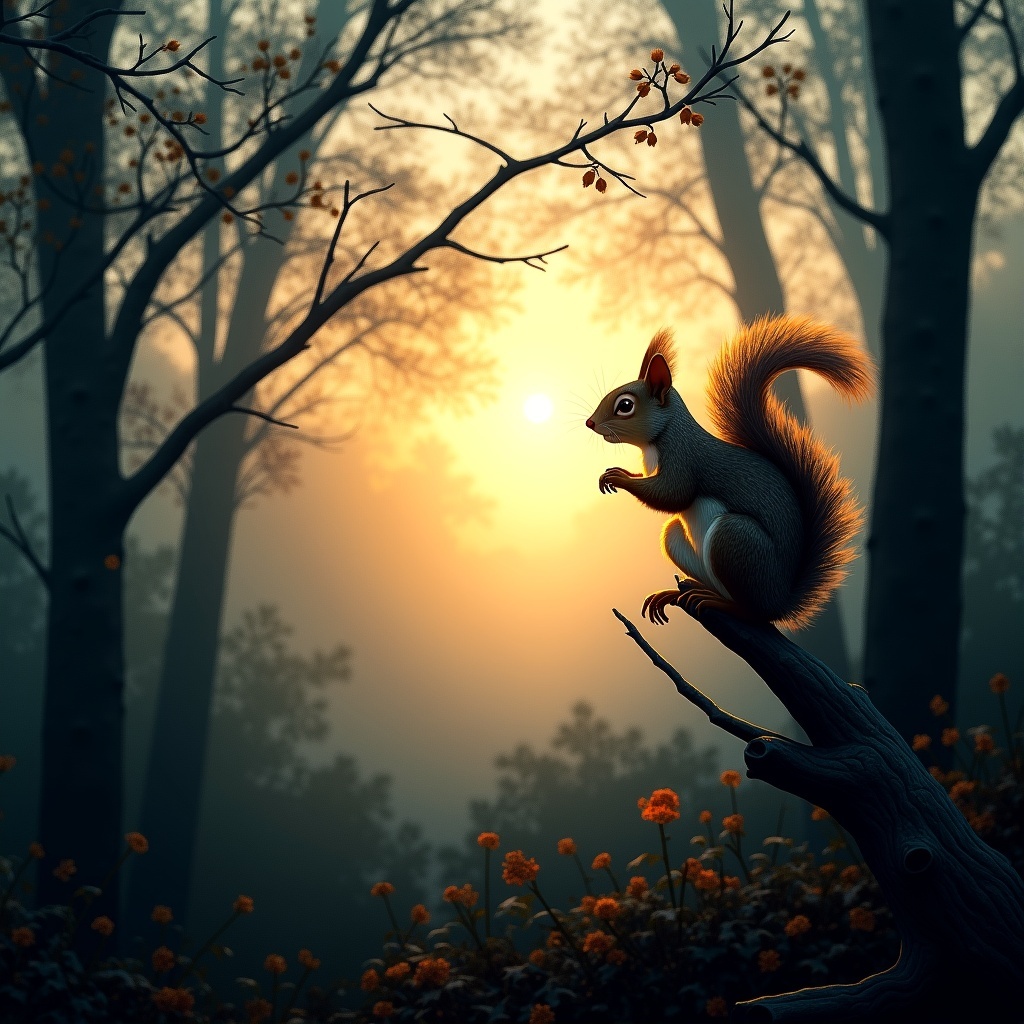}
        }
        \vspace{-0.6cm}
        \caption*{\centering \small center=5.0}
    \end{minipage}
    \begin{minipage}[b]{0.18\textwidth}
        \centering
        \subfigure{
            \includegraphics[width=\textwidth]{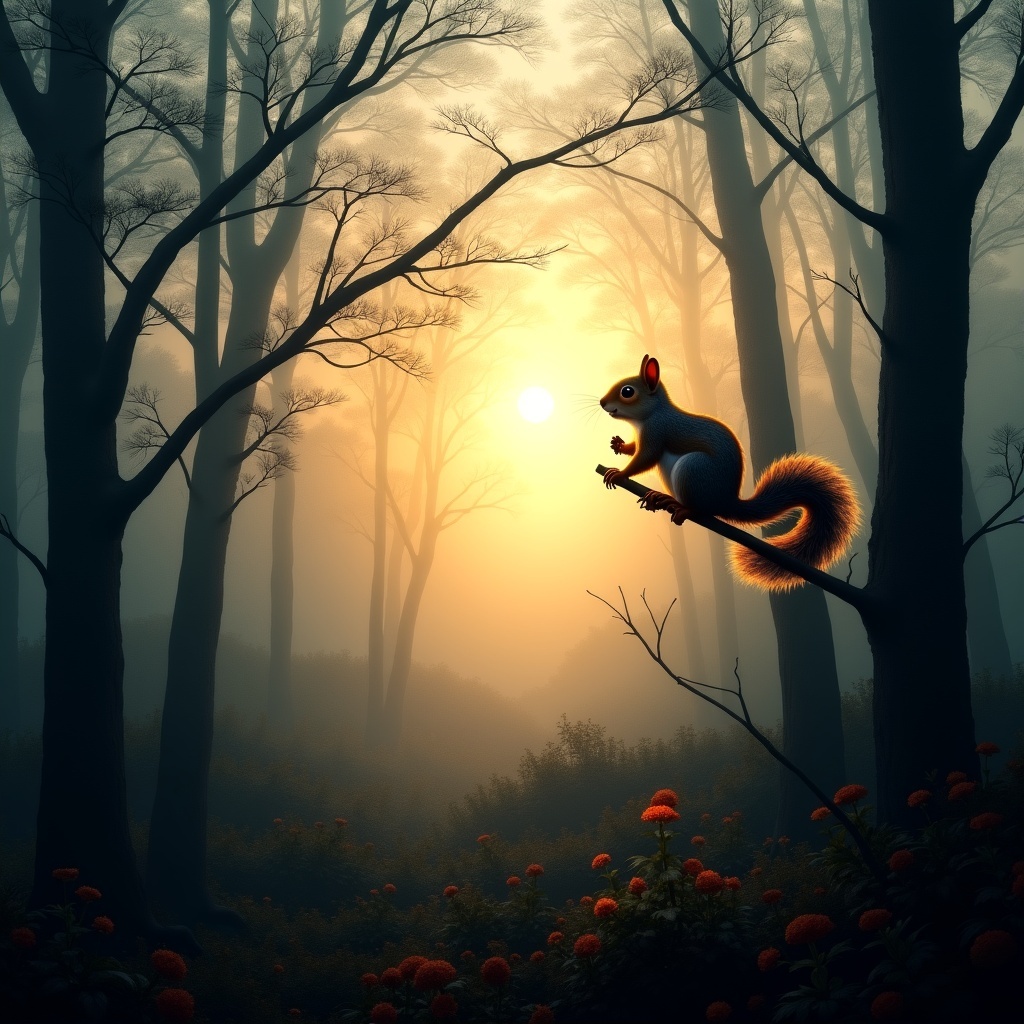}
        }
        \vspace{-0.6cm}
        \caption*{\centering \small center=5.6}
    \end{minipage}
    \begin{minipage}[b]{0.18\textwidth}
        \centering
        \subfigure{
            \includegraphics[width=\textwidth]{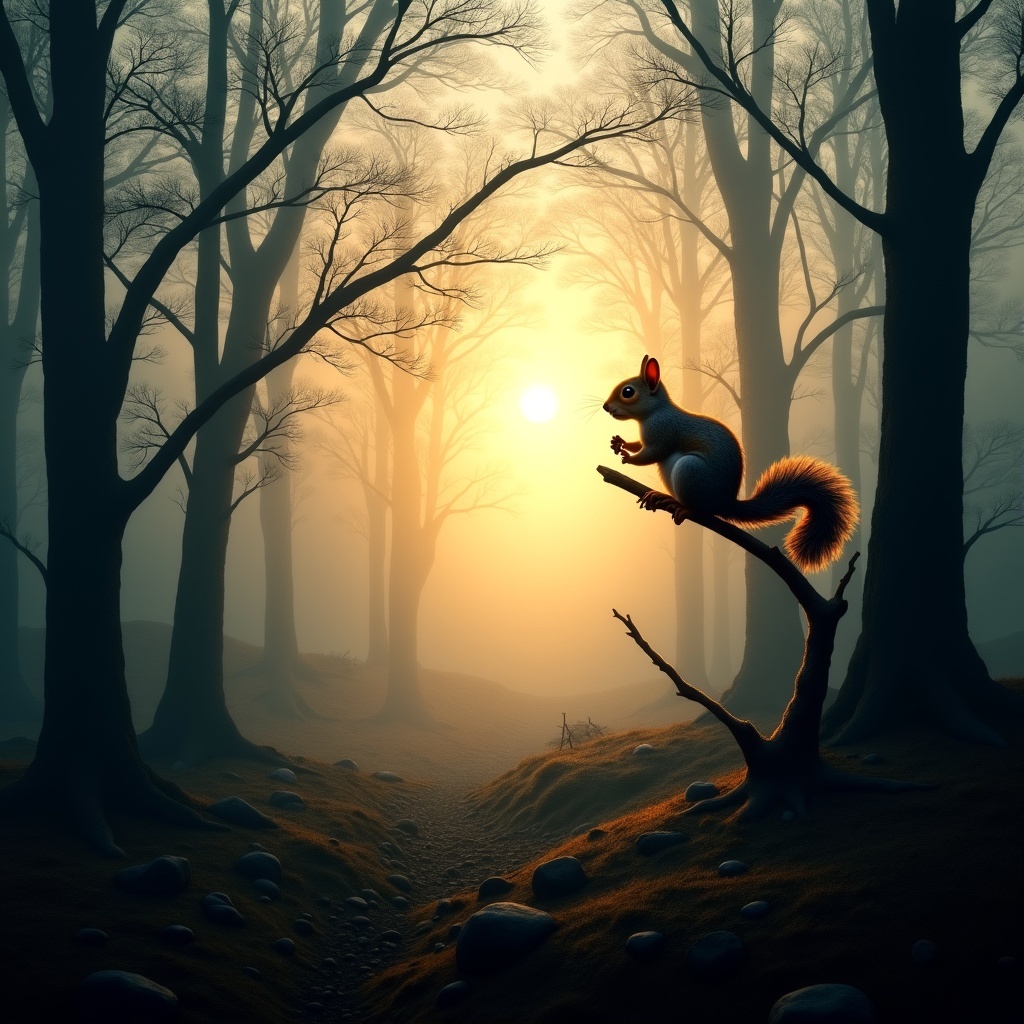}
        }
        \vspace{-0.6cm}
        \caption*{\centering \small center=6.1}
    \end{minipage}
    \begin{minipage}[b]{0.18\textwidth}
        \centering
        \subfigure{
            \includegraphics[width=\textwidth]{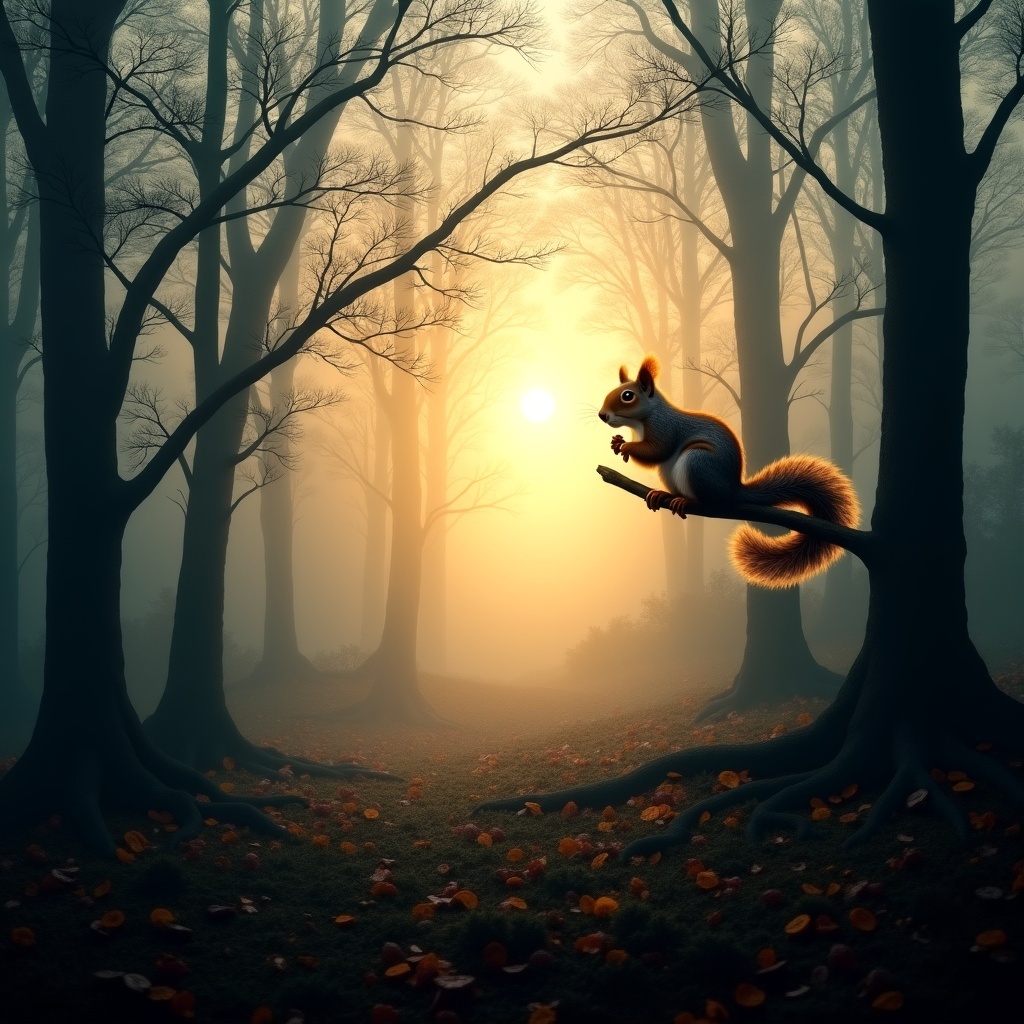}
        }
        \vspace{-0.6cm}
        \caption*{\centering \small center=6.7}
    \end{minipage}
    \begin{minipage}[b]{0.18\textwidth}
        \centering
        \subfigure{
            \includegraphics[width=\textwidth]{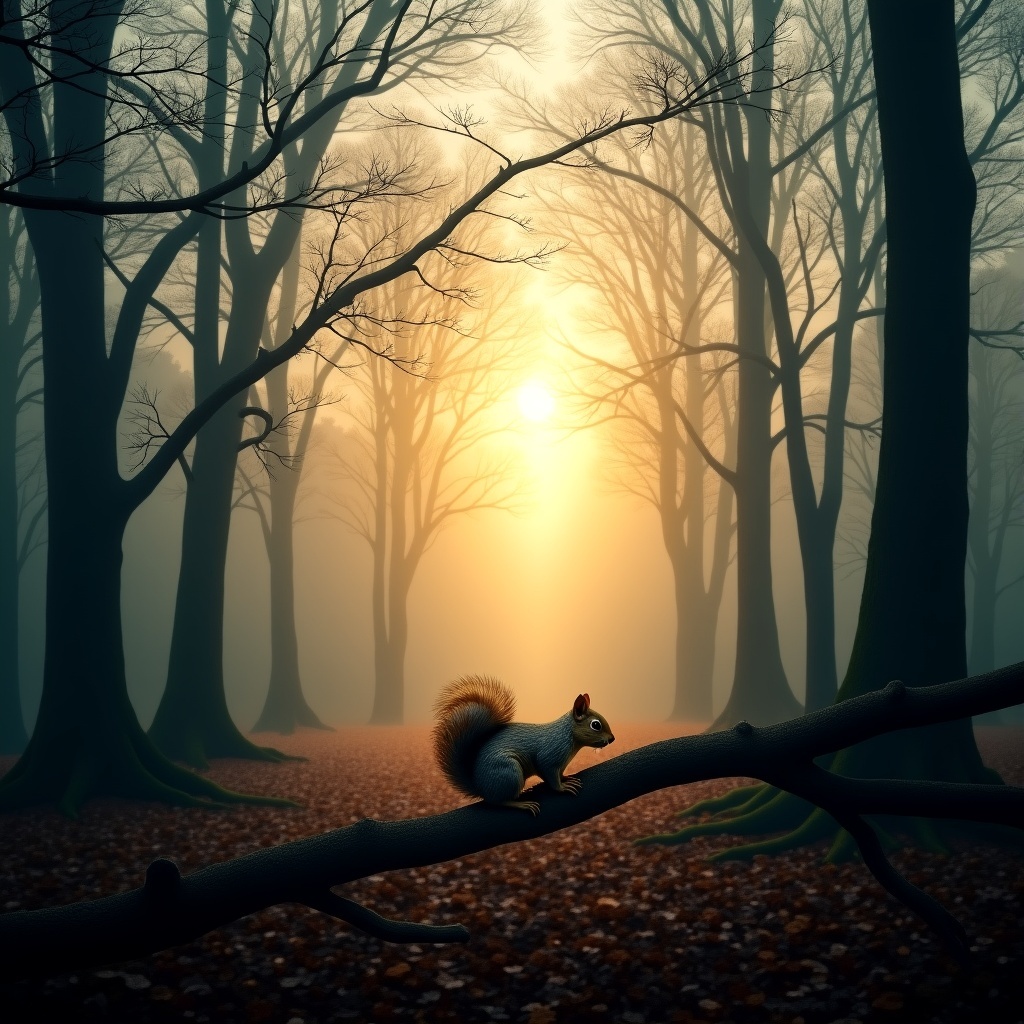}
        }
        \vspace{-0.6cm}
        \caption*{\centering \small center=7.2}
    \end{minipage}

    \vspace{0.1cm}
    
    \begin{minipage}[b]{0.18\textwidth}
        \centering
        \subfigure{
            \includegraphics[width=\textwidth]{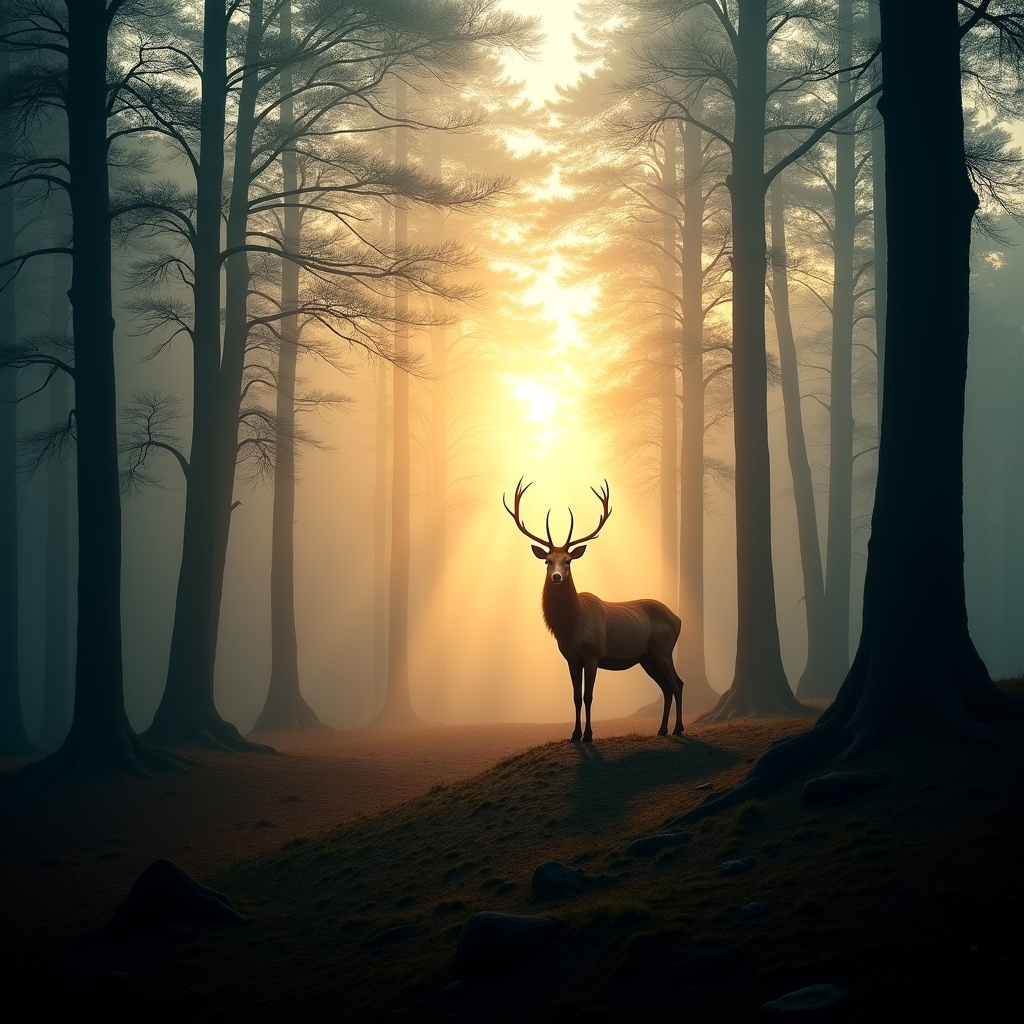}
        }
    \end{minipage}
    \begin{minipage}[b]{0.18\textwidth}
        \centering
        \subfigure{
            \includegraphics[width=\textwidth]{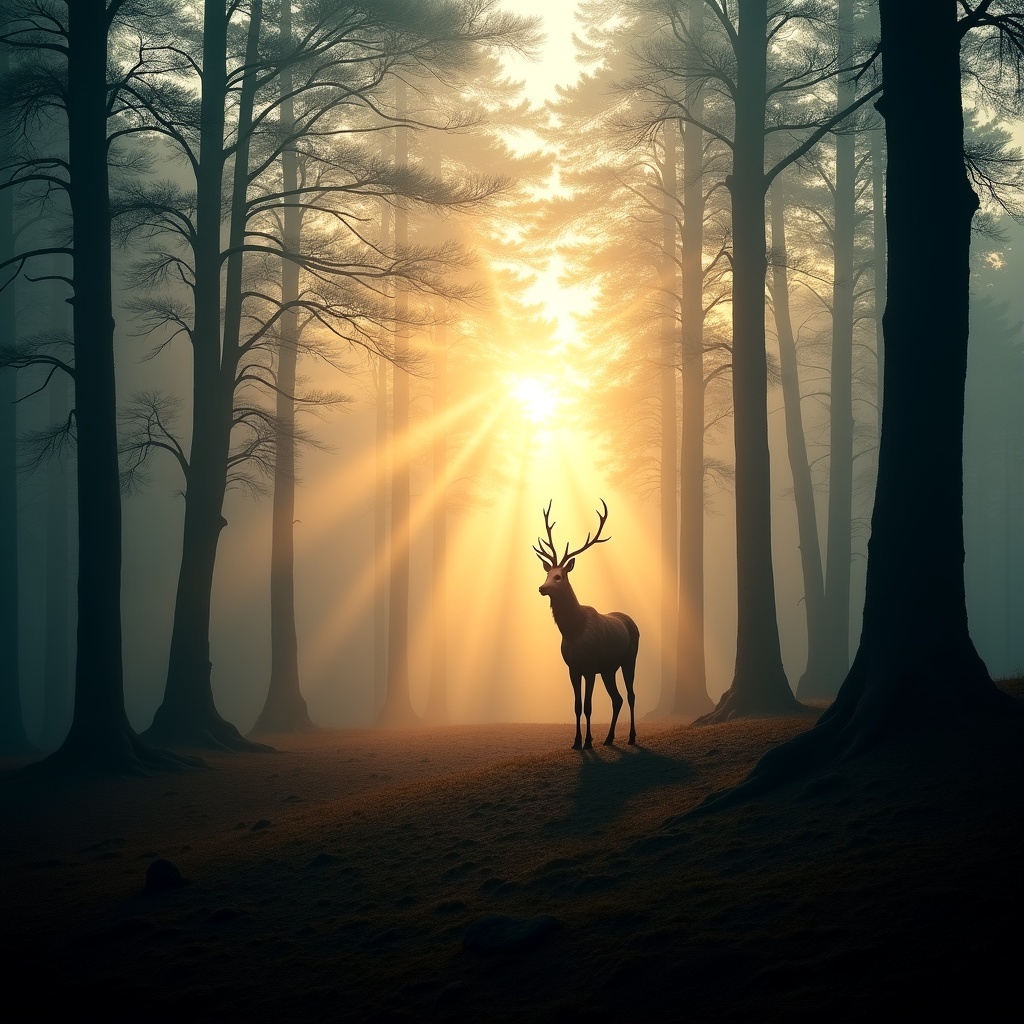}
        }
    \end{minipage}
    \begin{minipage}[b]{0.18\textwidth}
        \centering
        \subfigure{
            \includegraphics[width=\textwidth]{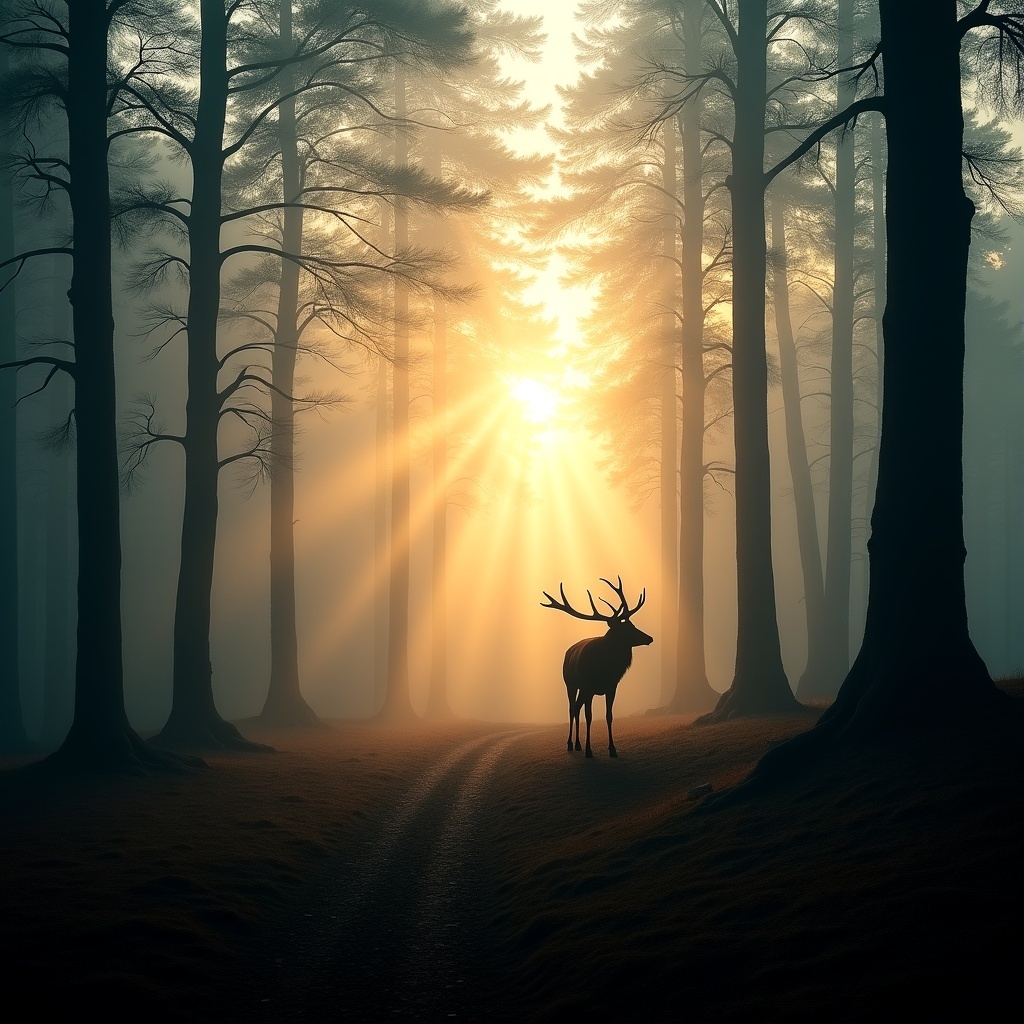}
        }
    \end{minipage}
    \begin{minipage}[b]{0.18\textwidth}
        \centering
        \subfigure{
            \includegraphics[width=\textwidth]{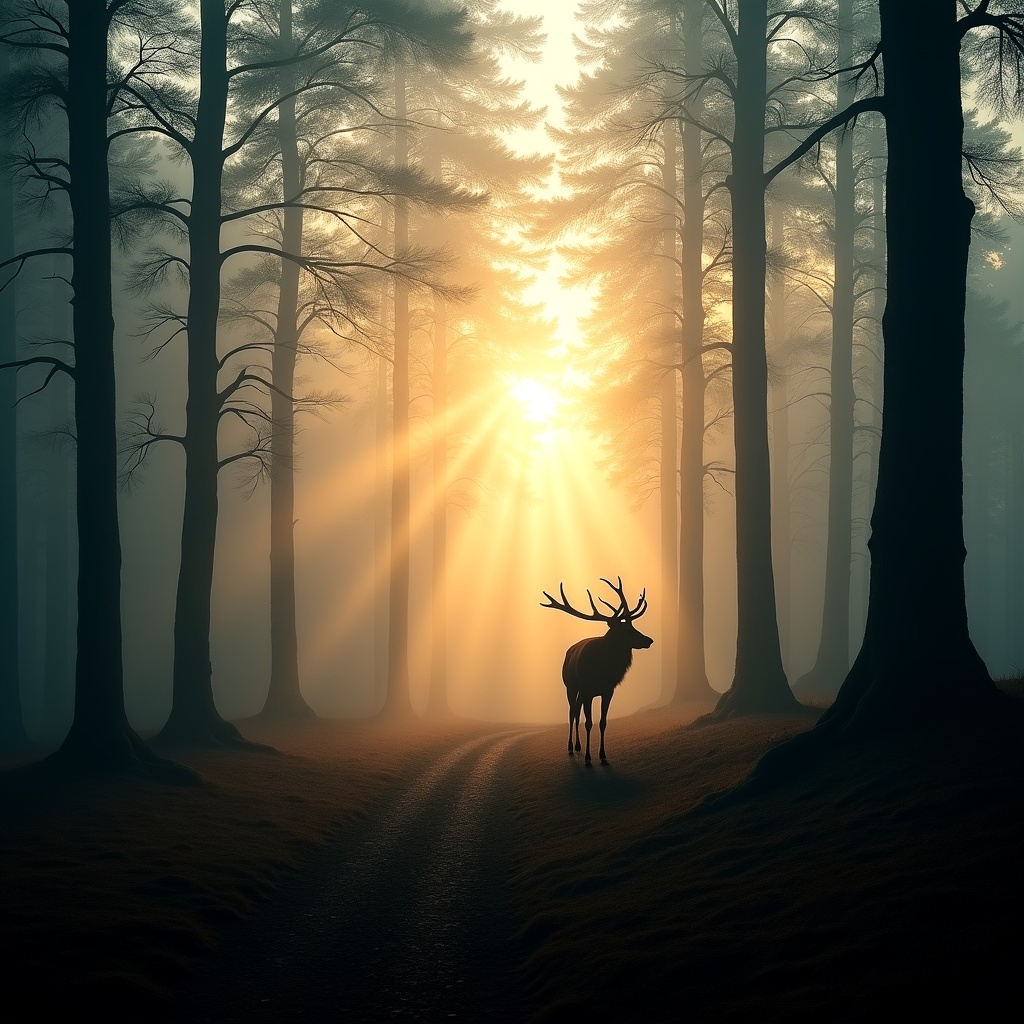}
        }
    \end{minipage}
    \begin{minipage}[b]{0.18\textwidth}
        \centering
        \subfigure{
            \includegraphics[width=\textwidth]{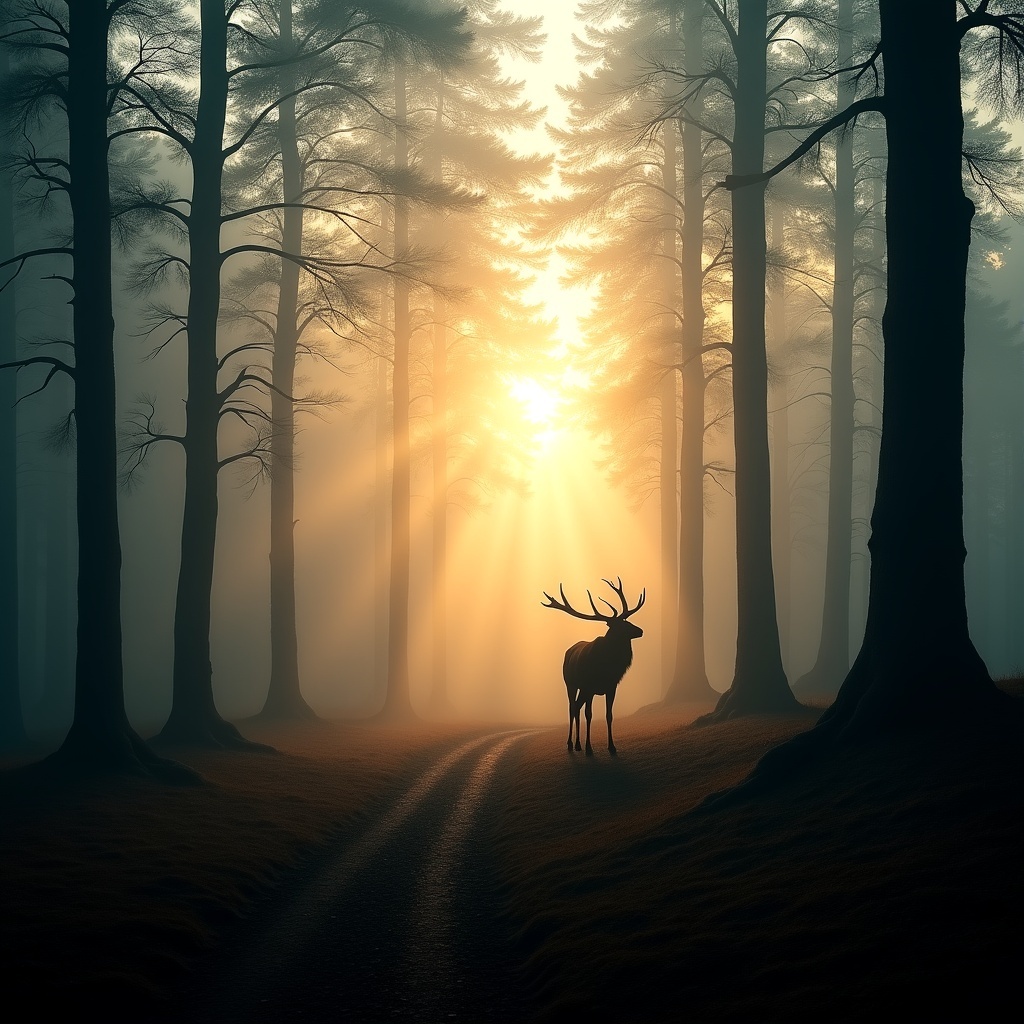}
        }
    \end{minipage}

    \begin{minipage}[b]{0.18\textwidth}
        \centering
        \subfigure{
            \includegraphics[width=\textwidth]{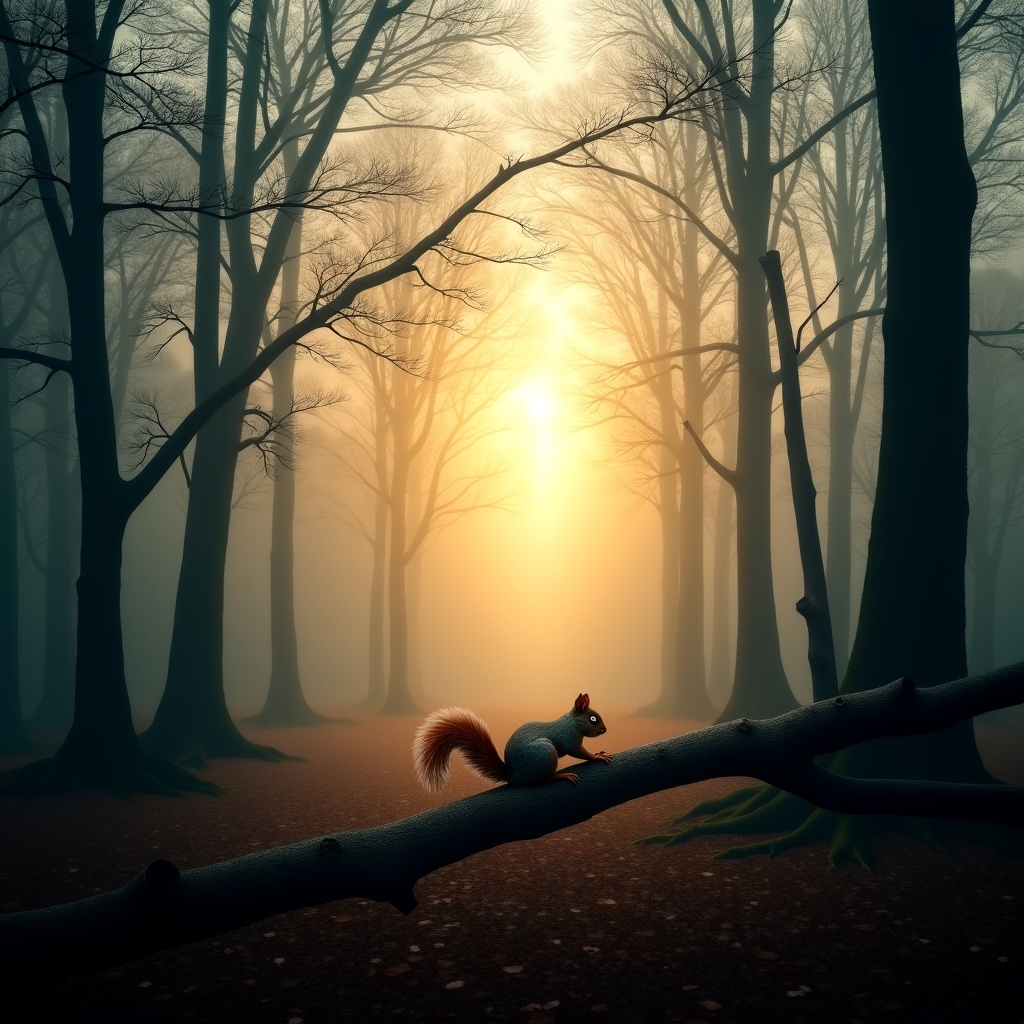}
        }
        \vspace{-0.6cm}
        \caption*{\centering \small center=7.8}
    \end{minipage}
    \begin{minipage}[b]{0.18\textwidth}
        \centering
        \subfigure{
            \includegraphics[width=\textwidth]{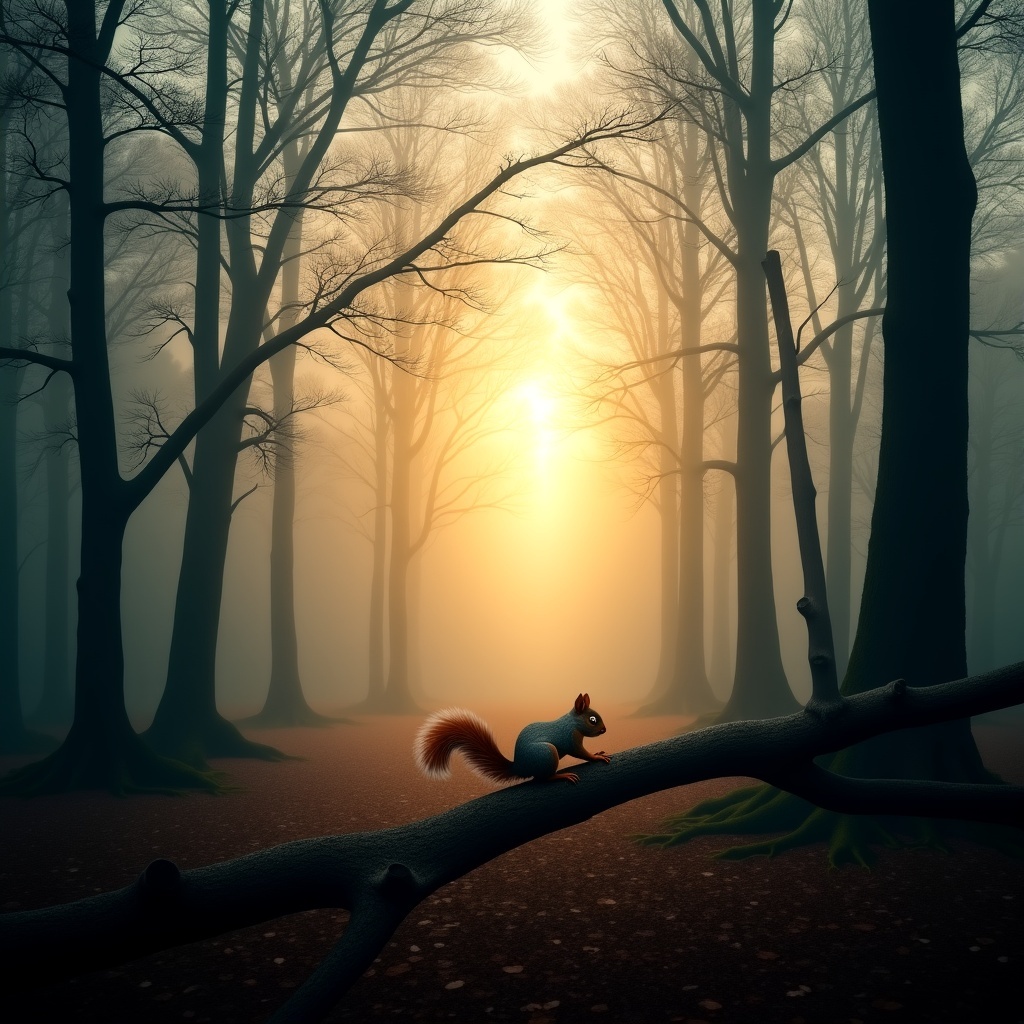}
        }
        \vspace{-0.6cm}
        \caption*{\centering \small center=8.3}
    \end{minipage}
    \begin{minipage}[b]{0.18\textwidth}
        \centering
        \subfigure{
            \includegraphics[width=\textwidth]{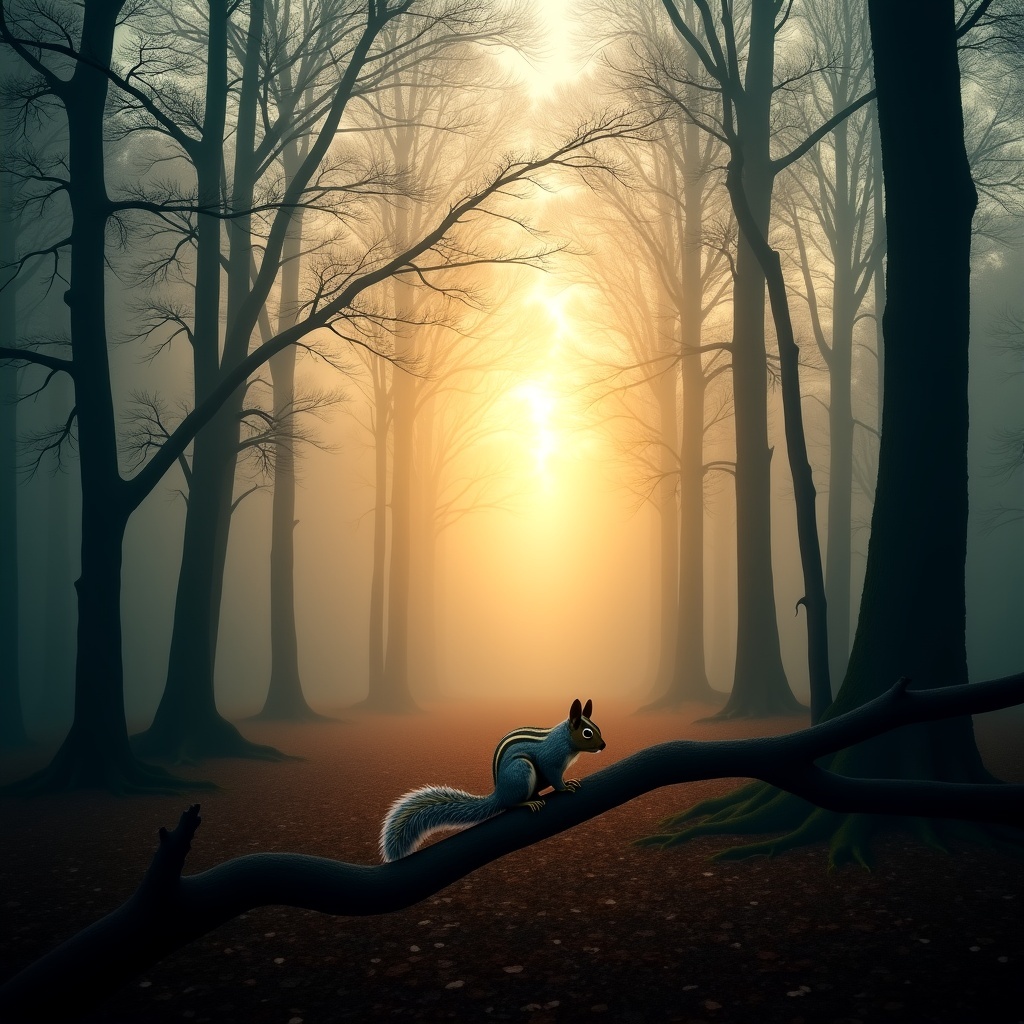}
        }
        \vspace{-0.6cm}
        \caption*{\centering \small center=8.9}
    \end{minipage}
    \begin{minipage}[b]{0.18\textwidth}
        \centering
        \subfigure{
            \includegraphics[width=\textwidth]{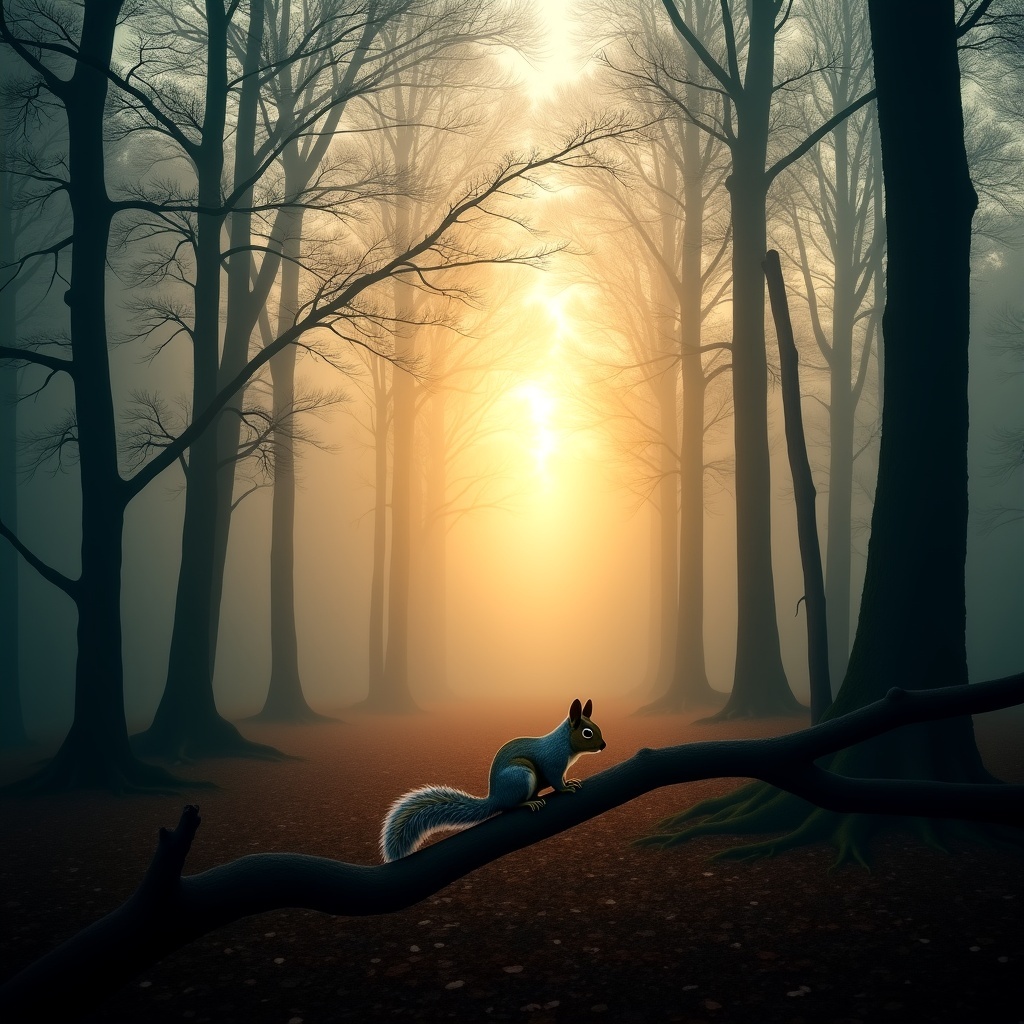}
        }
        \vspace{-0.6cm}
        \caption*{\centering \small center=9.4}
    \end{minipage}
    \begin{minipage}[b]{0.18\textwidth}
        \centering
        \subfigure{
            \includegraphics[width=\textwidth]{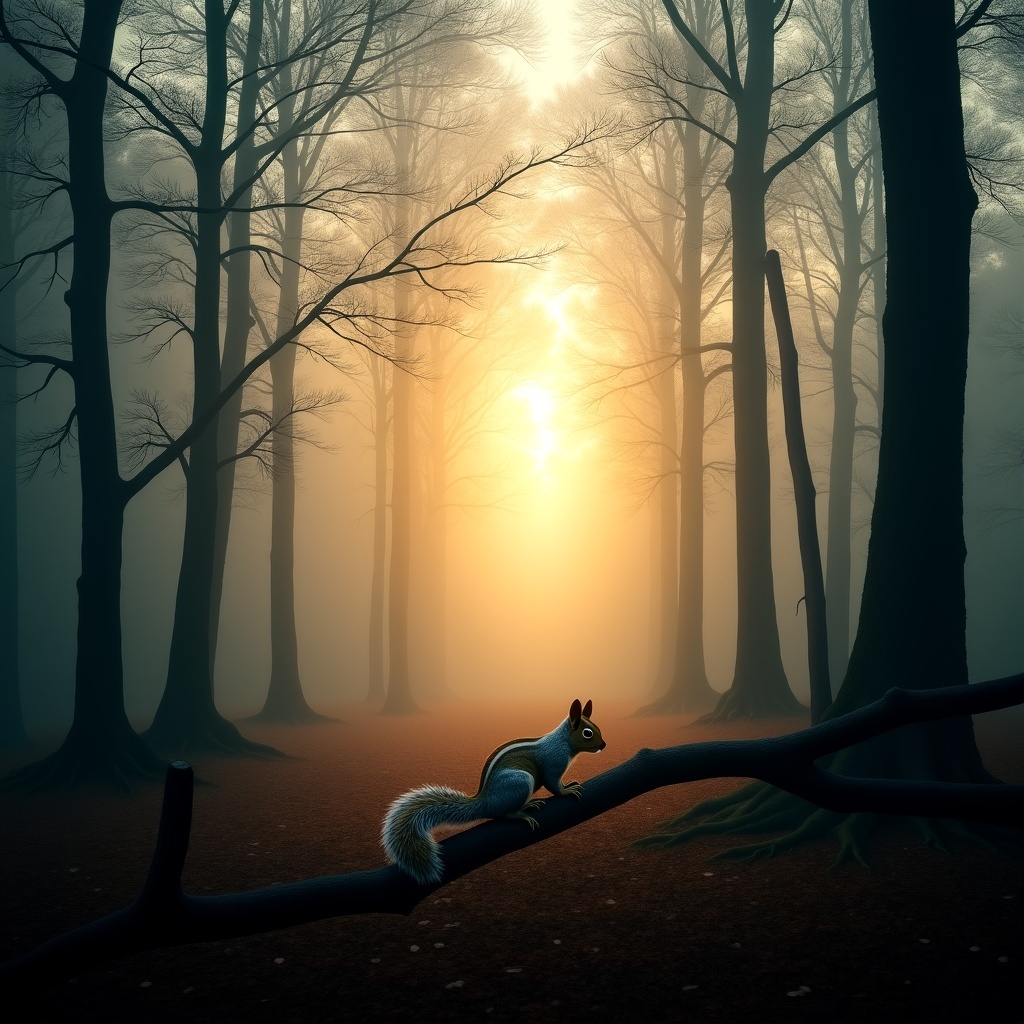}
        }
        \vspace{-0.6cm}
        \caption*{\centering \small center=10}
    \end{minipage}
    \caption{Comparison of different parameter centers. (sin) (scale=0.8)}
    \label{5}
\end{figure}

\begin{figure}[!htbp]
    \centering
    \begin{minipage}[b]{0.18\textwidth}
        \centering
        \subfigure{
            \includegraphics[width=\textwidth]{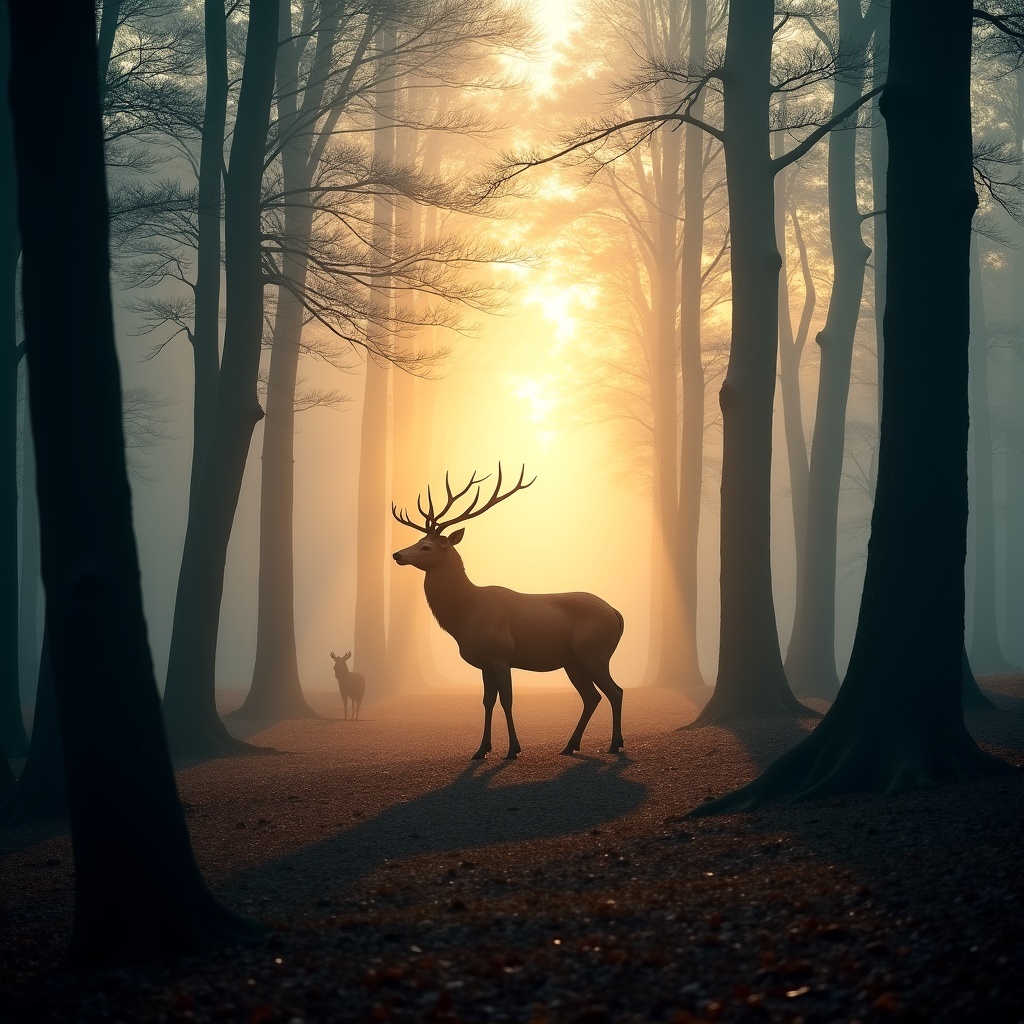}
        }
    \end{minipage}
    \begin{minipage}[b]{0.18\textwidth}
        \centering
        \subfigure{
            \includegraphics[width=\textwidth]{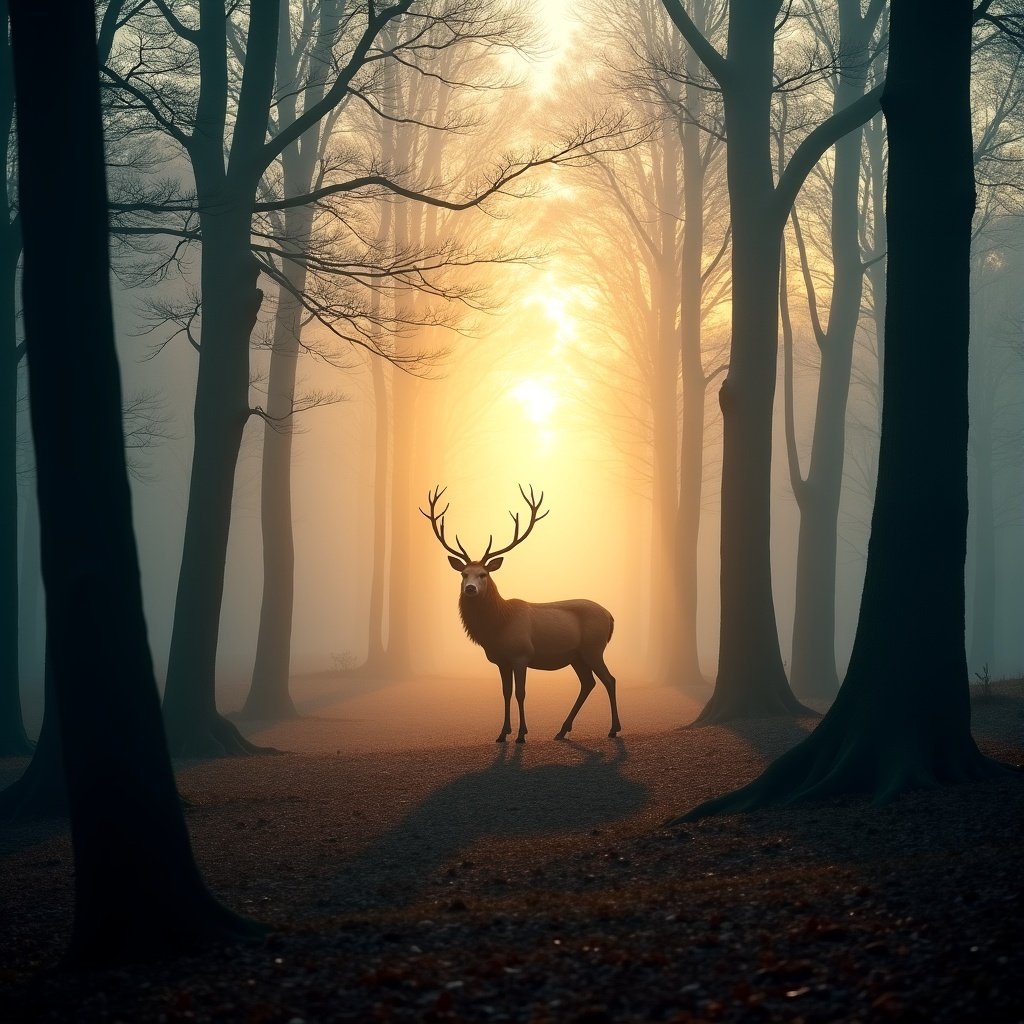}
        }
    \end{minipage}
    \begin{minipage}[b]{0.18\textwidth}
        \centering
        \subfigure{
            \includegraphics[width=\textwidth]{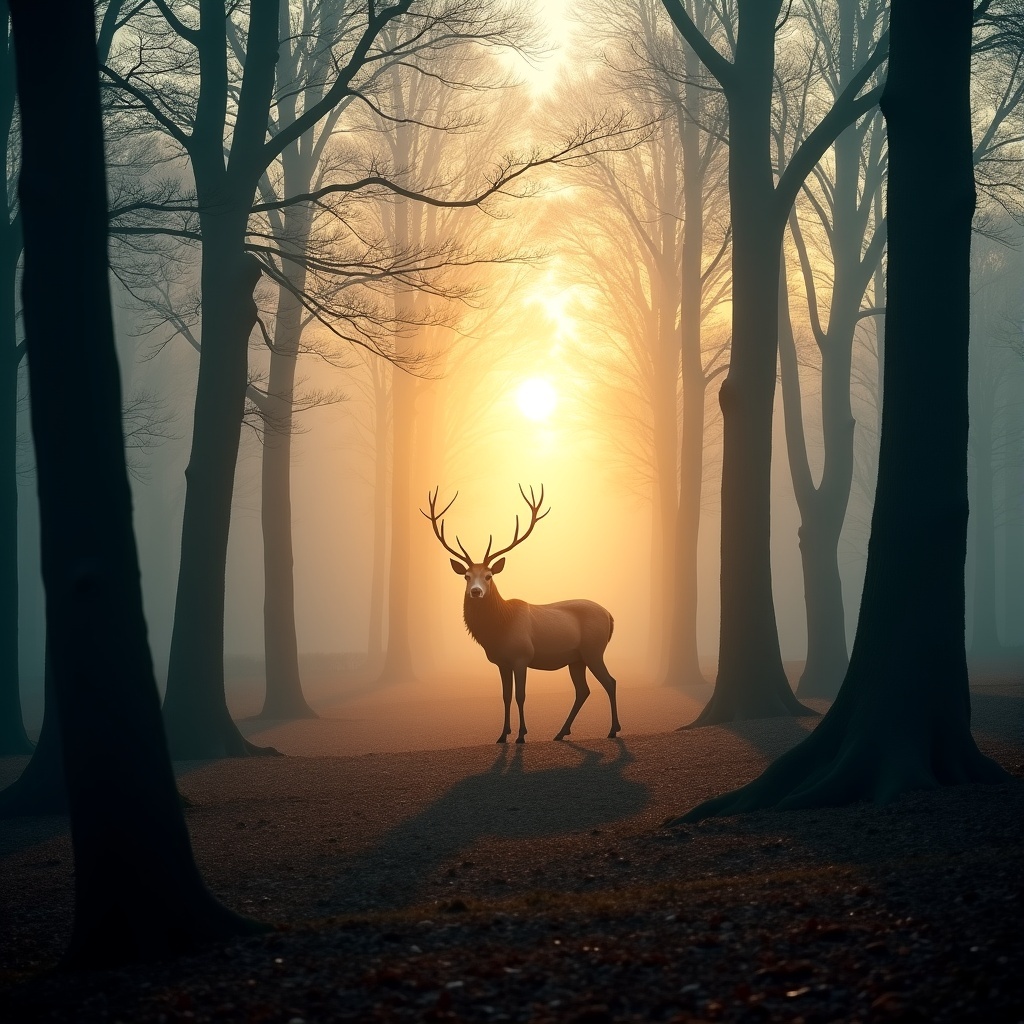}
        }
    \end{minipage}
    \begin{minipage}[b]{0.18\textwidth}
        \centering
        \subfigure{
            \includegraphics[width=\textwidth]{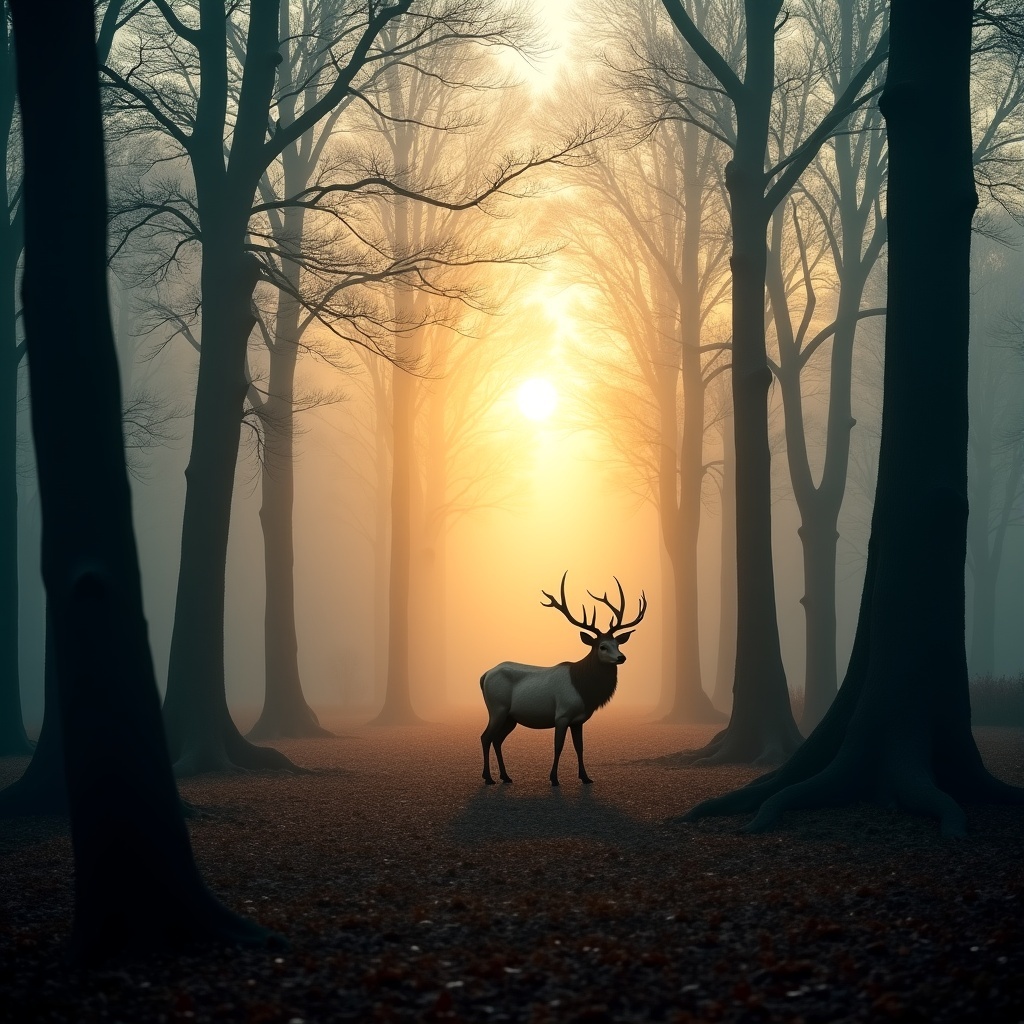}
        }
    \end{minipage}
    \begin{minipage}[b]{0.18\textwidth}
        \centering
        \subfigure{
            \includegraphics[width=\textwidth]{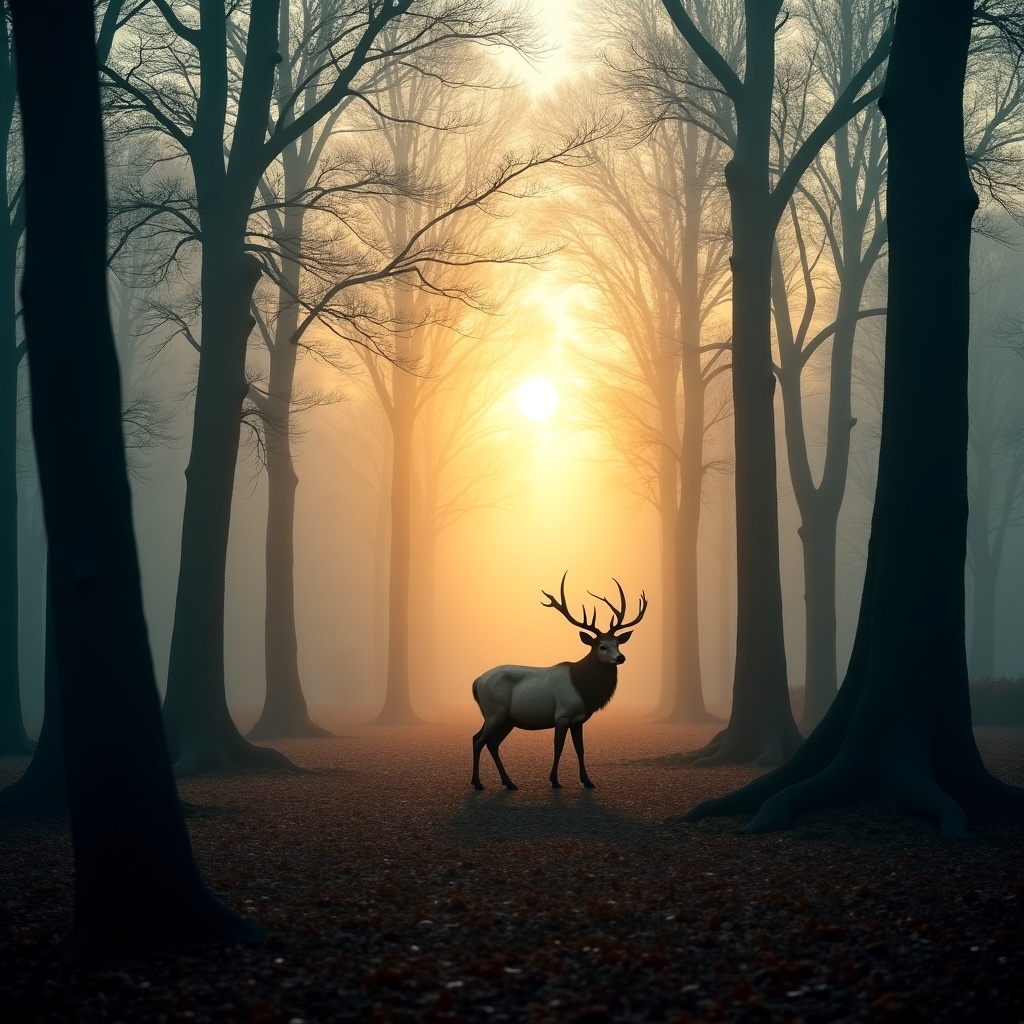}
        }
    \end{minipage}

    \begin{minipage}[b]{0.18\textwidth}
        \centering
        \subfigure{
            \includegraphics[width=\textwidth]{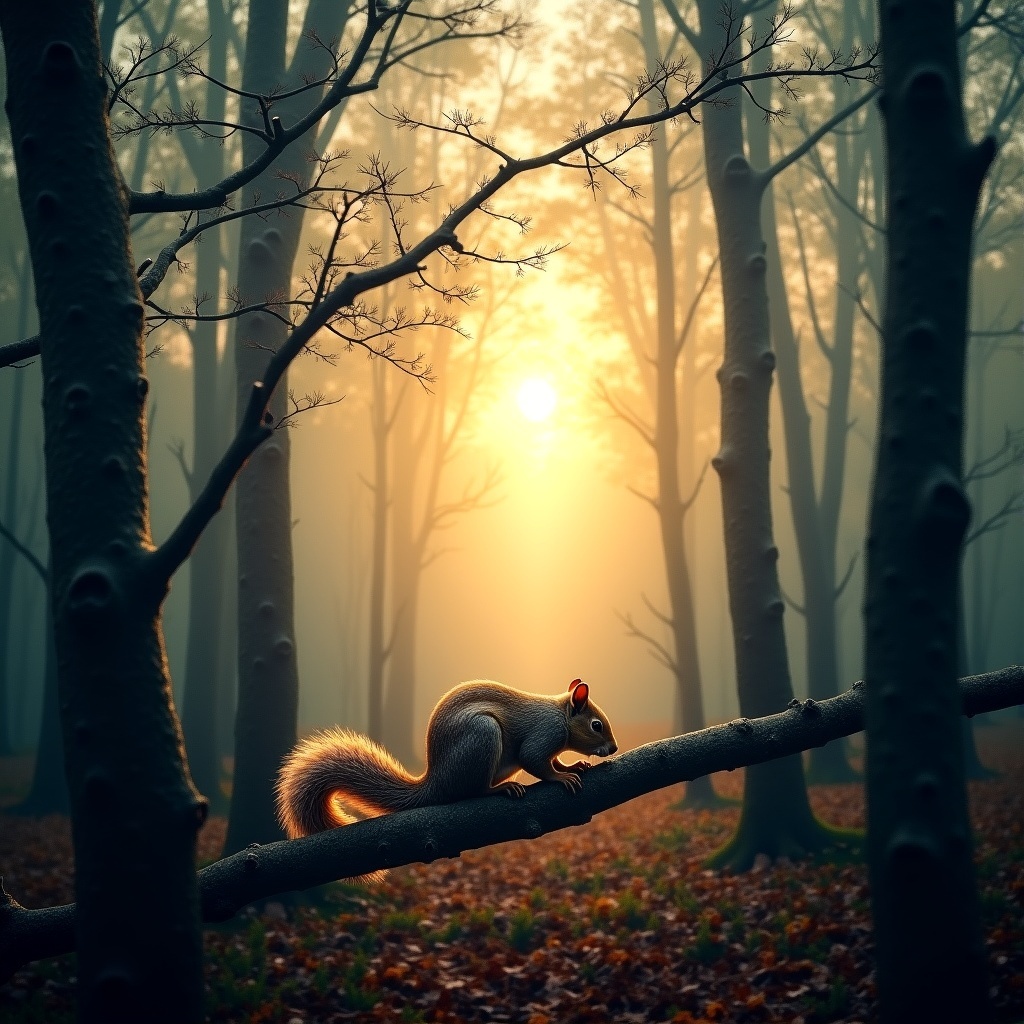}
        }
        \vspace{-0.6cm}
        \caption*{\centering \small center=5.0}
    \end{minipage}
    \begin{minipage}[b]{0.18\textwidth}
        \centering
        \subfigure{
            \includegraphics[width=\textwidth]{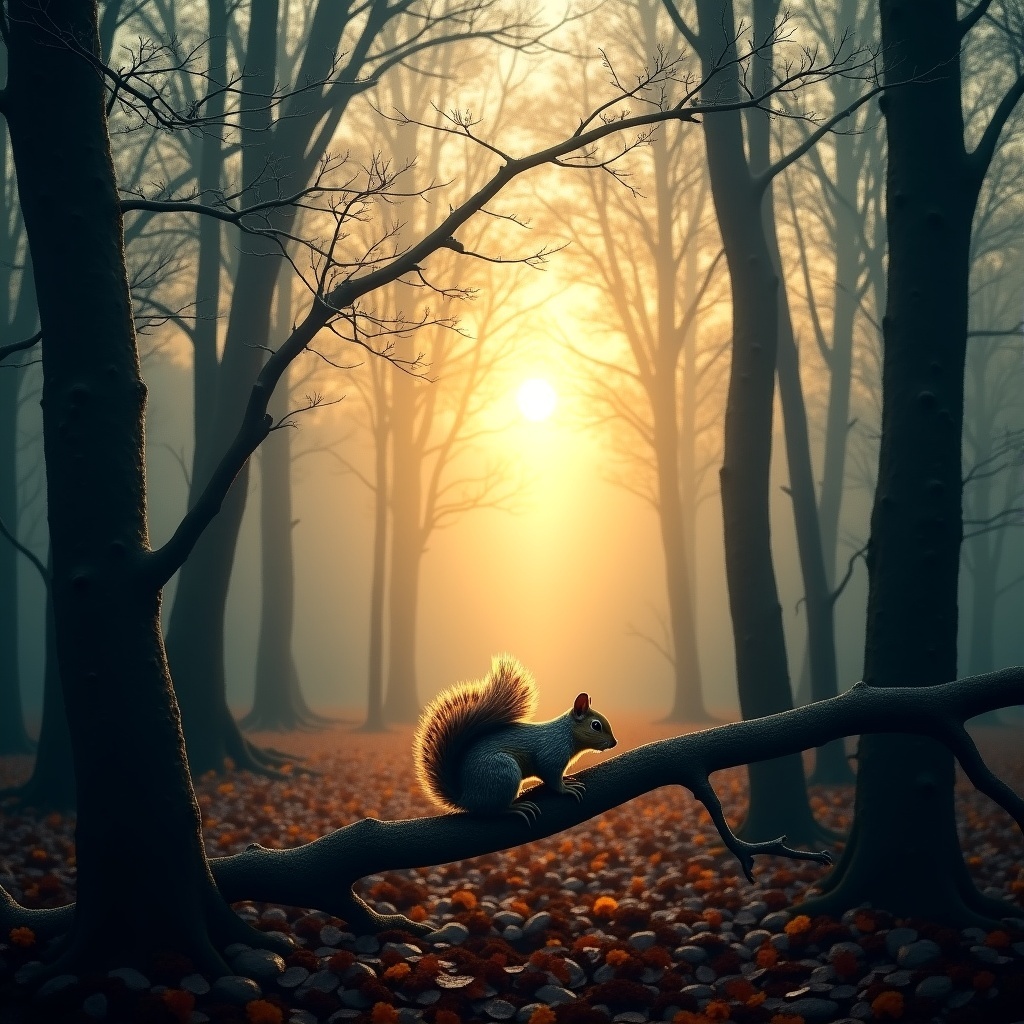}
        }
        \vspace{-0.6cm}
        \caption*{\centering \small center=5.6}
    \end{minipage}
    \begin{minipage}[b]{0.18\textwidth}
        \centering
        \subfigure{
            \includegraphics[width=\textwidth]{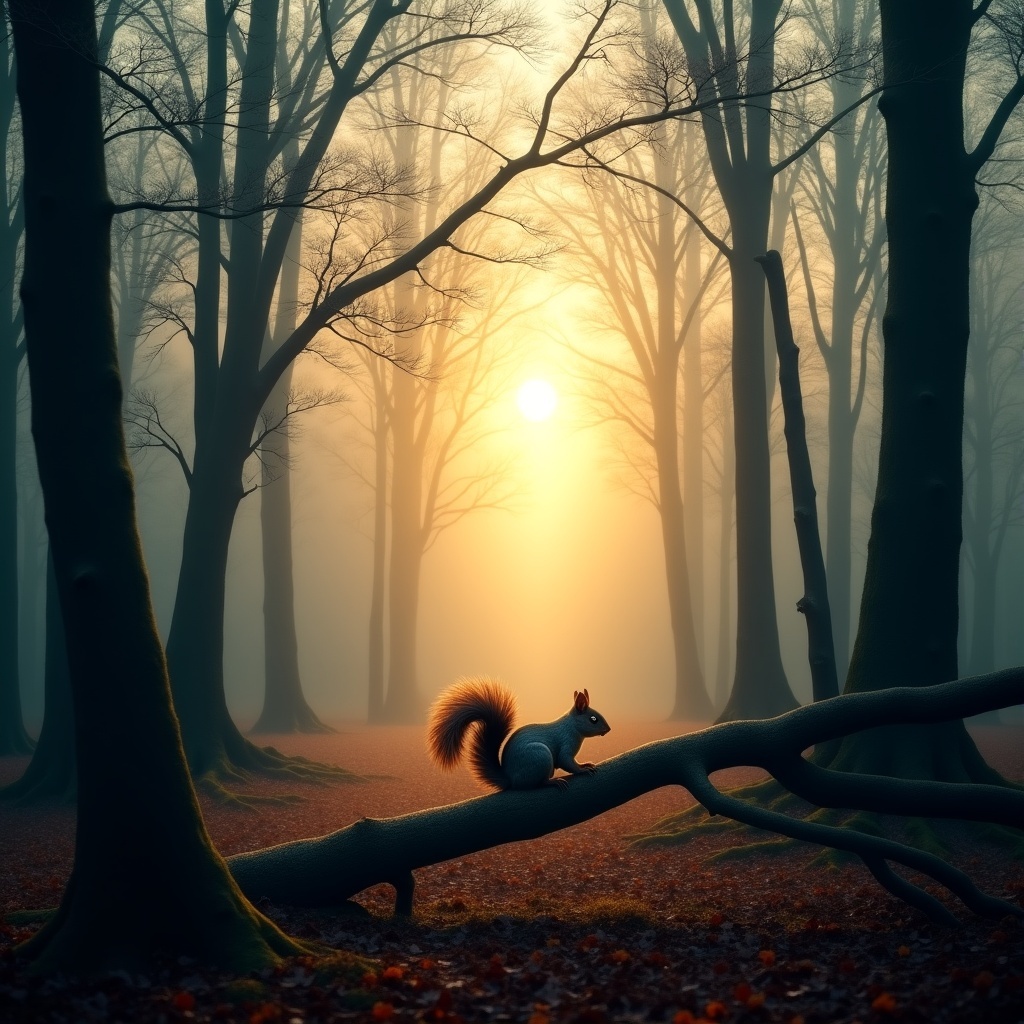}
        }
        \vspace{-0.6cm}
        \caption*{\centering \small center=6.1}
    \end{minipage}
    \begin{minipage}[b]{0.18\textwidth}
        \centering
        \subfigure{
            \includegraphics[width=\textwidth]{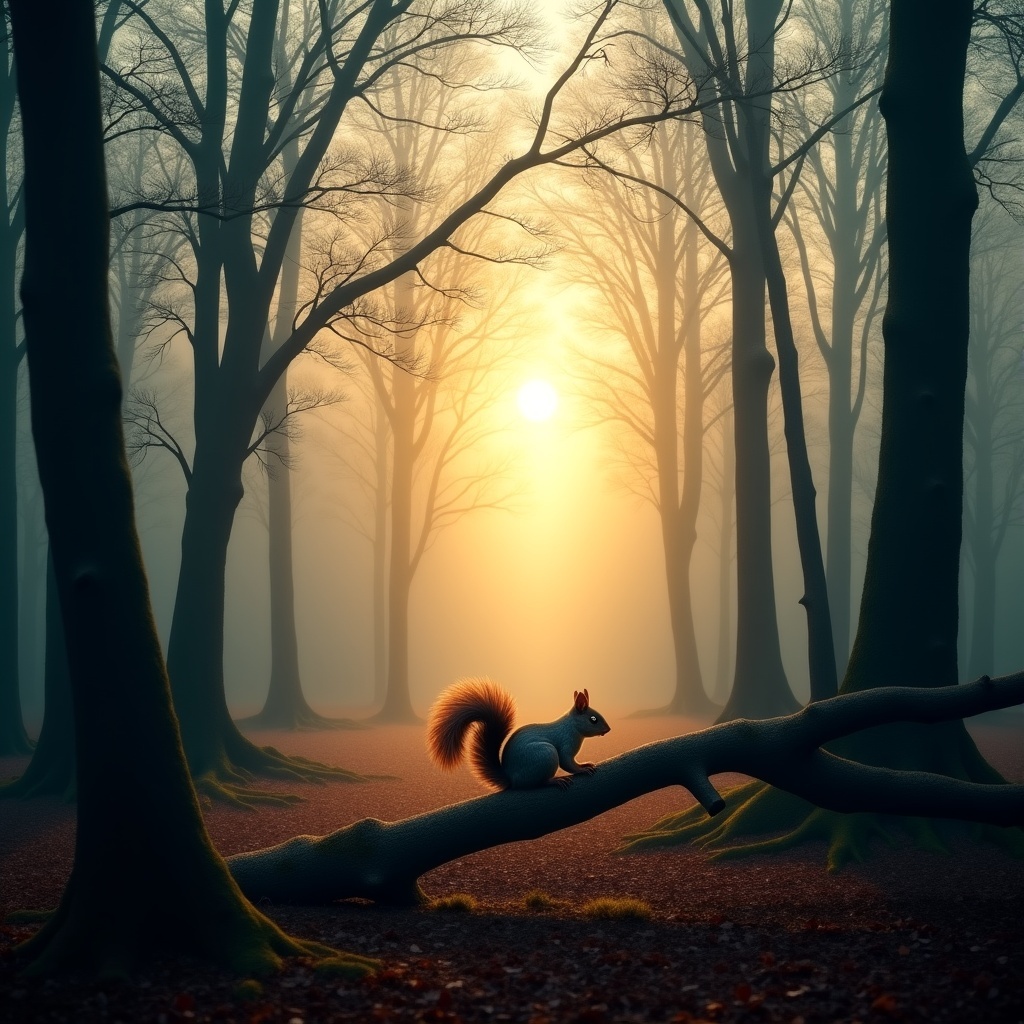}
        }
        \vspace{-0.6cm}
        \caption*{\centering \small center=6.7}
    \end{minipage}
    \begin{minipage}[b]{0.18\textwidth}
        \centering
        \subfigure{
            \includegraphics[width=\textwidth]{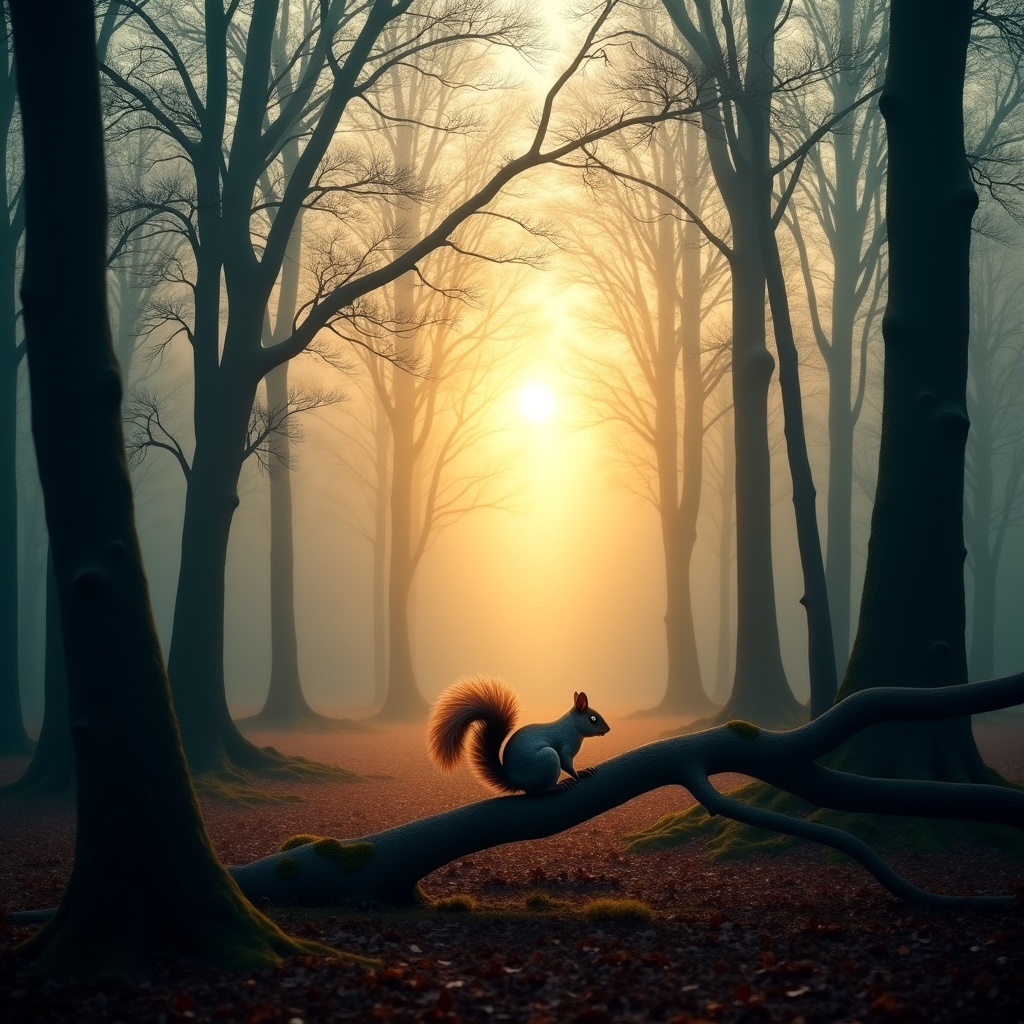}
        }
        \vspace{-0.6cm}
        \caption*{\centering \small center=7.2}
    \end{minipage}
    
    \vspace{0.1cm}
    
    \begin{minipage}[b]{0.18\textwidth}
        \centering
        \subfigure{
            \includegraphics[width=\textwidth]{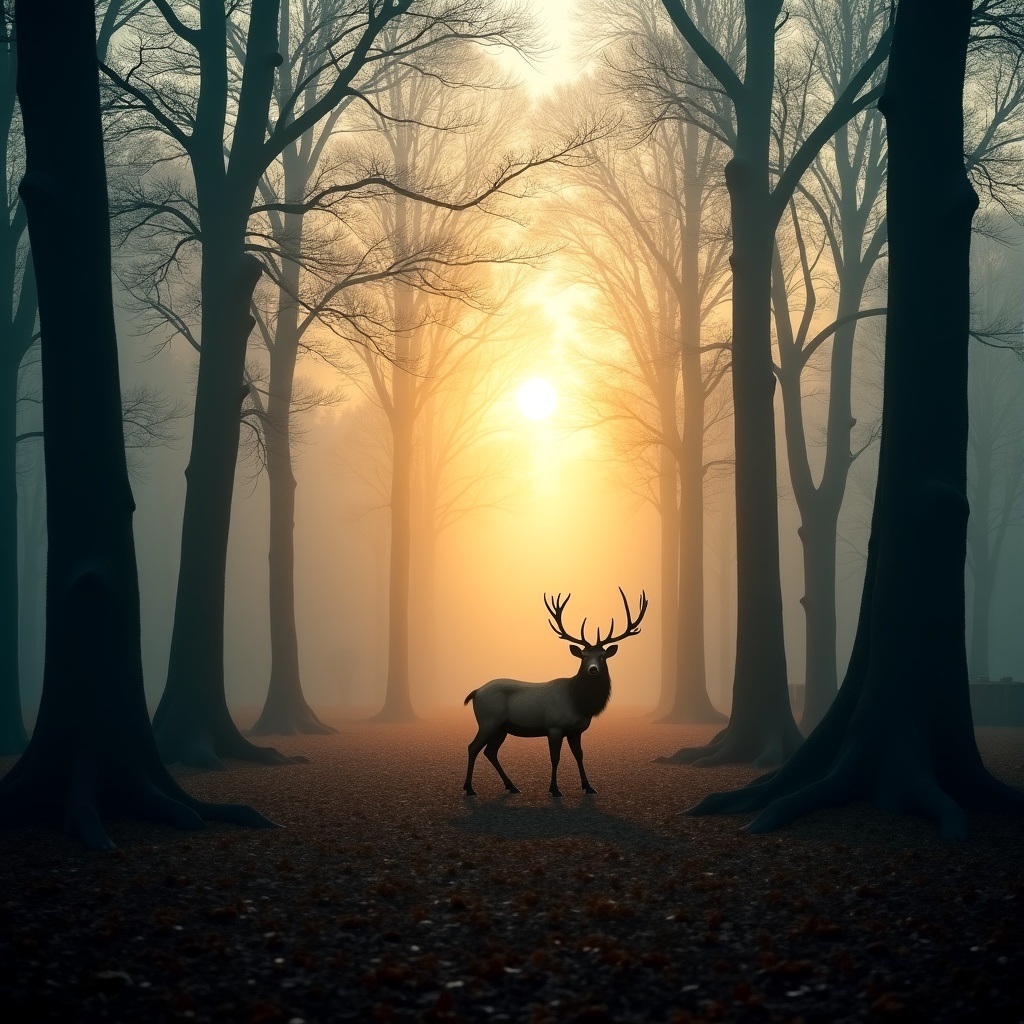}
        }
    \end{minipage}
    \begin{minipage}[b]{0.18\textwidth}
        \centering
        \subfigure{
            \includegraphics[width=\textwidth]{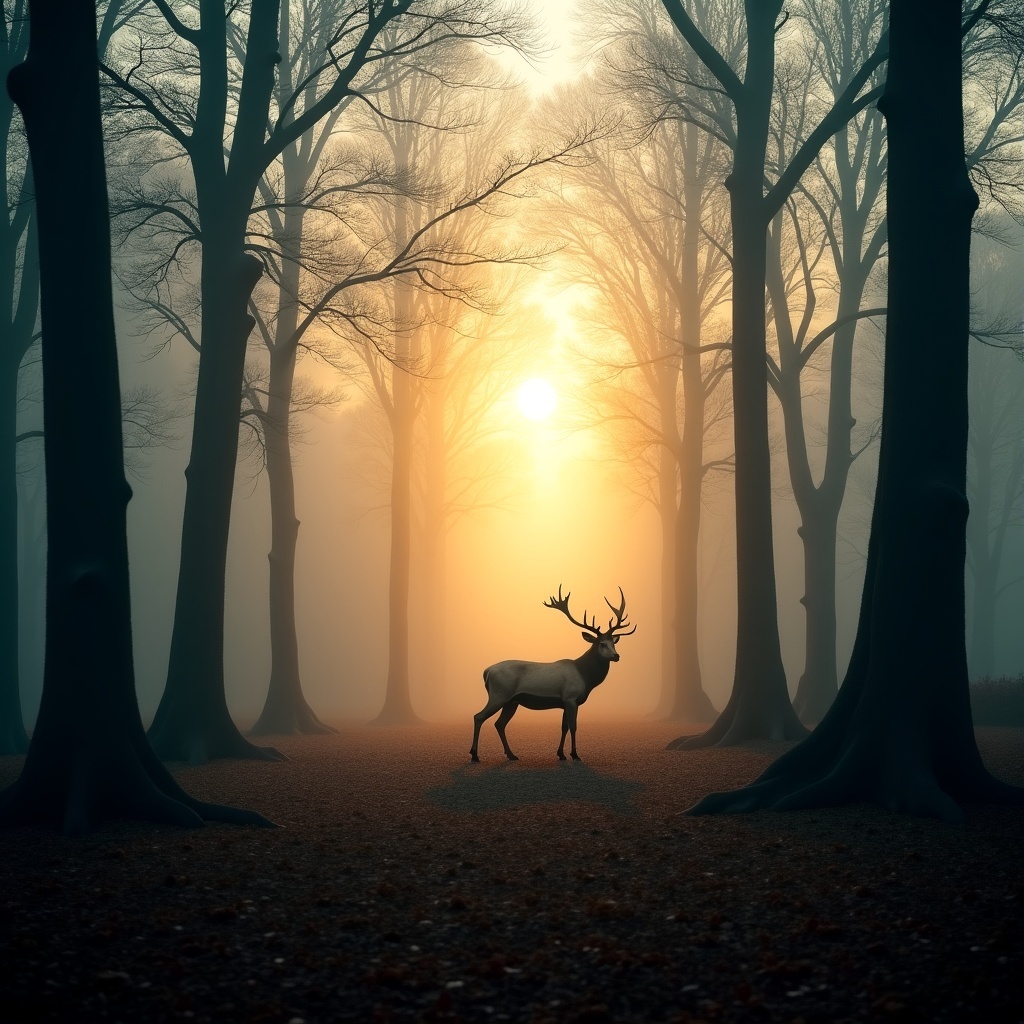}
        }
    \end{minipage}
    \begin{minipage}[b]{0.18\textwidth}
        \centering
        \subfigure{
            \includegraphics[width=\textwidth]{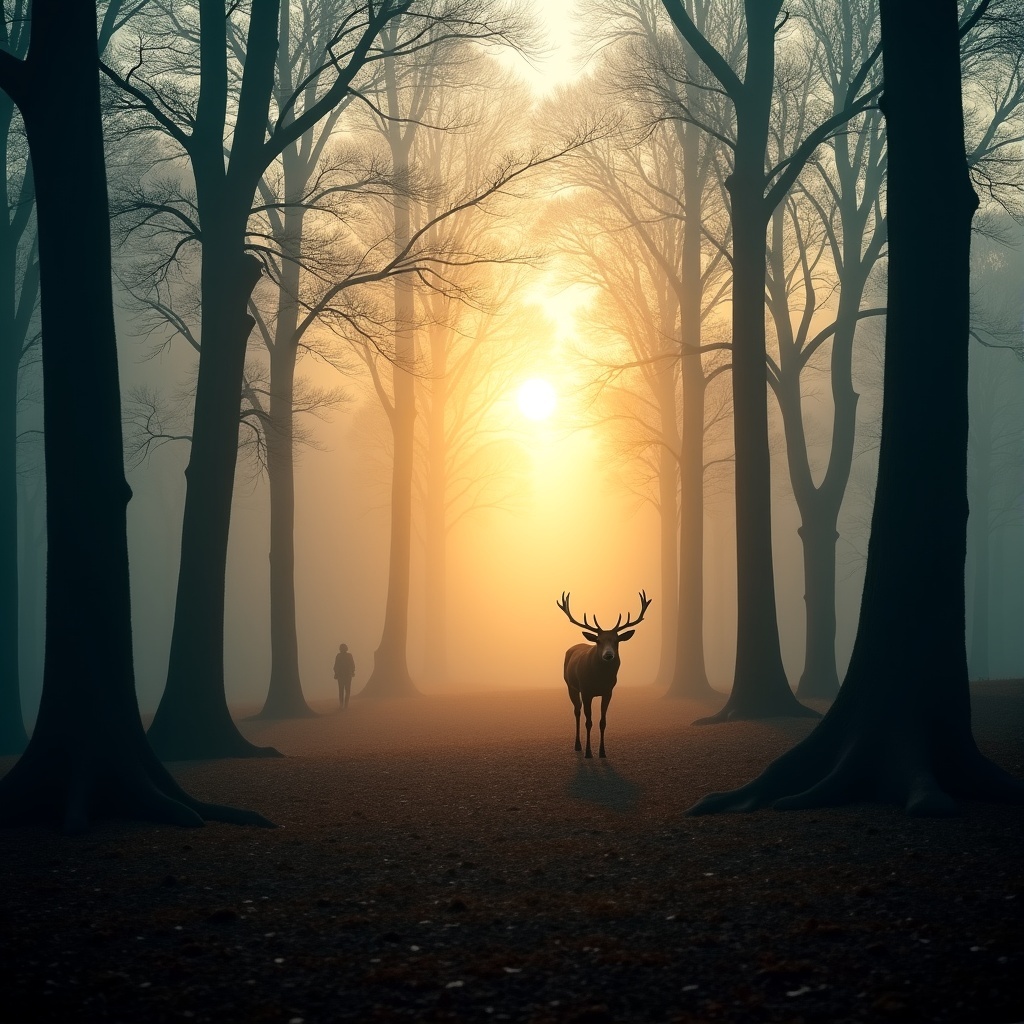}
        }
    \end{minipage}
    \begin{minipage}[b]{0.18\textwidth}
        \centering
        \subfigure{
            \includegraphics[width=\textwidth]{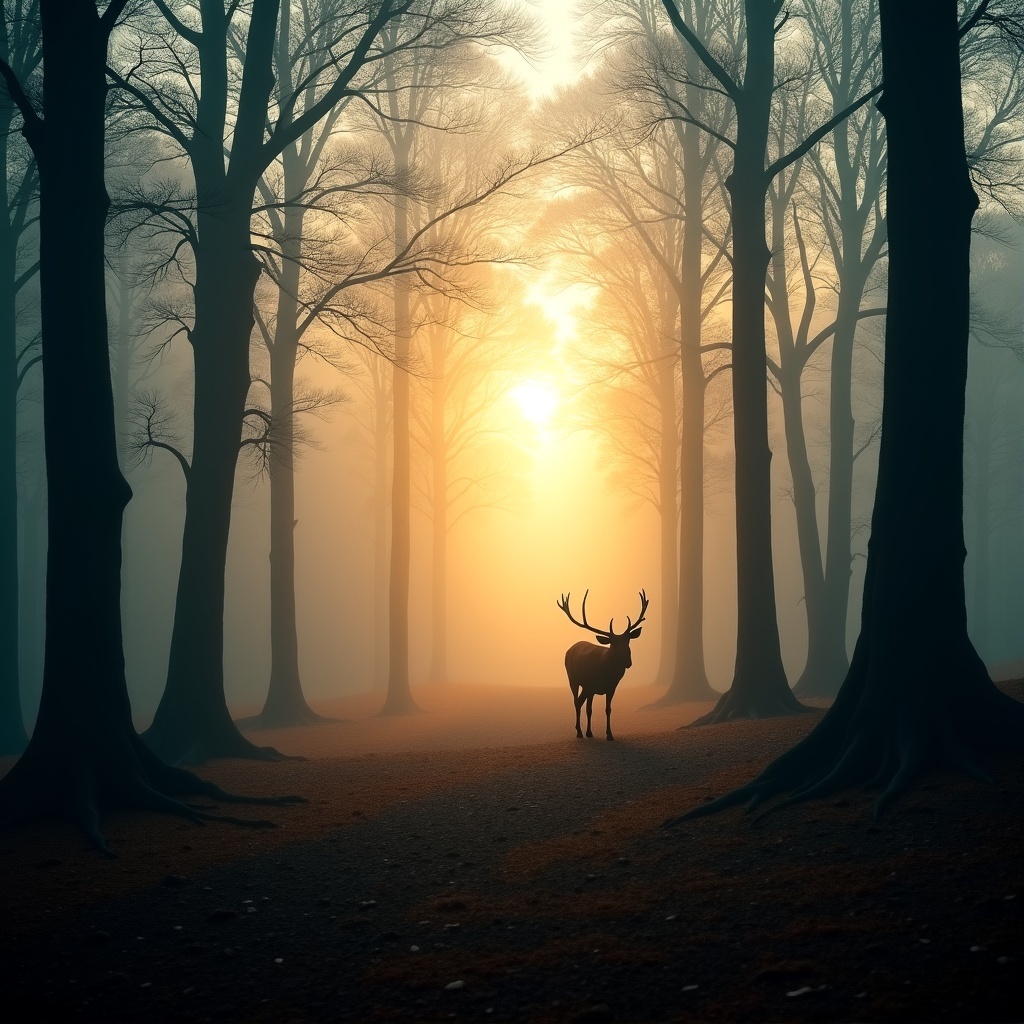}
        }
    \end{minipage}
    \begin{minipage}[b]{0.18\textwidth}
        \centering
        \subfigure{
            \includegraphics[width=\textwidth]{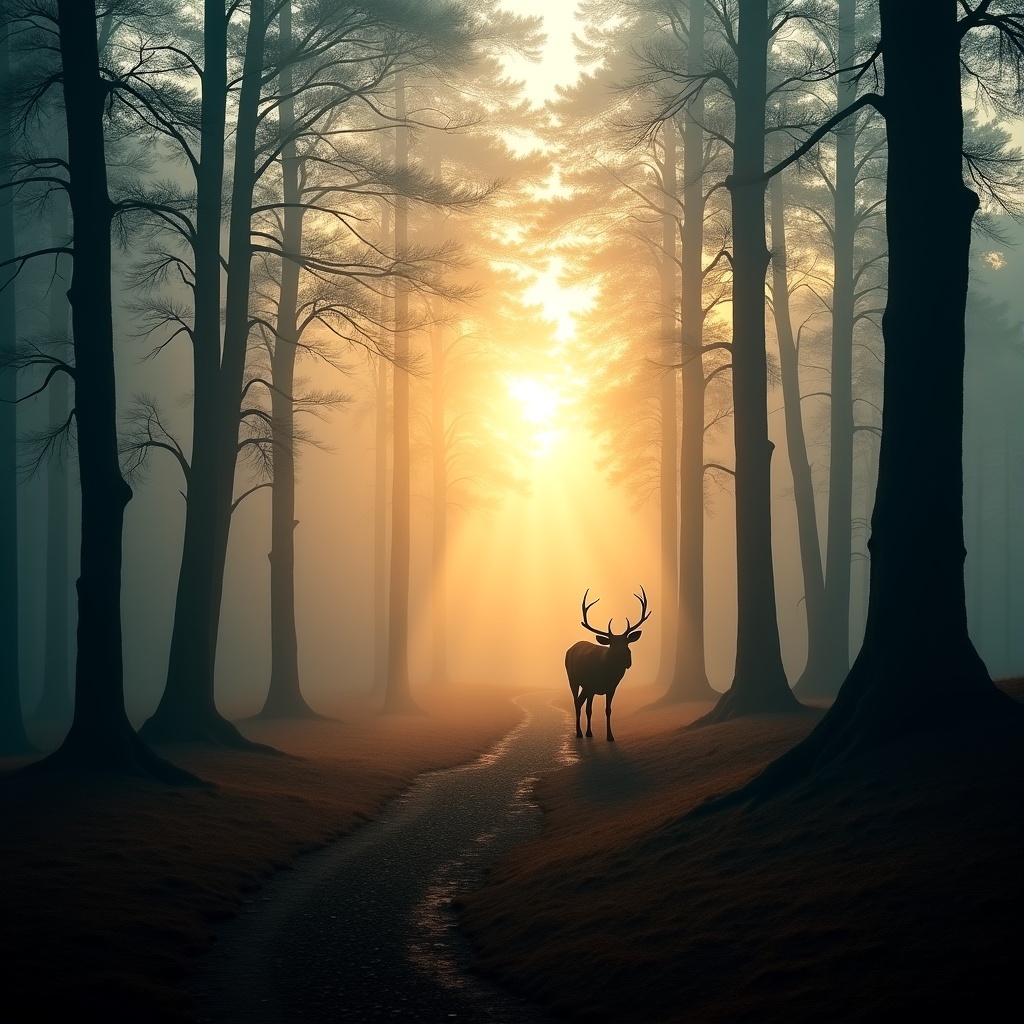}
        }
    \end{minipage}

    \begin{minipage}[b]{0.18\textwidth}
        \centering
        \subfigure{
            \includegraphics[width=\textwidth]{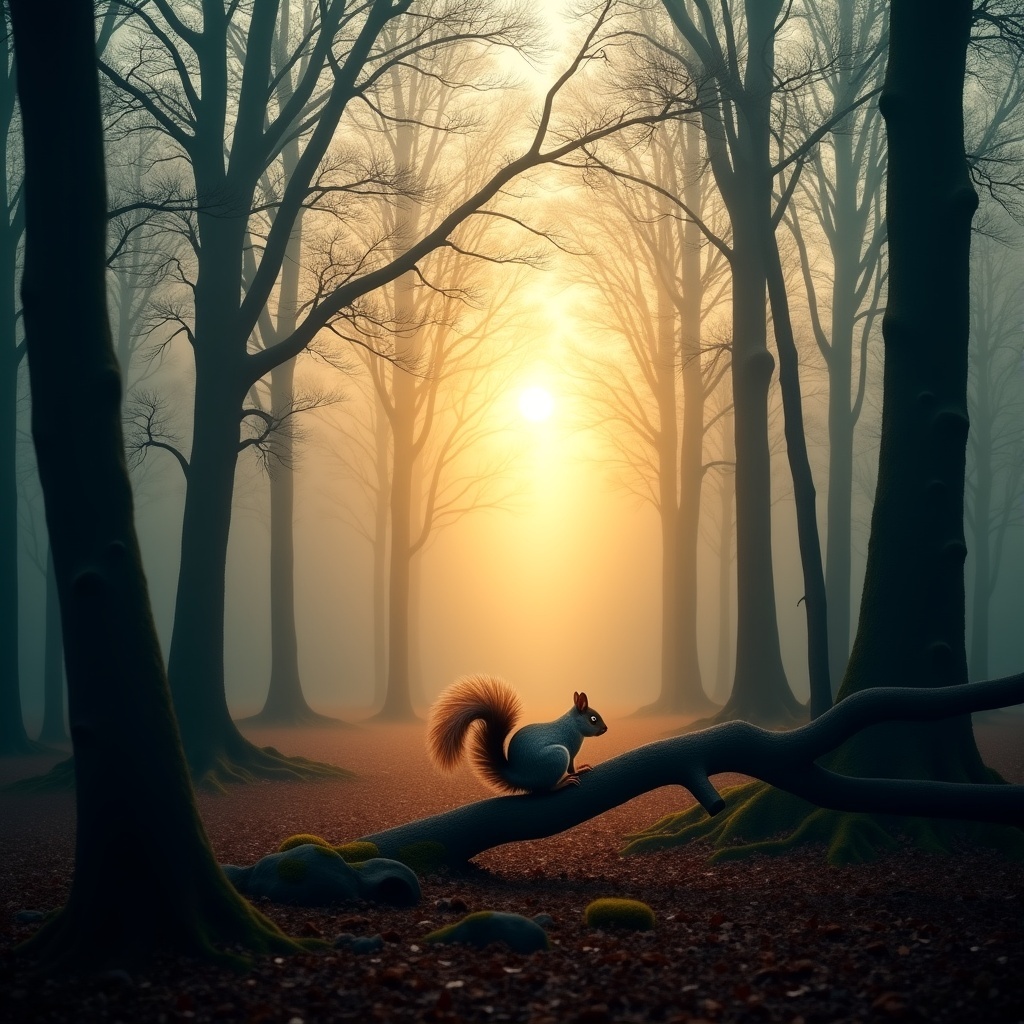}
        }
        \vspace{-0.6cm}
        \caption*{\centering \small center=7.8}
    \end{minipage}
    \begin{minipage}[b]{0.18\textwidth}
        \centering
        \subfigure{
            \includegraphics[width=\textwidth]{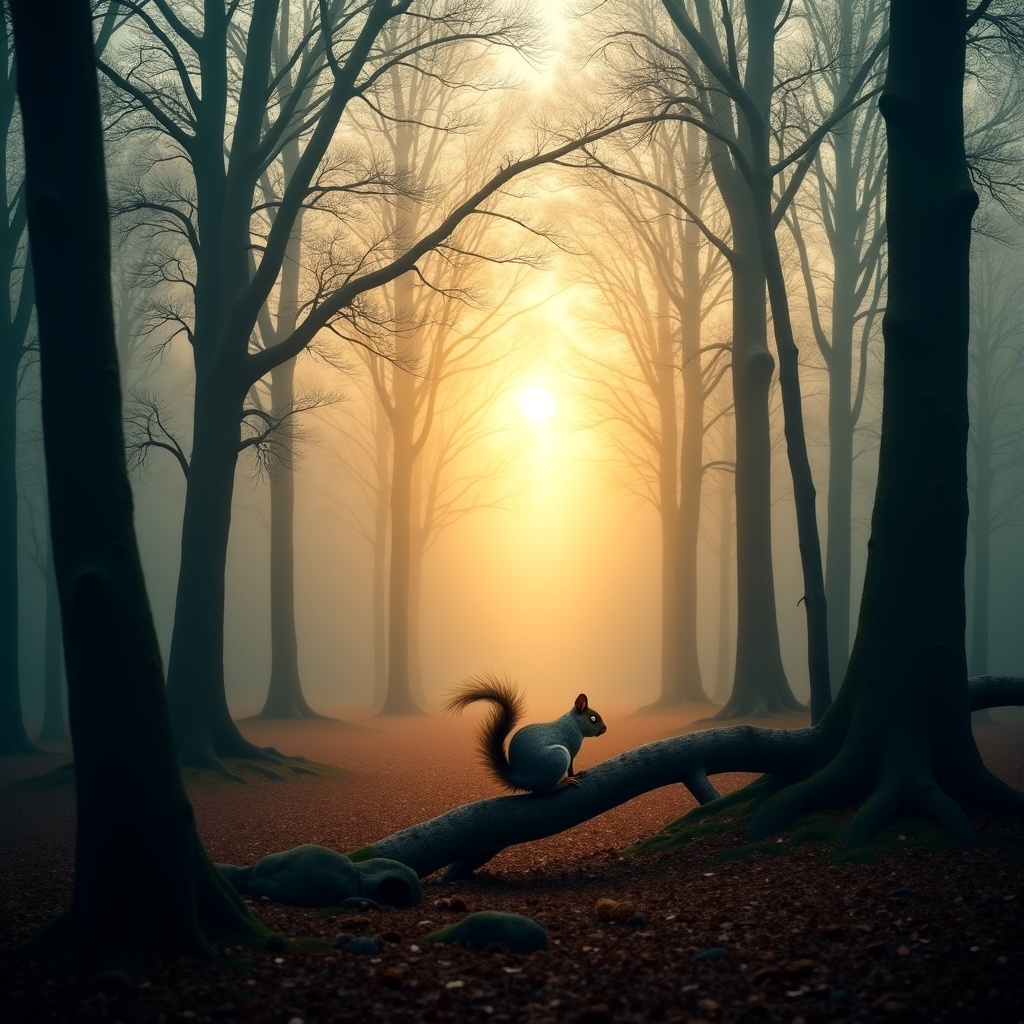}
        }
        \vspace{-0.6cm}
        \caption*{\centering \small center=8.3}
    \end{minipage}
    \begin{minipage}[b]{0.18\textwidth}
        \centering
        \subfigure{
            \includegraphics[width=\textwidth]{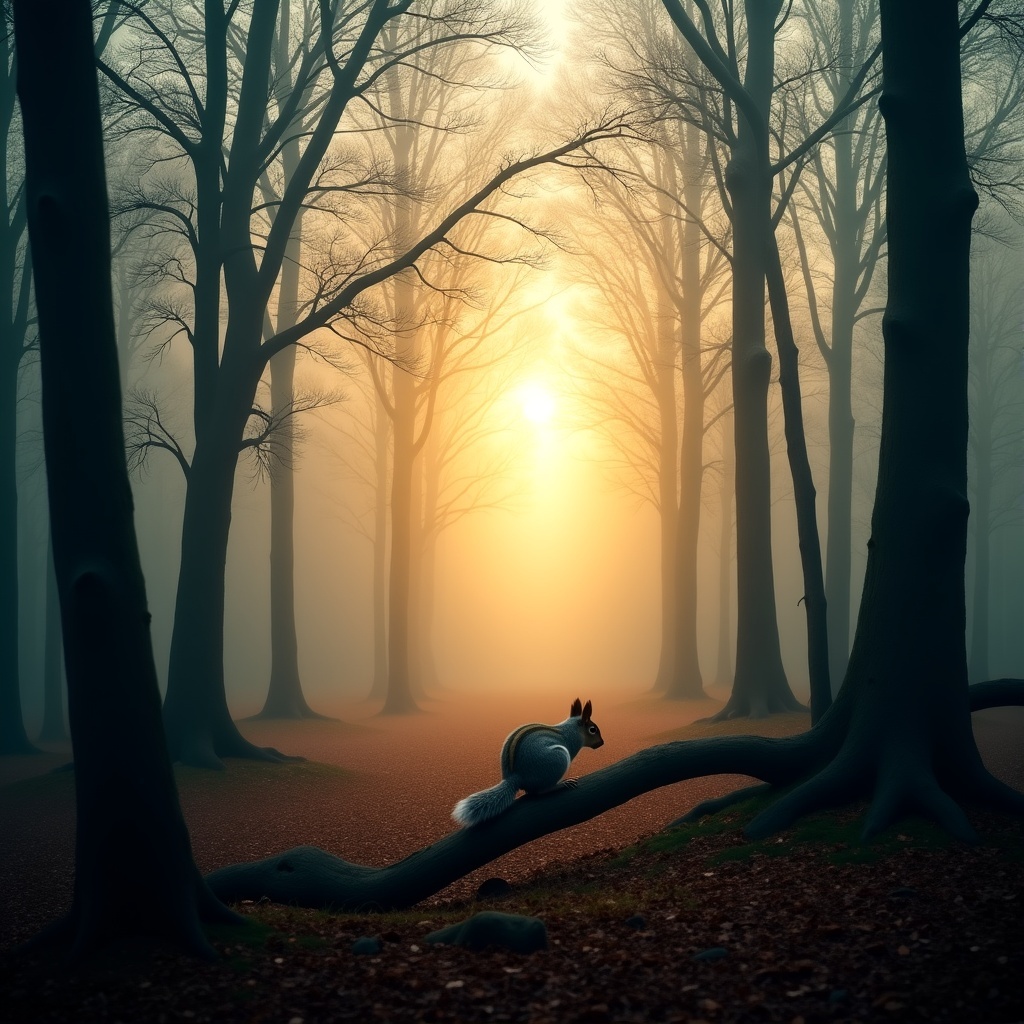}
        }
        \vspace{-0.6cm}
        \caption*{\centering \small center=8.9}
    \end{minipage}
    \begin{minipage}[b]{0.18\textwidth}
        \centering
        \subfigure{
            \includegraphics[width=\textwidth]{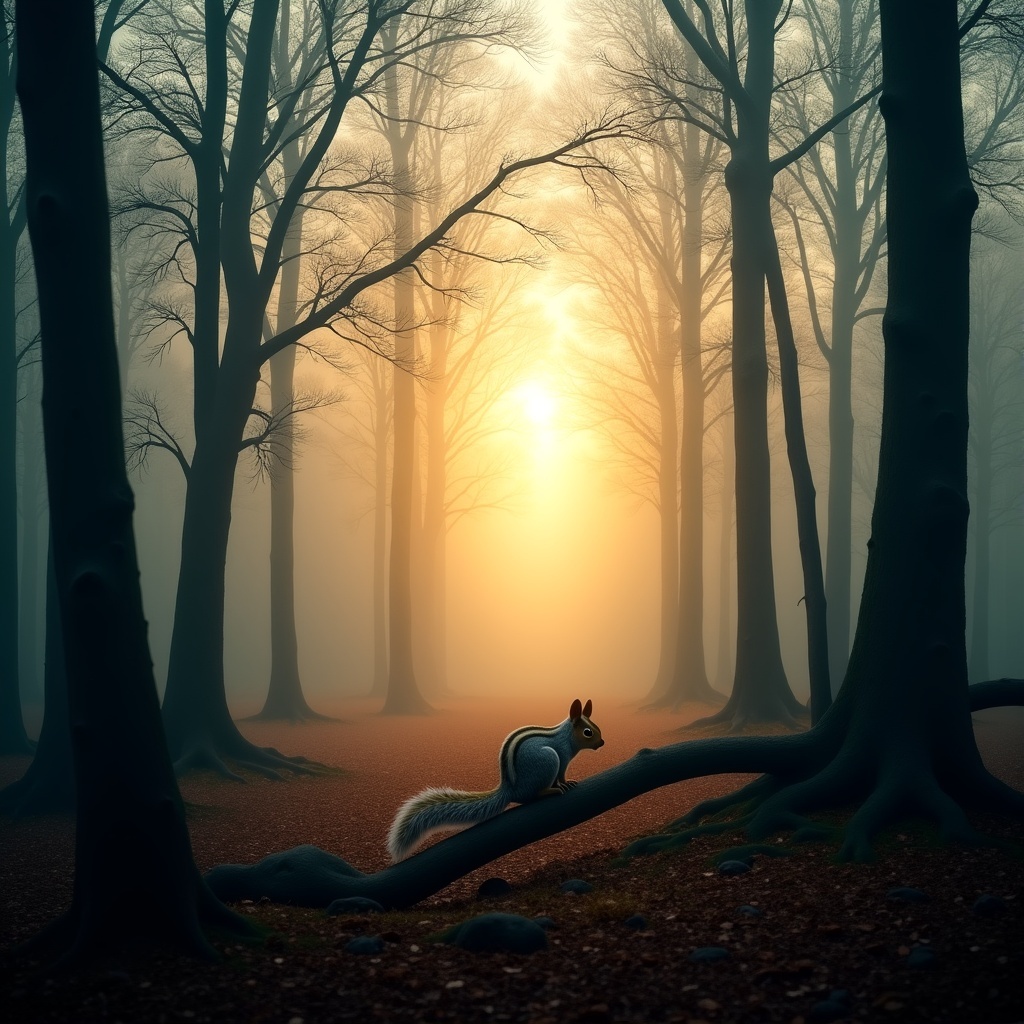}
        }
        \vspace{-0.6cm}
        \caption*{\centering \small center=9.4}
    \end{minipage}
    \begin{minipage}[b]{0.18\textwidth}
        \centering
        \subfigure{
            \includegraphics[width=\textwidth]{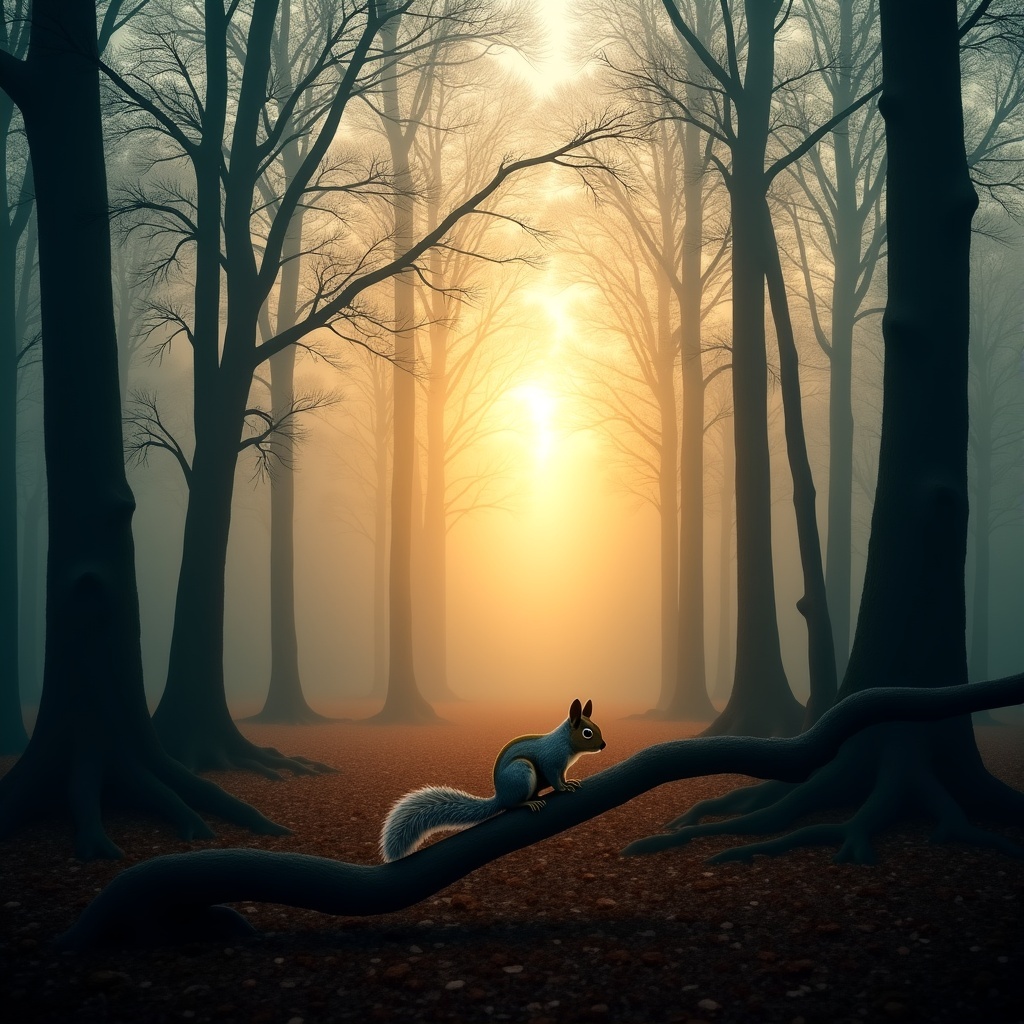}
        }
        \vspace{-0.6cm}
        \caption*{\centering \small center=10}
    \end{minipage}
    \caption{Comparison of different parameter centers. (arctan) (scale=0.5)}
    \label{6}
\end{figure}

\begin{figure}[!htbp]
    \centering
    \begin{minipage}[b]{0.18\textwidth}
        \centering
        \subfigure{
            \includegraphics[width=\textwidth]{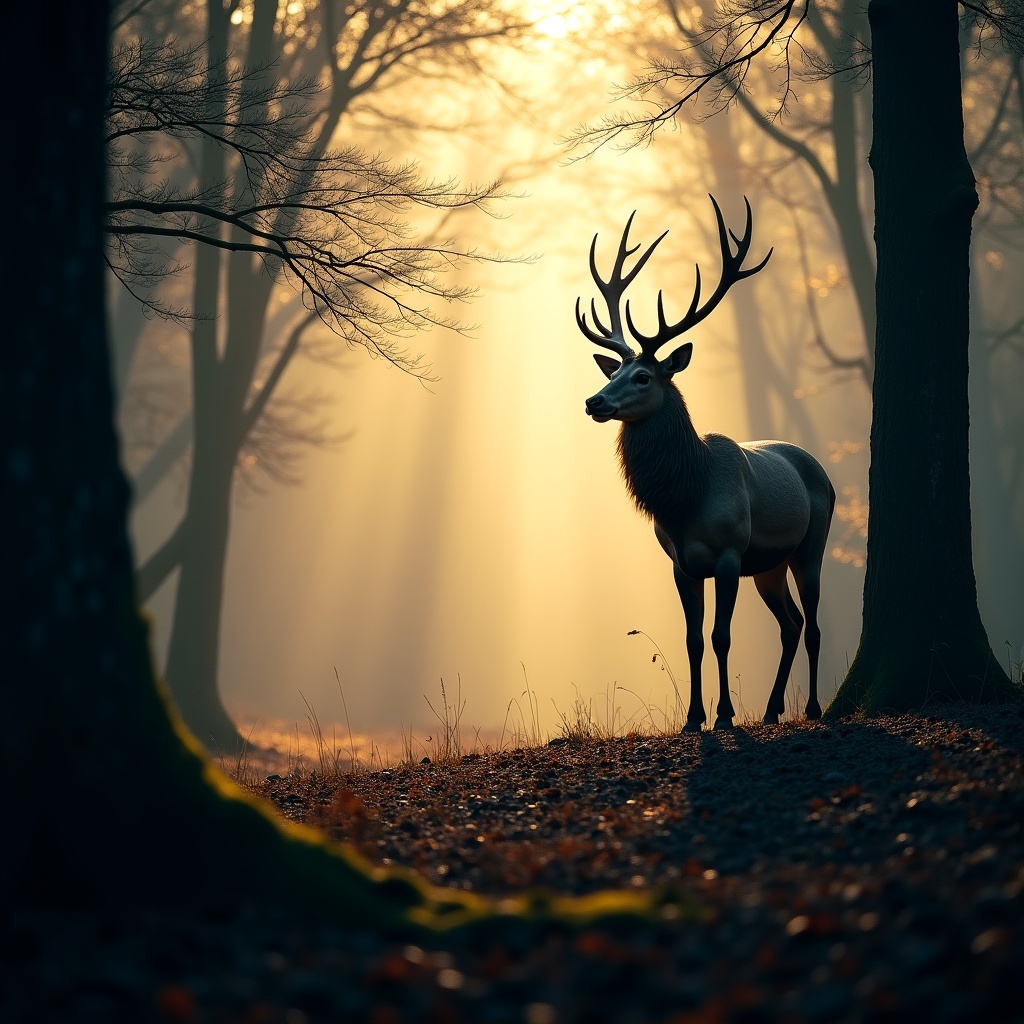}
        }
    \end{minipage}
    \begin{minipage}[b]{0.18\textwidth}
        \centering
        \subfigure{
            \includegraphics[width=\textwidth]{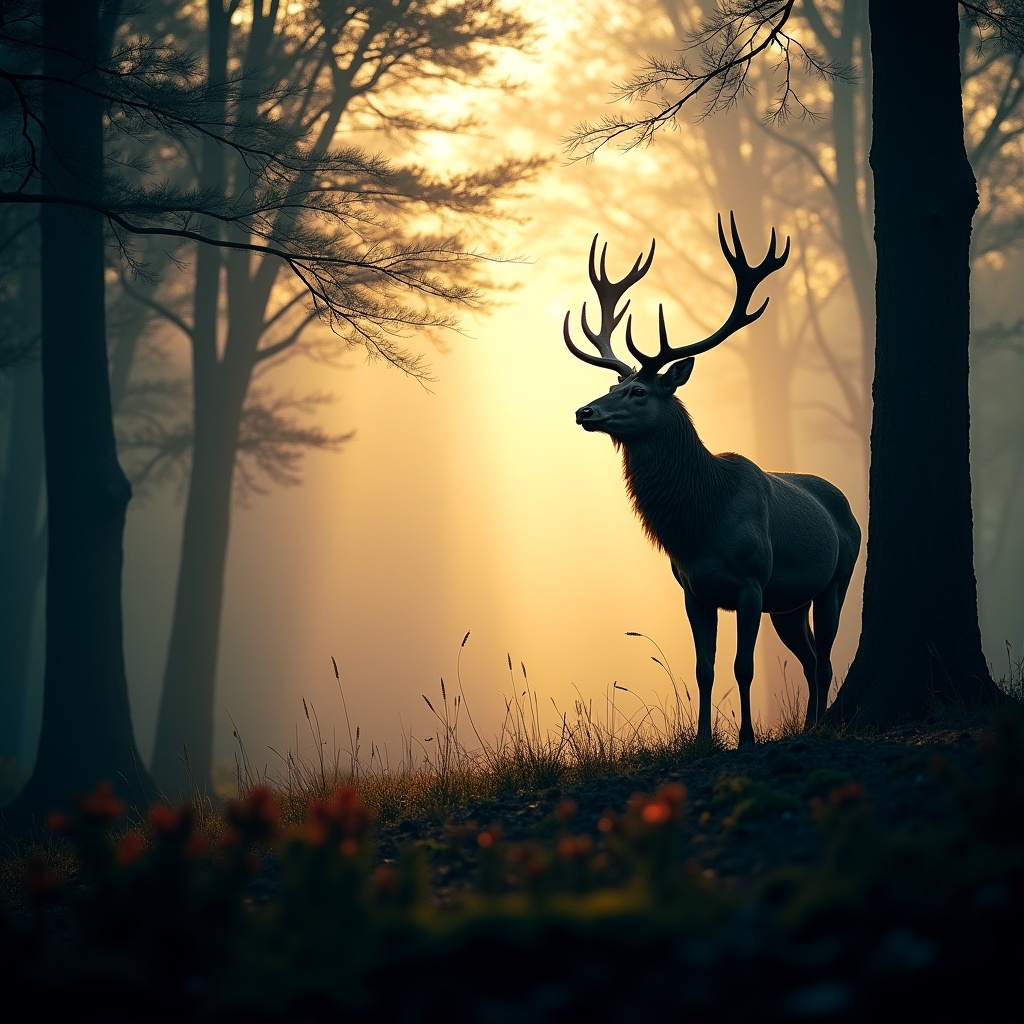}
        }
    \end{minipage}
    \begin{minipage}[b]{0.18\textwidth}
        \centering
        \subfigure{
            \includegraphics[width=\textwidth]{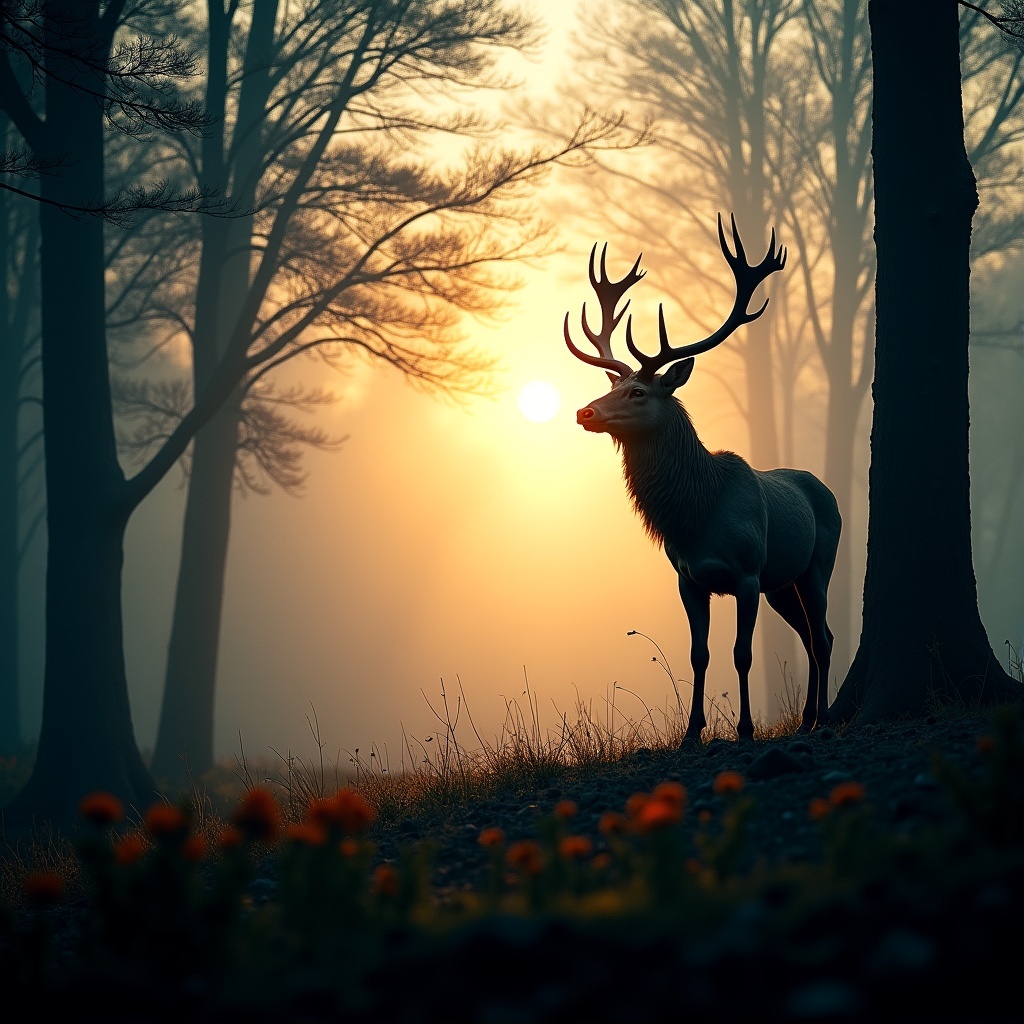}
        }
    \end{minipage}
    \begin{minipage}[b]{0.18\textwidth}
        \centering
        \subfigure{
            \includegraphics[width=\textwidth]{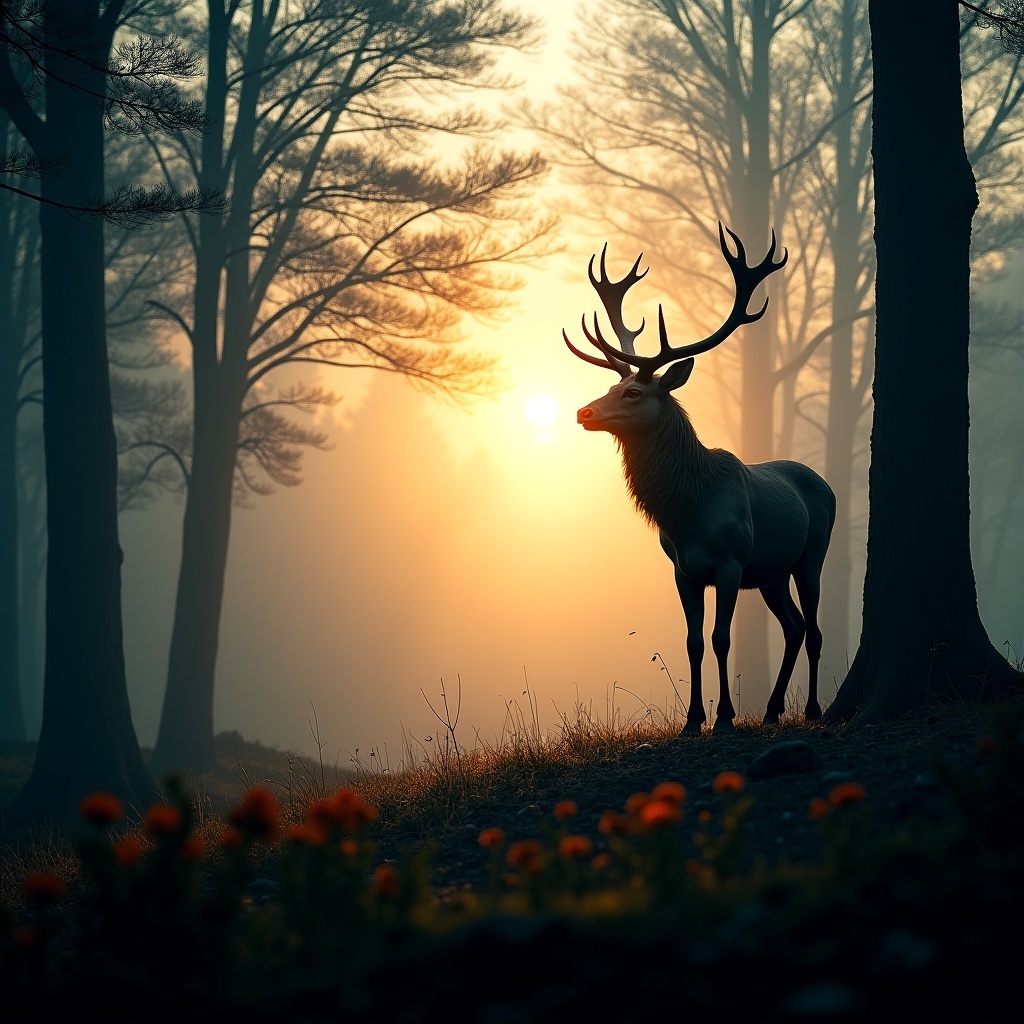}
        }
    \end{minipage}
    \begin{minipage}[b]{0.18\textwidth}
        \centering
        \subfigure{
            \includegraphics[width=\textwidth]{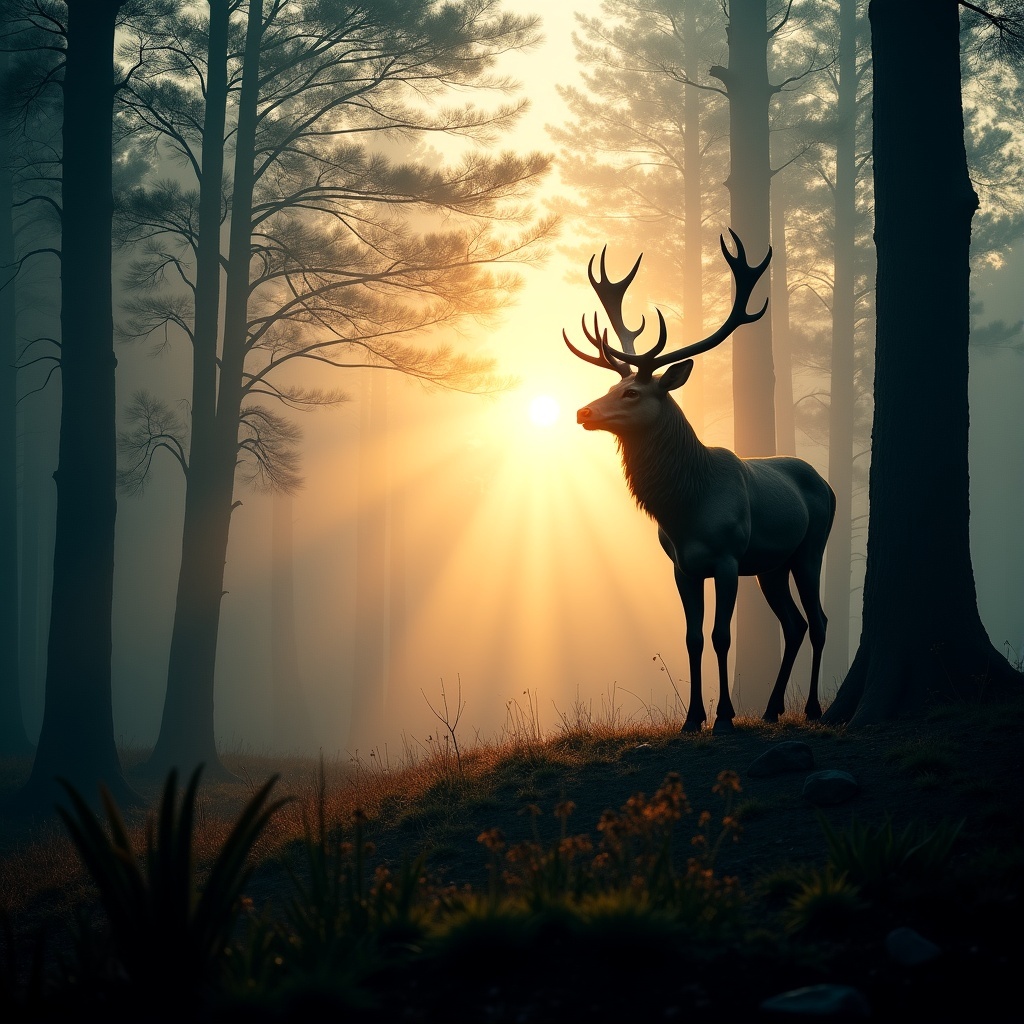}
        }
    \end{minipage}

    \begin{minipage}[b]{0.18\textwidth}
        \centering
        \subfigure{
            \includegraphics[width=\textwidth]{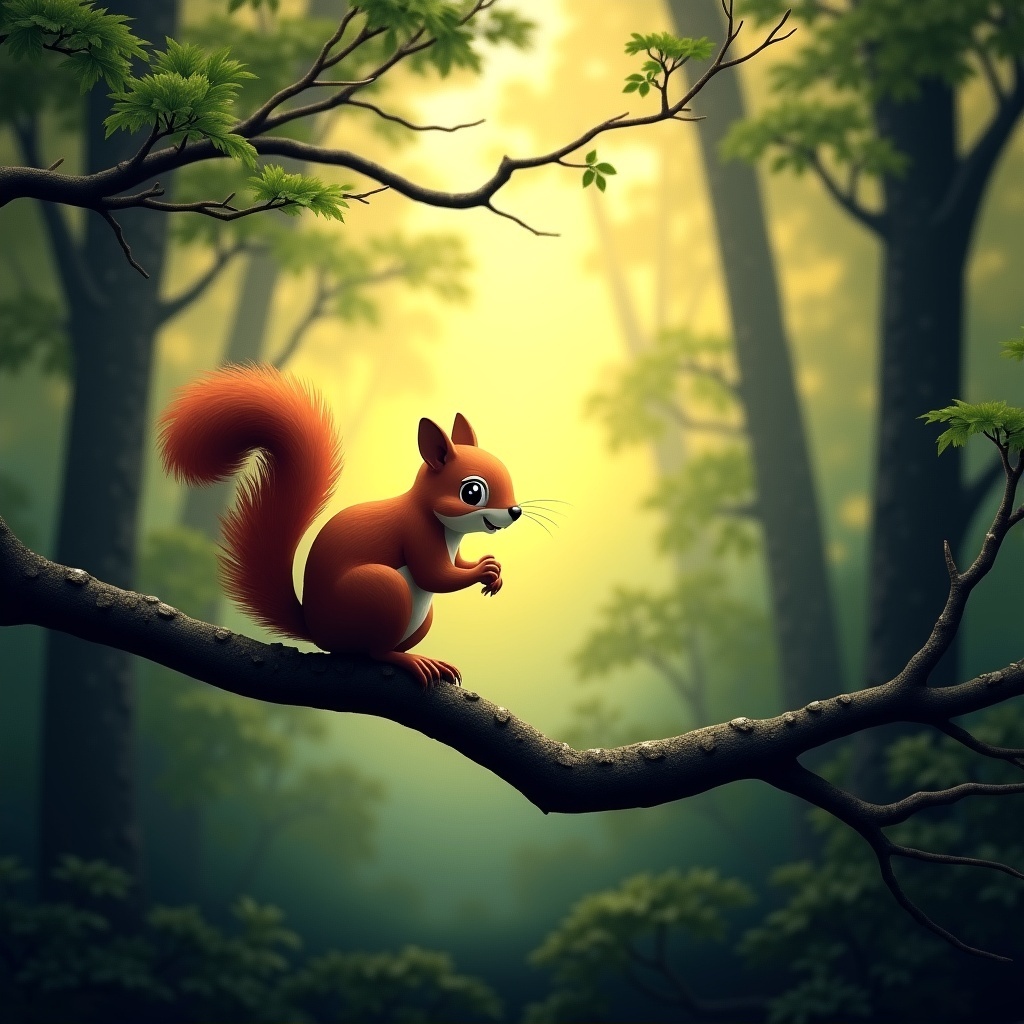}
        }
        \vspace{-0.6cm}
        \caption*{\centering \small center=3}
    \end{minipage}
    \begin{minipage}[b]{0.18\textwidth}
        \centering
        \subfigure{
            \includegraphics[width=\textwidth]{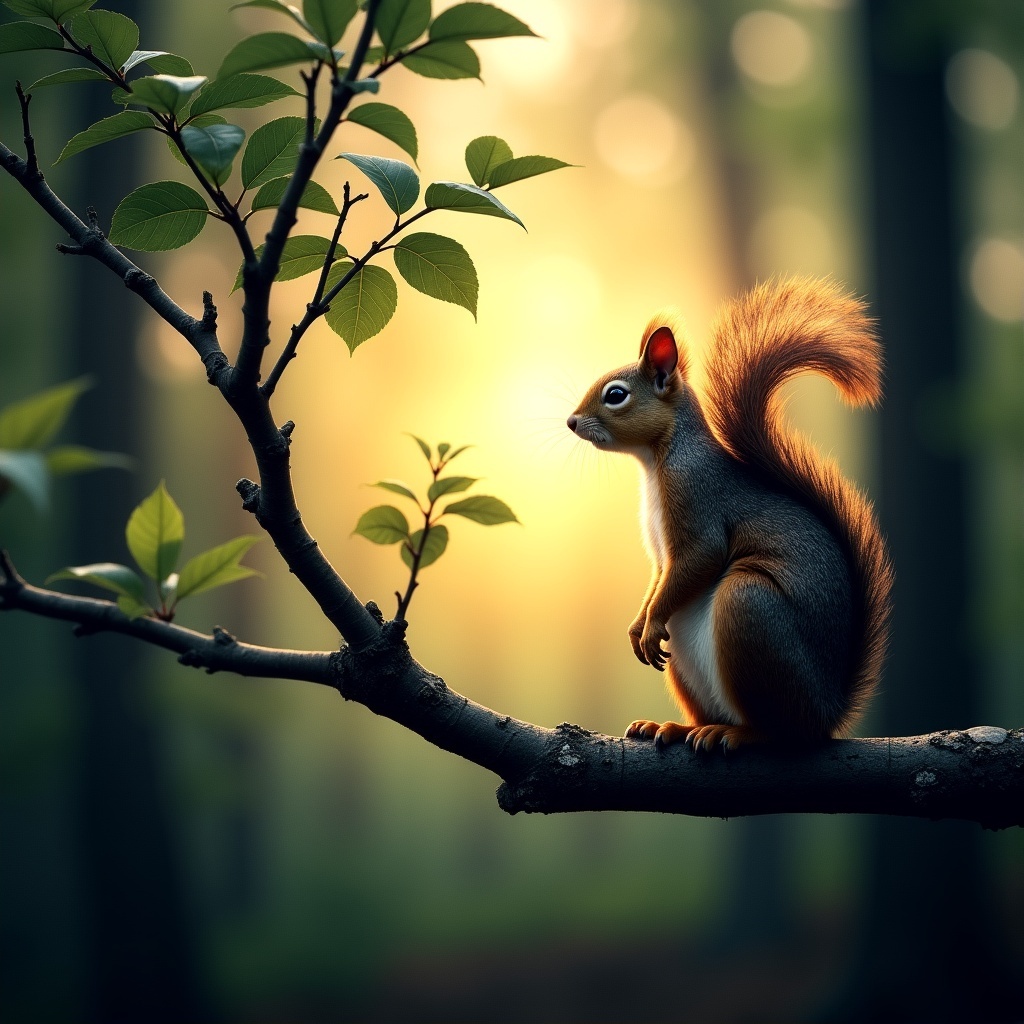}
        }
        \vspace{-0.6cm}
        \caption*{\centering \small center=4}
    \end{minipage}
    \begin{minipage}[b]{0.18\textwidth}
        \centering
        \subfigure{
            \includegraphics[width=\textwidth]{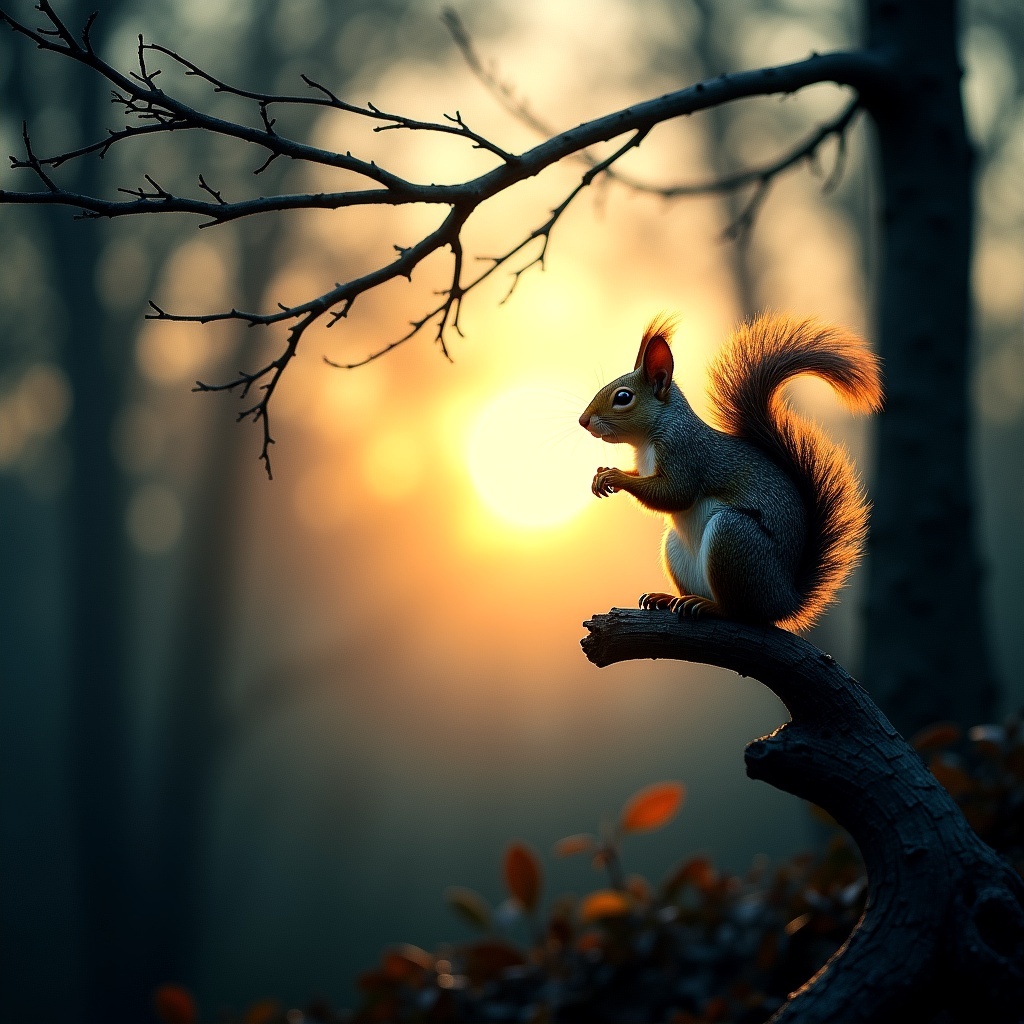}
        }
        \vspace{-0.6cm}
        \caption*{\centering \small center=5}
    \end{minipage}
    \begin{minipage}[b]{0.18\textwidth}
        \centering
        \subfigure{
            \includegraphics[width=\textwidth]{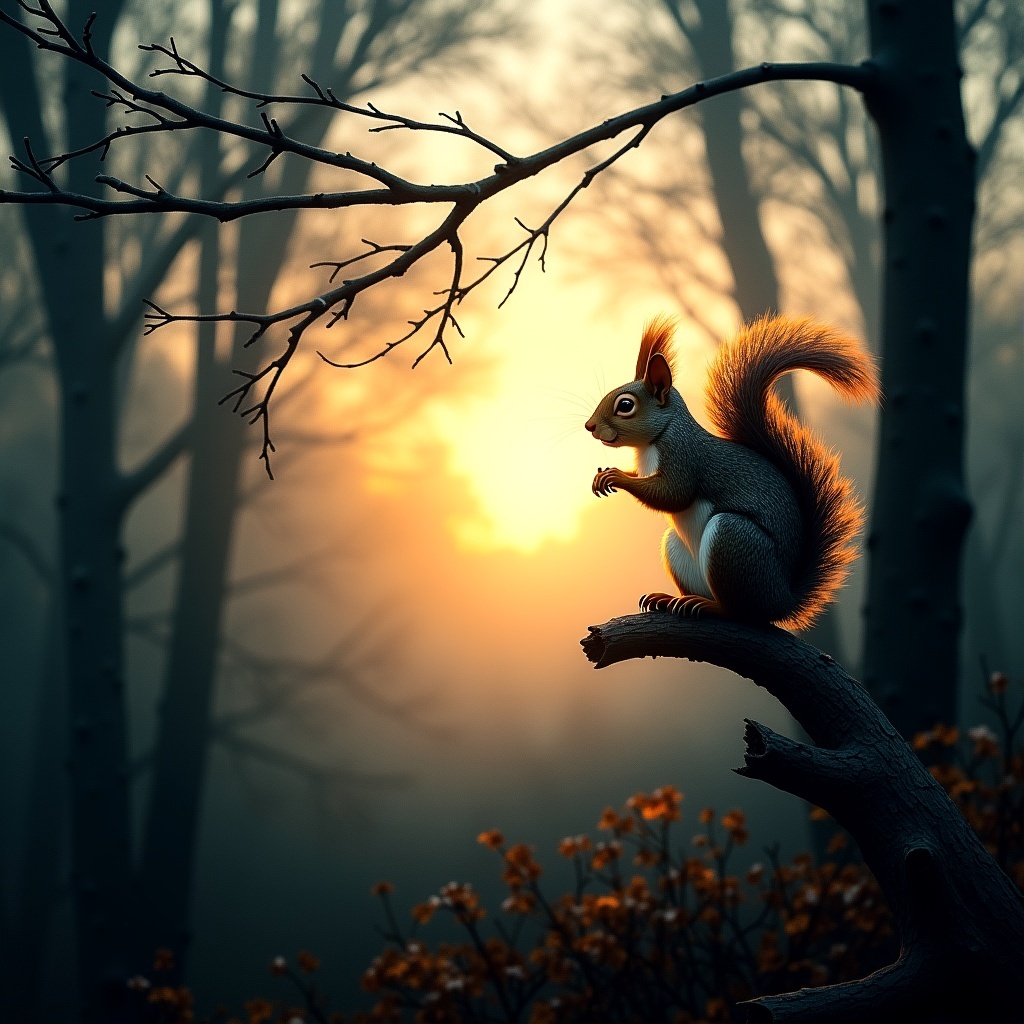}
        }
        \vspace{-0.6cm}
        \caption*{\centering \small center=6}
    \end{minipage}
    \begin{minipage}[b]{0.18\textwidth}
        \centering
        \subfigure{
            \includegraphics[width=\textwidth]{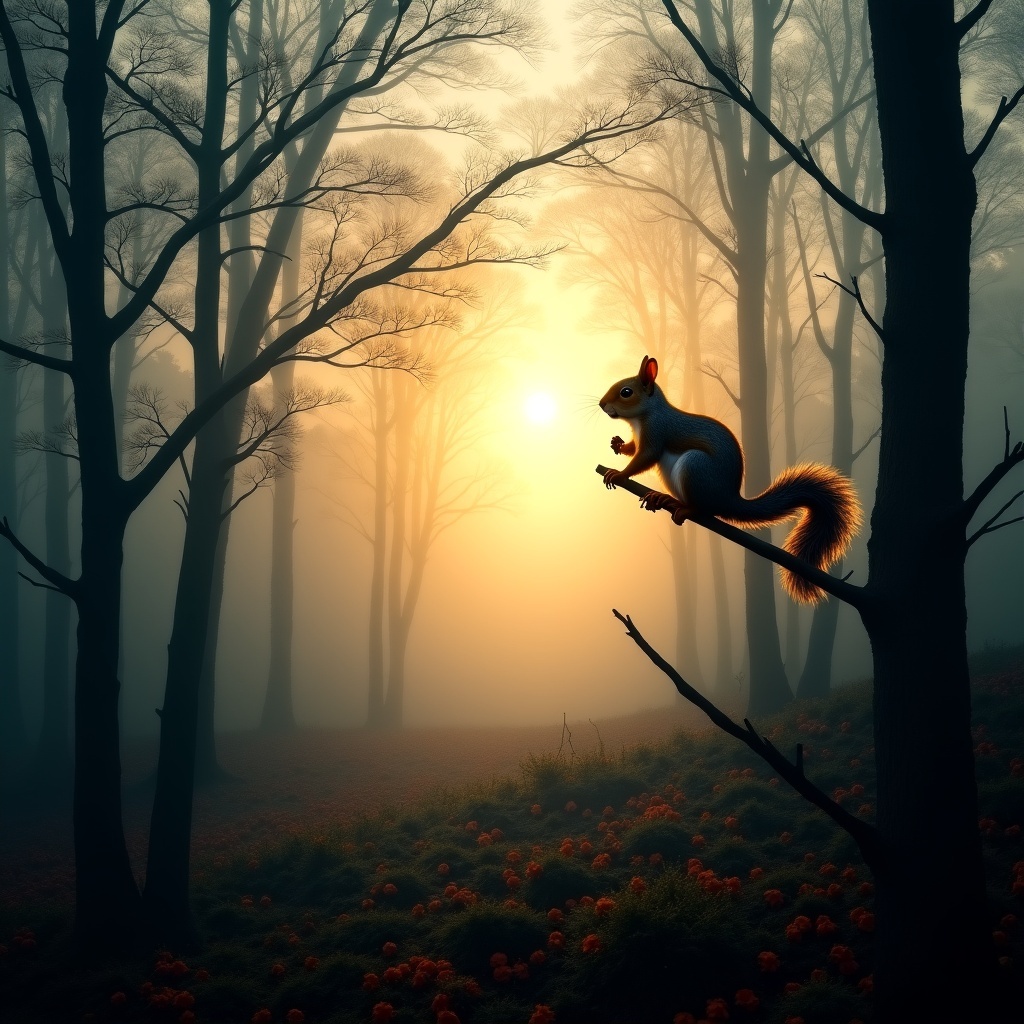}
        }
        \vspace{-0.6cm}
        \caption*{\centering \small center=7}
    \end{minipage}

    \vspace{0.1cm}
    
    \begin{minipage}[b]{0.18\textwidth}
        \centering
        \subfigure{
            \includegraphics[width=\textwidth]{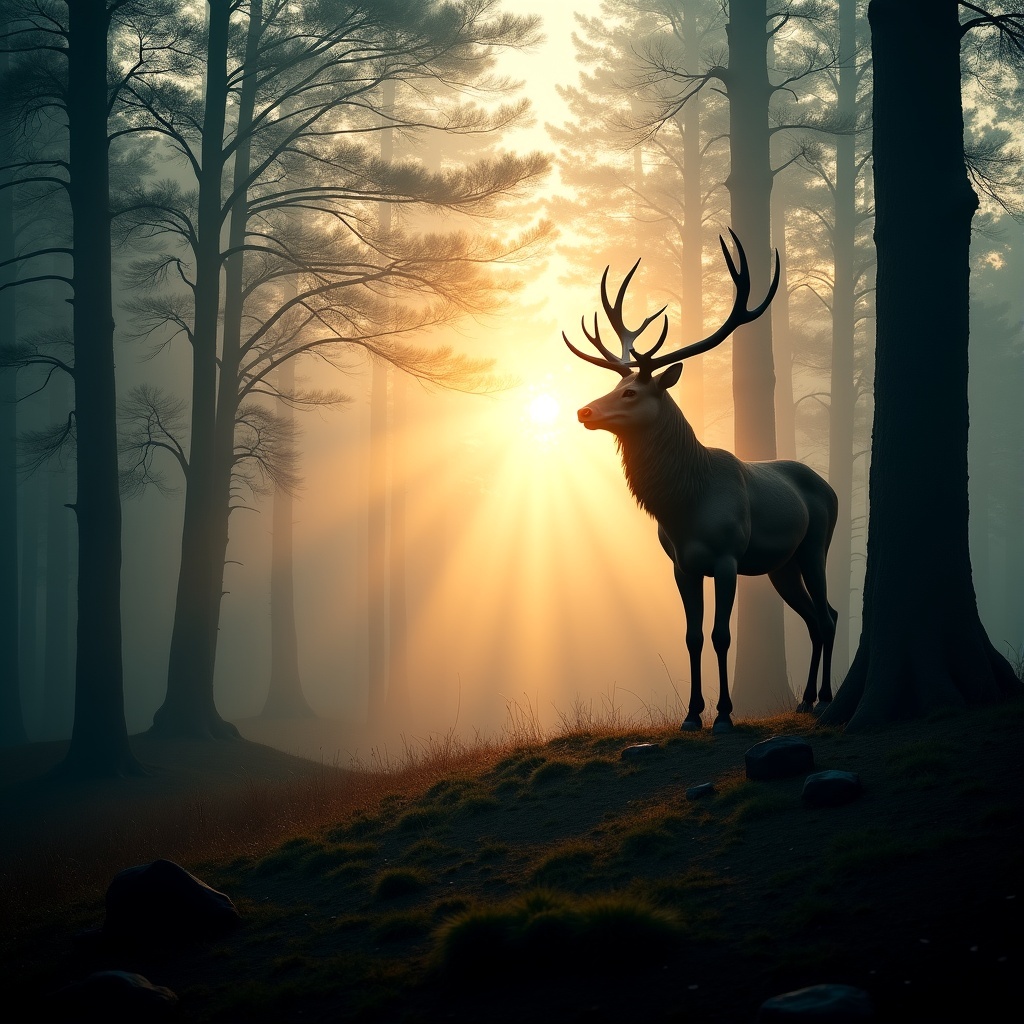}
        }
    \end{minipage}
    \begin{minipage}[b]{0.18\textwidth}
        \centering
        \subfigure{
            \includegraphics[width=\textwidth]{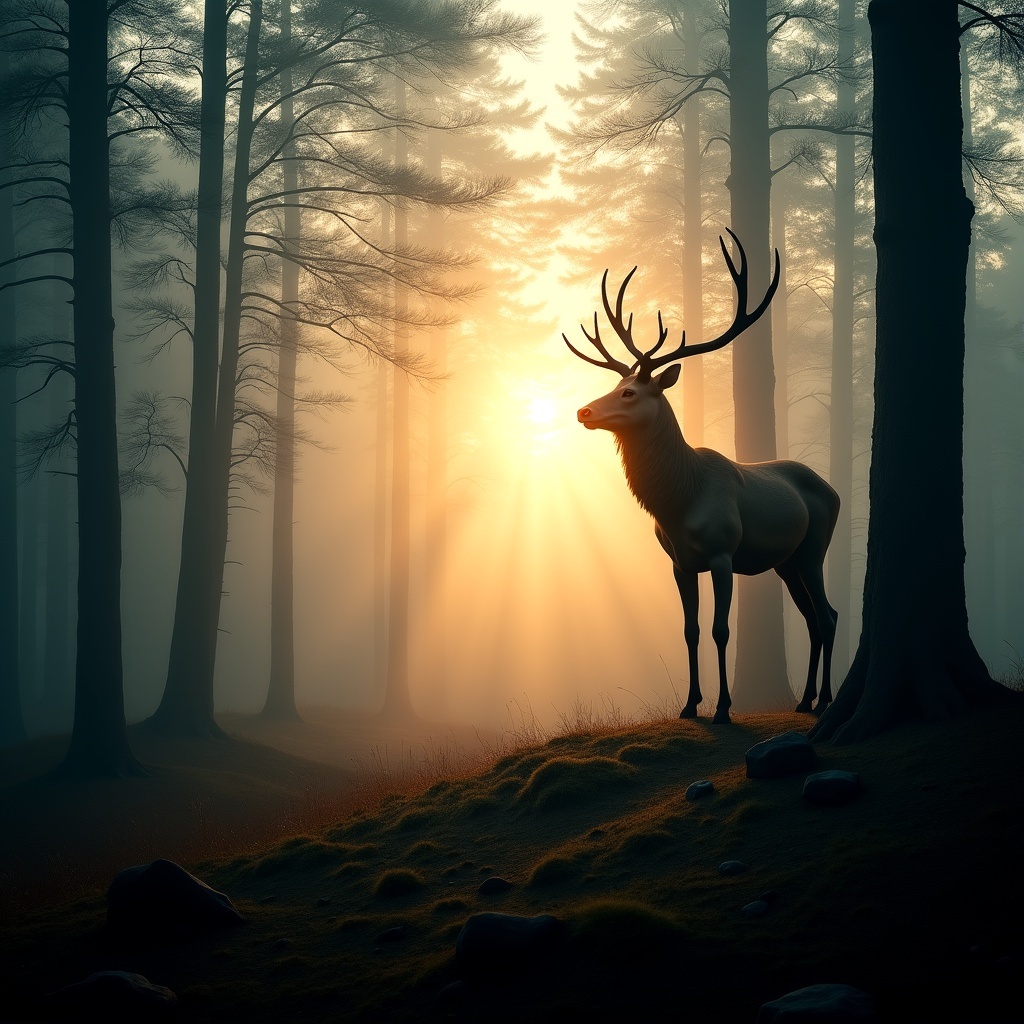}
        }
    \end{minipage}
    \begin{minipage}[b]{0.18\textwidth}
        \centering
        \subfigure{
            \includegraphics[width=\textwidth]{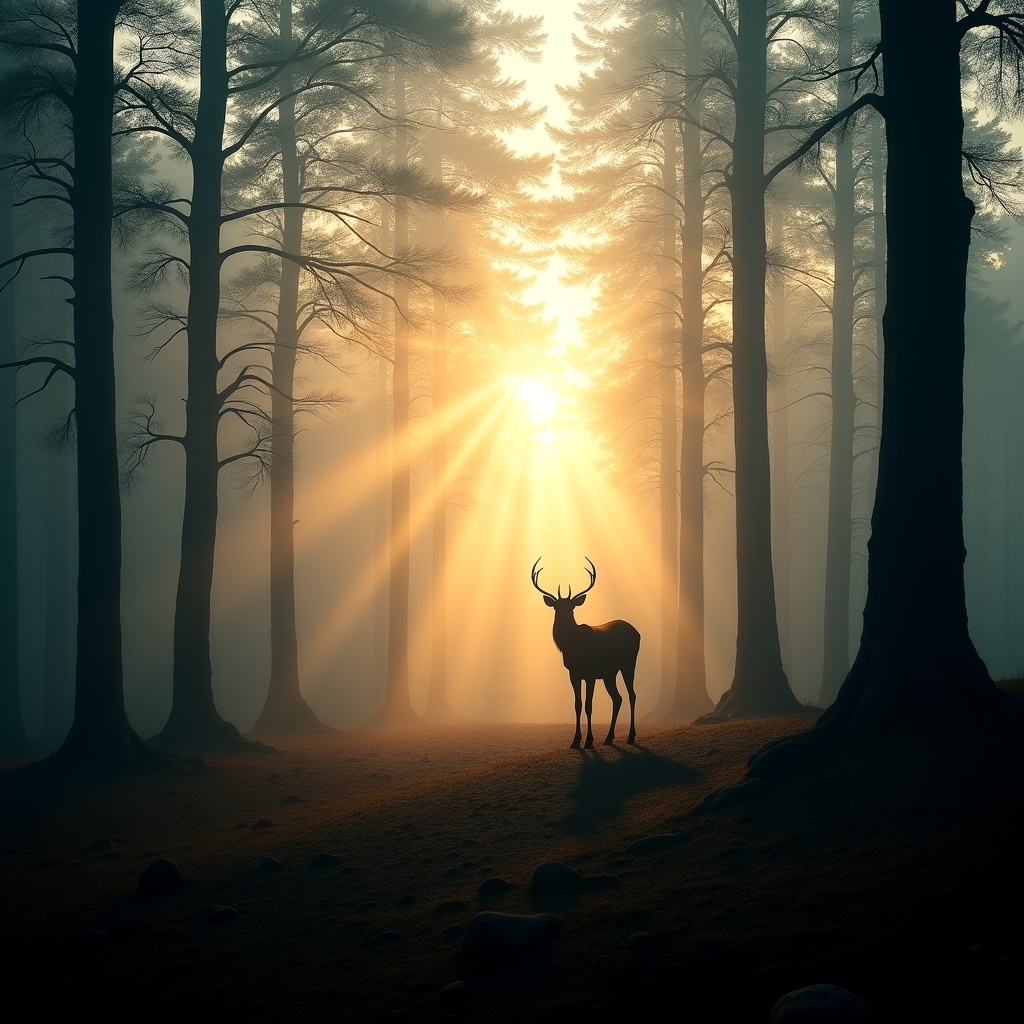}
        }
    \end{minipage}
    \begin{minipage}[b]{0.18\textwidth}
        \centering
        \subfigure{
            \includegraphics[width=\textwidth]{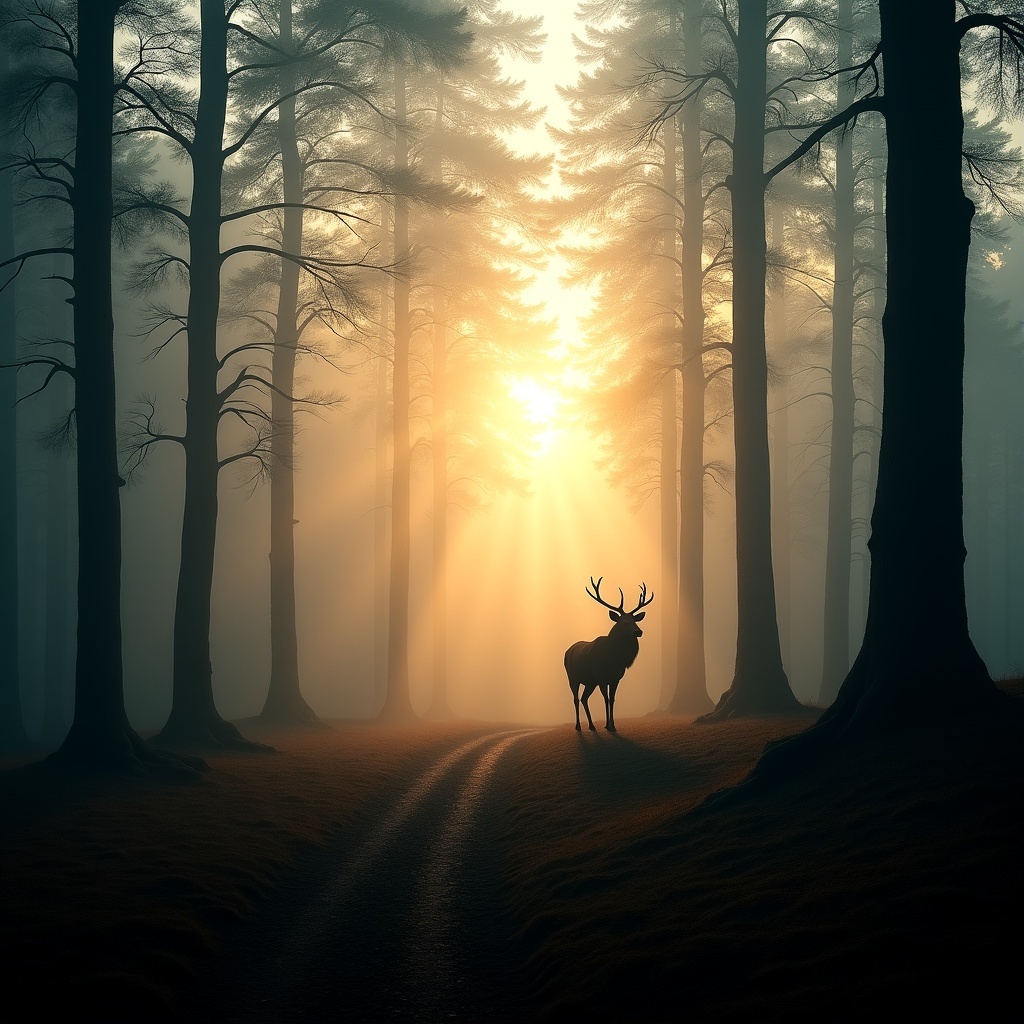}
        }
    \end{minipage}
    \begin{minipage}[b]{0.18\textwidth}
        \centering
        \subfigure{
            \includegraphics[width=\textwidth]{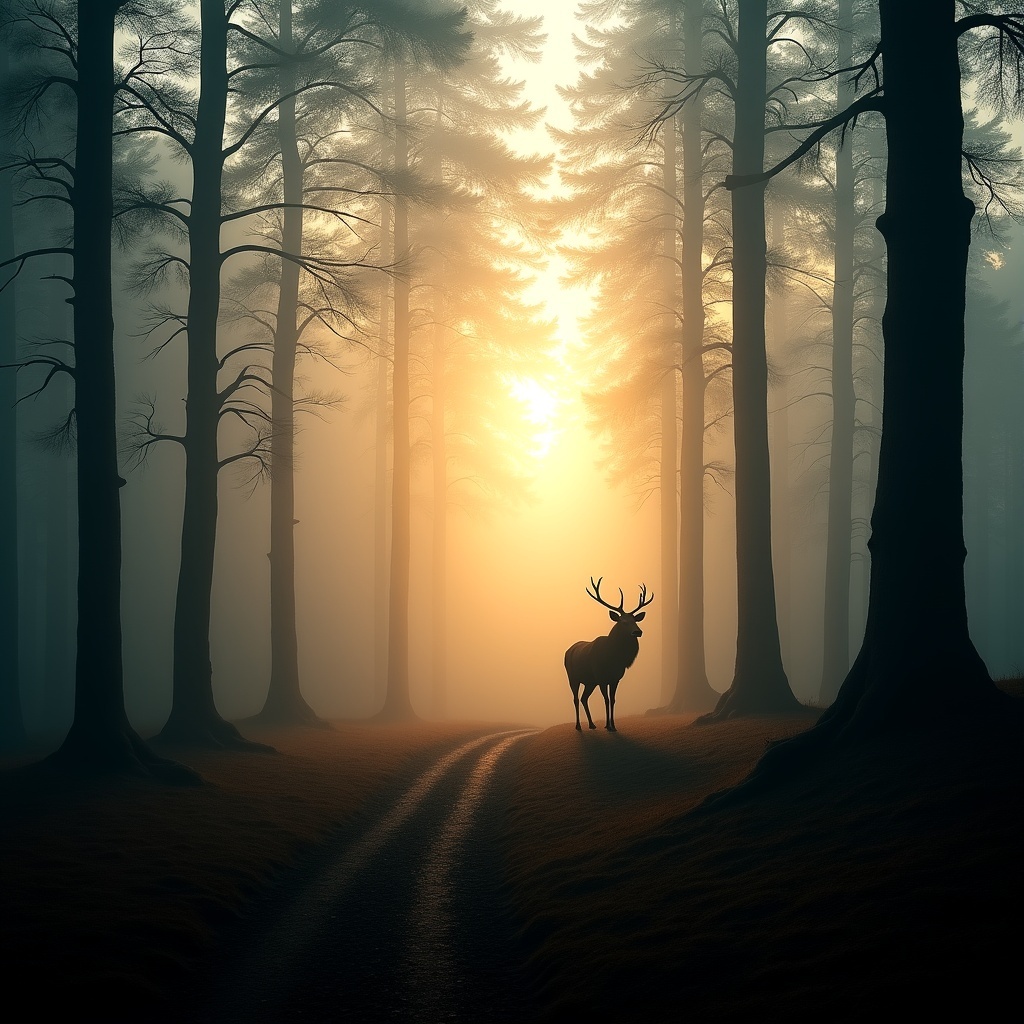}
        }
    \end{minipage}

    \begin{minipage}[b]{0.18\textwidth}
        \centering
        \subfigure{
            \includegraphics[width=\textwidth]{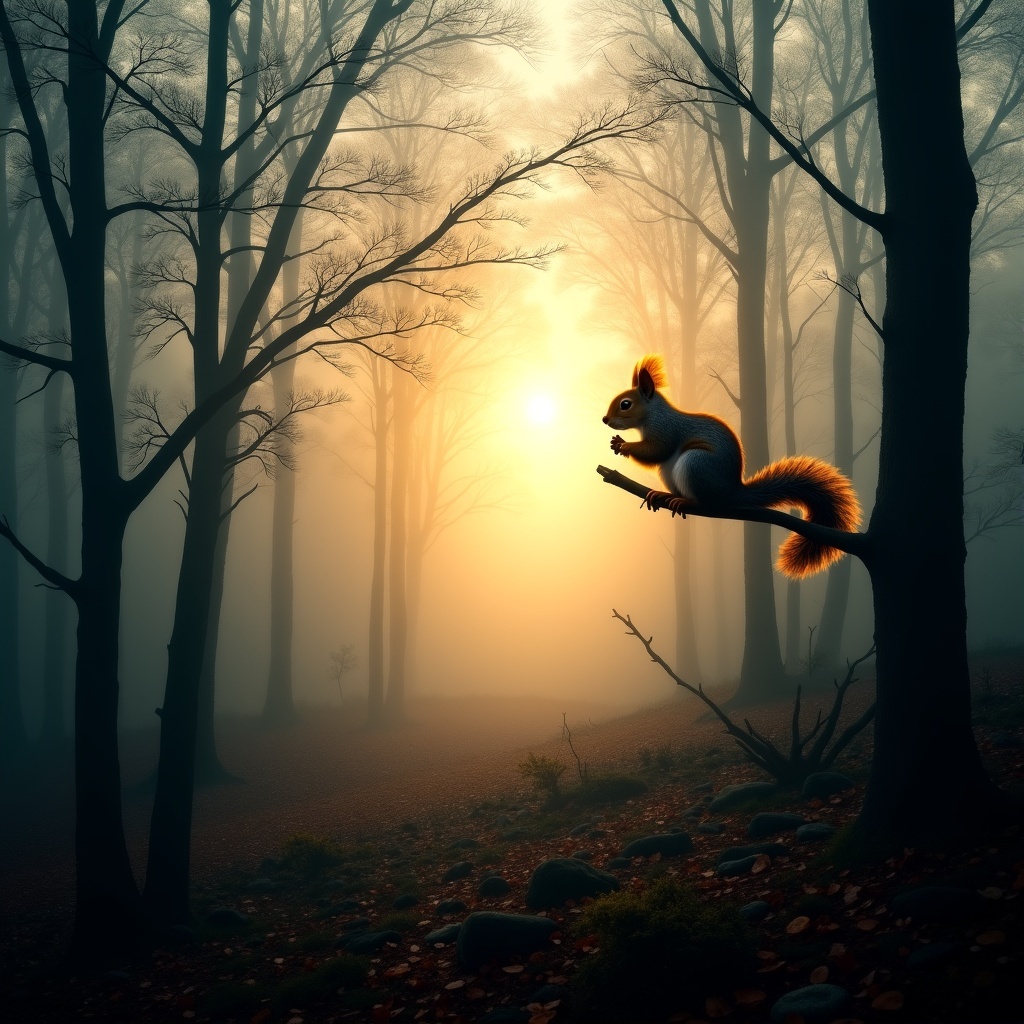}
        }
        \vspace{-0.6cm}
        \caption*{\centering \small center=8}
    \end{minipage}
    \begin{minipage}[b]{0.18\textwidth}
        \centering
        \subfigure{
            \includegraphics[width=\textwidth]{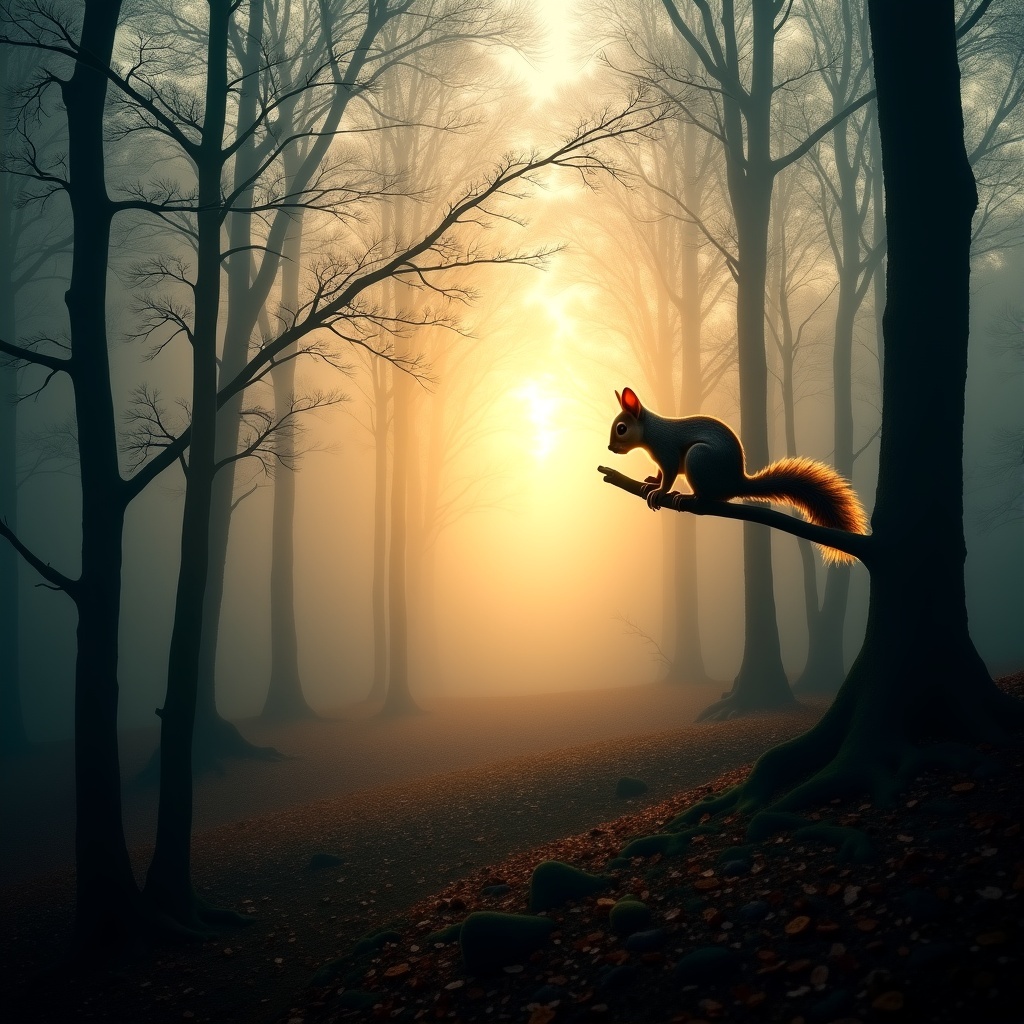}
        }
        \vspace{-0.6cm}
        \caption*{\centering \small center=9}
    \end{minipage}
    \begin{minipage}[b]{0.18\textwidth}
        \centering
        \subfigure{
            \includegraphics[width=\textwidth]{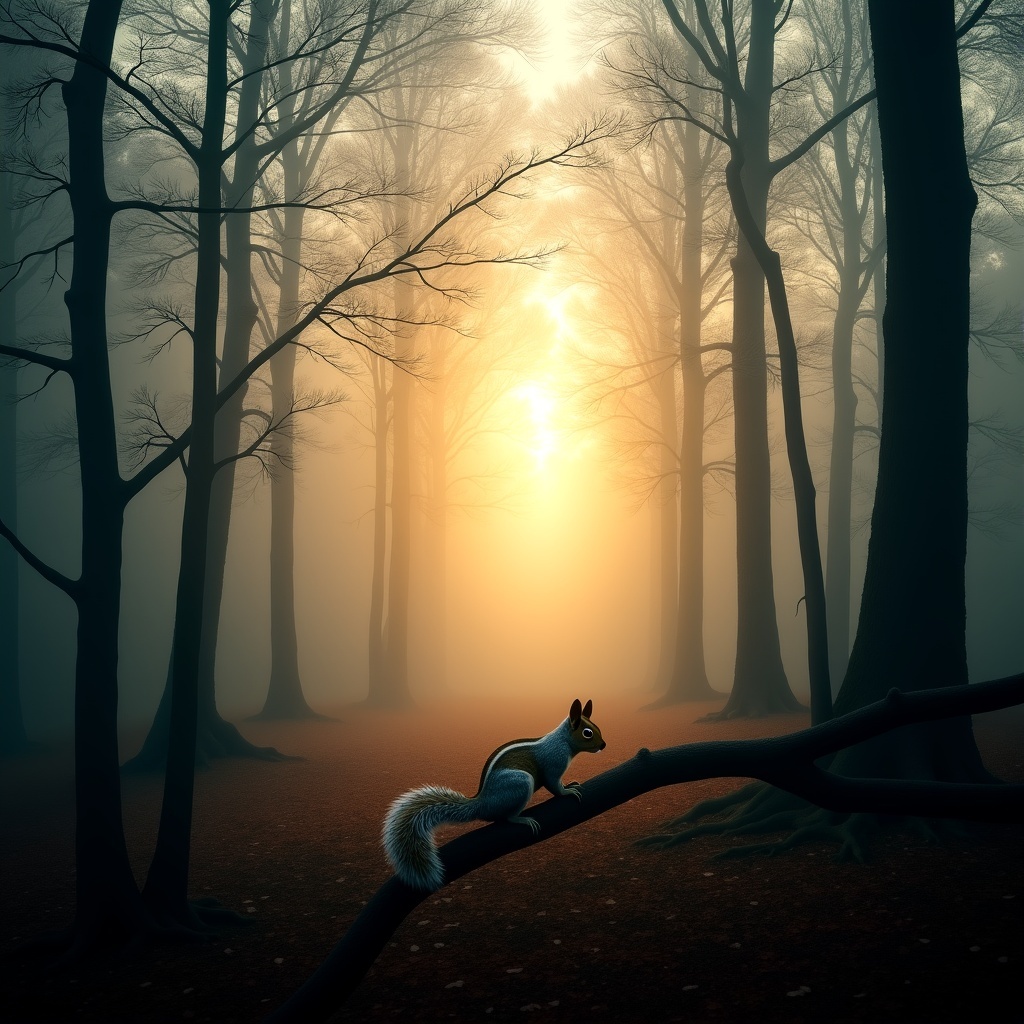}
        }
        \vspace{-0.6cm}
        \caption*{\centering \small center=10}
    \end{minipage}
    \begin{minipage}[b]{0.18\textwidth}
        \centering
        \subfigure{
            \includegraphics[width=\textwidth]{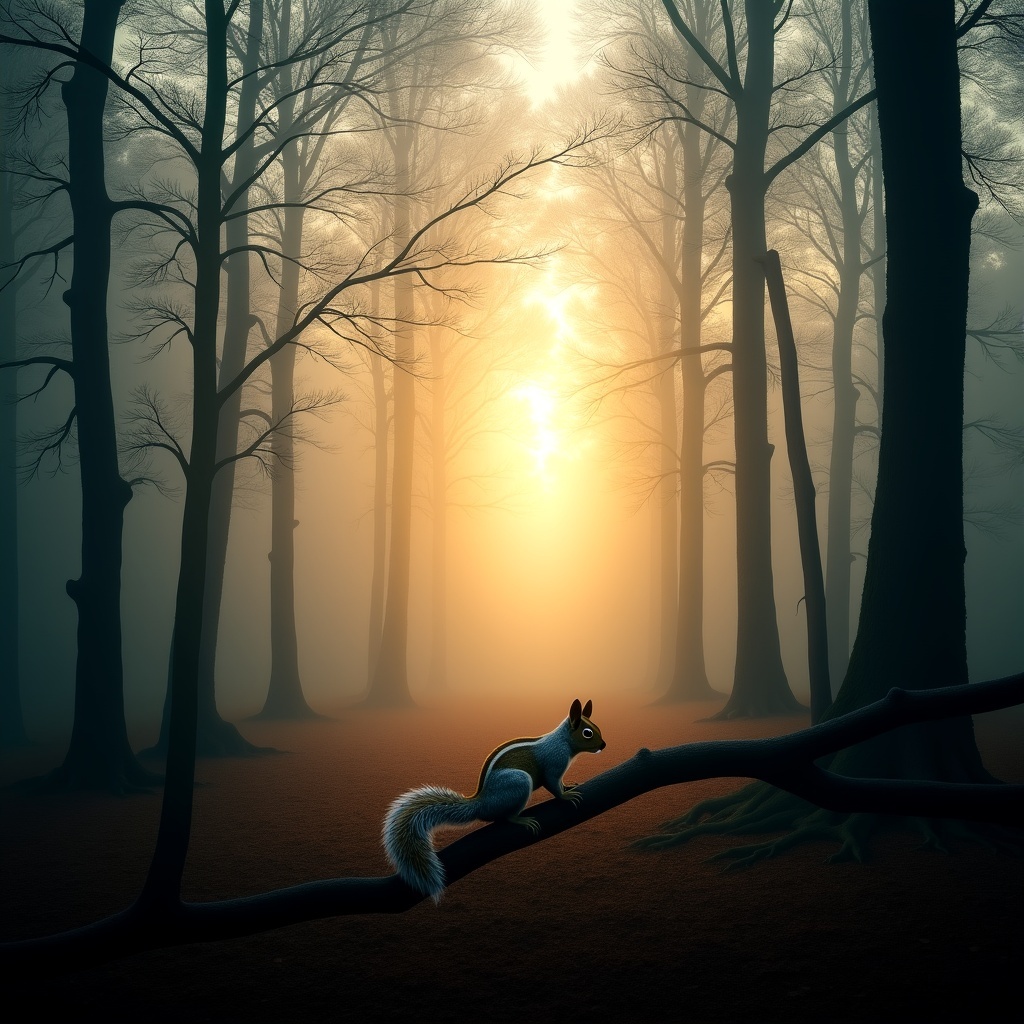}
        }
        \vspace{-0.6cm}
        \caption*{\centering \small center=11}
    \end{minipage}
    \begin{minipage}[b]{0.18\textwidth}
        \centering
        \subfigure{
            \includegraphics[width=\textwidth]{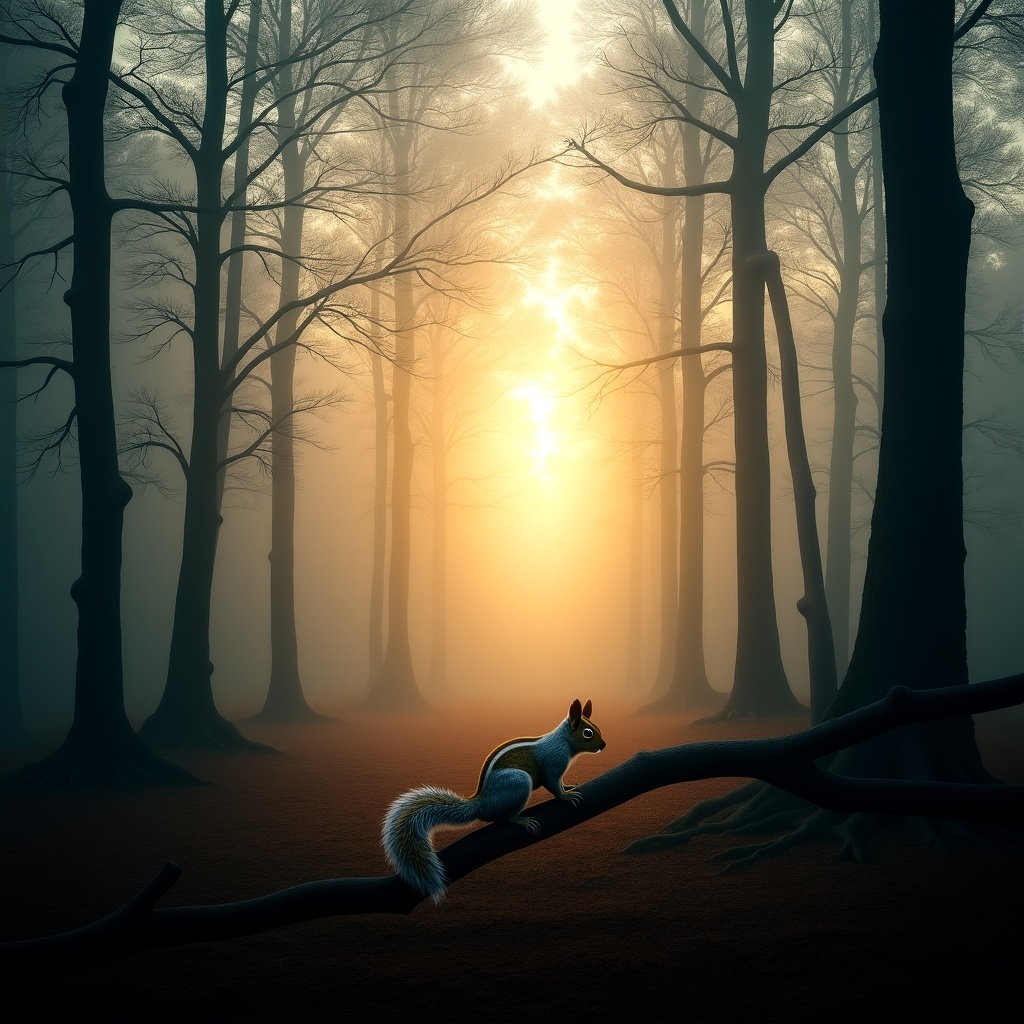}
        }
        \caption*{\centering \small center=12}
        \vspace{-0.6cm}
    \end{minipage}
    \caption{Comparison of different parameter centers. (01)}
    \label{7}
\end{figure}

In this ablation study, we compare the effects of varying the center of the parameterized functions. The prompts we used are \textit{"A majestic stag standing proud in a misty forest at dawn, sunlight through trees."} and \textit{"A squirrel perched on a branch in a misty forest at dawn, sunlight through trees."}. The generated results are shown in Figures \ref{5}, \ref{6}, and \ref{7}.

We observe a clear trade-off between background similarity and text-image alignment. When the center value is small, fewer parameters remain near 0 while more approach 1. This leads to stronger text-image alignment but reduced background similarity. Conversely, increasing the center causes more parameters to stay near 0 and fewer to approach 1, which improves background similarity at the cost of alignment with the text prompt. To navigate this trade-off, we formulate and solve the isotonic optimization problem and identify a balance that achieves both high background similarity and strong text-image alignment.

\section{Prompt Example}
\begin{figure}[htbp]
    \centering
    \begin{tikzpicture}[
        user/.style={rectangle, rounded corners=3mm, fill=blue!15, draw=blue, minimum width=12cm, align=left, text width=11cm, inner sep=5pt},
        llm/.style={rectangle, rounded corners=3mm, fill=gray!15, draw=gray, minimum width=12cm, align=left, text width=11cm, inner sep=5pt},
        node distance=0.6cm and 0.6cm]

        \node[user] (user1) {
            \textbf{User Input:}\\[2pt]
        role: system, content: You are a helpful assistant to arrange prompts for text-to-image generation.\\
        role: user, content: You are given a set of prompts for text-to-image generation. Your task is to first identify and extract the common background shared across all prompts. Then, for each individual prompt, present the distinct entity description. Please format your output as follows:
        Background: [shared background]
        Entity 1: [description of the first unique entity]
        Entity 2: [description of the second unique entity]
        ... and so on.\\
        role: user, content: Prompt 1: A cute Pikachu sits in a cozy room bathed in warm sunshine. The room has wooden flooring and a peaceful, homely atmosphere. Prompt 2: A beautiful girl stands in a cozy room bathed in warm sunshine. The room has wooden flooring and a peaceful, homely atmosphere."
       };

        \node[llm, below=of user1] (llm1) {
            \textbf{LLM Output:}\\[2pt]
            Background: A cozy room bathed in warm sunshine with wooden flooring and a peaceful, homely atmosphere.\\
            Entity 1: A cute Pikachu sits.\\
            Entity 2: A beautiful girl stands.
        };

        \draw[->, thick] (user1.south) -- (llm1.north);

    \end{tikzpicture}
    \caption{A zero-shot example of prompting a pre-trained LLM to extract the shared background from a set of prompts and present the distinct entity descriptions for each individual prompt.}
    \label{example;prompt}

\end{figure}
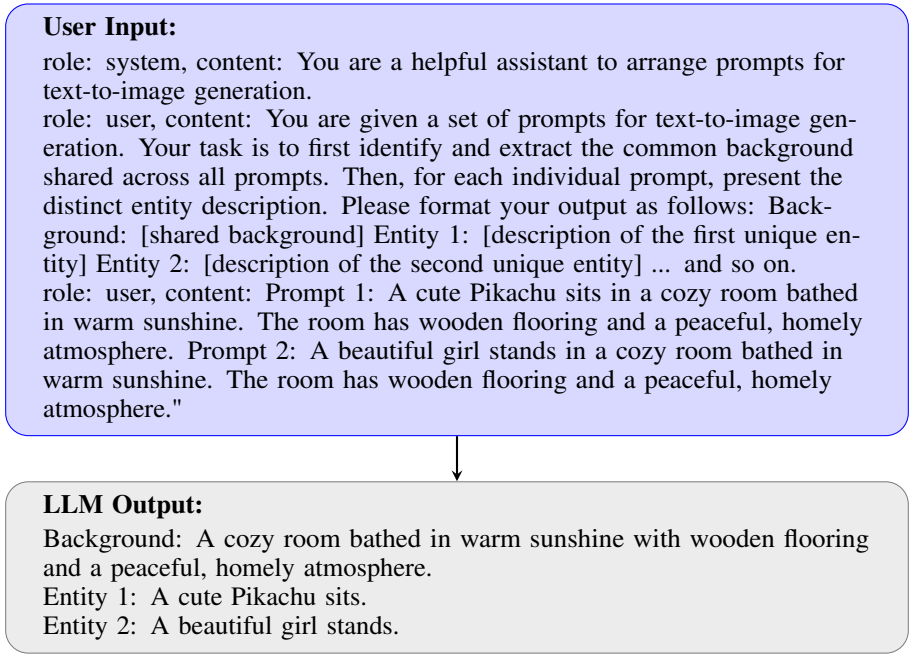

We present the example of prompting a pre-trained LLM to extract the shared background and present the distinct entity descriptions in Figure \ref{example;prompt}.

\end{document}

%% file: math_commands.tex
\usepackage{amsmath,amsfonts,bm}

\def\eqref#1{equation~\ref{#1}}

\def\1{\bm{1}}

\DeclareMathAlphabet{\mathsfit}{\encodingdefault}{\sfdefault}{m}{sl}
\SetMathAlphabet{\mathsfit}{bold}{\encodingdefault}{\sfdefault}{bx}{n}

\newcommand{\softmax}{\mathrm{softmax}}

